\documentclass[journal]{IEEEtran}
\usepackage[numbers,sort]{natbib}

\usepackage{url}%
\usepackage{graphicx}%
\usepackage{multirow}%
\usepackage{amsmath,amssymb,amsfonts}%
\usepackage{amsthm}%
\usepackage{mathrsfs}%
\usepackage{xcolor}%
\usepackage{textcomp}%
\usepackage{booktabs}%
\usepackage{array}%
\usepackage{rotating}%
\usepackage{algorithm}%
\usepackage{algorithmicx}%
\usepackage{algpseudocode}%
\usepackage{listings}%
\usepackage{stfloats}

\usepackage{cite}

\usepackage[hidelinks]{hyperref}
\usepackage{cleveref}
\crefname{figure}{Fig.}{Figs.}
\Crefname{figure}{Fig.}{Figs.}

\begin{document}

\title{Guiding Multi-Objective Genetic Programming with Description Length Improves Symbolic Regression Solutions}

\author{Gabriel Kronberger, Fabricio Olivetti de Fran\c{c}a, Deaglan J. Bartlett, Harry Desmond, and Pedro G. Ferreira%
\thanks{This work has been submitted to the IEEE for possible publication. Copyright may be transferred without notice, after which this version may no longer be accessible.}%
\thanks{Gabriel Kronberger is with the Heuristic and Evolutionary Algorithms Laboratory, University of Applied Sciences Upper Austria, Softwarepark 11, 4232 Hagenberg, Austria (e-mail: gabriel.kronberger@fh-hagenberg.at).}%
\thanks{Fabricio Olivetti de Fran\c{c}a is with the Center of Mathematics, Computing and Cognition, Federal University of ABC, Av. dos Estados 5001, Santo Andre, 09280-560, SP, Brazil (e-mail: olivetti@ufabc.edu.br).}%
\thanks{Deaglan J. Bartlett and Pedro G. Ferreira are with Astrophysics, University of Oxford, Denys Wilkinson Building, Keble Road, Oxford, OX1 3RH, UK (e-mail: deaglan.bartlett@physics.ox.ac.uk; pedro.ferreira@physics.ox.ac.uk).}%
\thanks{Harry Desmond is with the Institute of Cosmology \& Gravitation, University of Portsmouth, Dennis Sciama Building, Portsmouth, PO1 3FX, UK (e-mail: harry.desmond@port.ac.uk).}%
}
\maketitle
\begin{abstract}
Symbolic regression with genetic programming (GPSR) may suffer from overfitting and structural bloat, especially when noise is present. In this paper we evaluate description length (DL) and fractional Bayes factor (FBF) criteria as principled, data-efficient alternatives to heuristics for selecting compact expressions that generalise well. We implement DL using a Fisher-information--based parameter encoding and compare it to the Akaike Information Criterion (AIC) and the Bayesian Information Criterion (BIC) across multiple datasets, including noisy synthetic benchmarks and real-world regression problems. We study three search/selection strategies: (i) multi-objective search for accuracy and program length followed by DL/FBF selection; (ii) multi-objective search using DL directly as an objective; and (iii) single-objective optimisation with DL/FBF as the fitness. Across datasets, we find that DL/FBF post-selection  improves test performance compared to AIC/BIC baseline and that BIC in combination with the same function complexity penalty from DL/FBF produces similar results. In contrast, using DL/FBF directly as a fitness function in single-objective GPSR frequently induces premature convergence to overly simple models. We conclude with practical guidance for using DL/FBF as robust model-selection tools in genetic programming workflows.
\end{abstract}


\section{Introduction}\label{sec:intro}
\noindent
\IEEEPARstart{S}{ymbolic} Regression (SR) aims to recover the functional relation underlying observed data as a nonlinear regression model~\citep{Koza1992,Kronberger2024}. Formally, given a search space of parametric functions defined by a symbol set (e.g., $+, \times,\sin$) and structural constraints (e.g., maximum size), SR seeks $\hat{f}(x; \theta)$ that best approximates a dataset $\{x_i, y_i\}_{i=1..m}$ under a loss $\mathcal{L}(x, y, f, \theta)$.

A common search method for $\hat{f}$ is genetic programming (GP)~\citep{Koza1992}, a metaheuristic inspired by evolution. GP starts from a randomly generated population of candidate expressions and iteratively selects, recombines, and mutates them according to fitness. Variants of GP rank among the strongest methods in SR benchmarks~\citep{SRBench_LaCava,deFranca2025_srbenchplusplus,yu2025_mdlformer}.

Because observed data usually contain irreducible noise, a model that fits the training data perfectly is often more complex than necessary to represent the underlying data-generating process (overfitting). Model selection is therefore required to balance predictive accuracy and complexity. In GPSR, this also interacts with bloat, i.e., growth in expression size without commensurate accuracy gains. Effective model selection should control both overfitting and bloat. Recent GP approaches reduce overfitting by penalising functional complexity, quantified with an improved slope-based measure, through multi-objective selection~\citep{Silva2026}, or by jointly minimising the empirical risk and the vicinal Jensen gap in evolutionary feature construction~\citep{Zhang2026}.

The minimum description length (MDL) principle provides a model-selection framework grounded in Shannon information and Kolmogorov complexity. It favours the model that yields the shortest description of the data~\citep{Rissanen1978,Grunwald2007} and is closely related to Solomonoff induction~\citep{Solomonoff1964_part1, Solomonoff1964_part2}. \citet{Guimera2026} argue that such probabilistic approaches are a principled alternative to the heuristic choices of loss function, complexity measure, and model selection criterion that are common in SR.
Our main question is whether GPSR performance can be improved by using MDL for model selection or even as a search driver for GP.

A simple MDL variant for GP appeared in early GP literature~\citep{Iba1995Stroganoff}, where
beneficial effects were observed. \citet{Udrescu2020} used MDL with a $k \log n$ function complexity penalty to select models in their SR system AI Feynman 2.0.
More recently, description length (DL) has been used to rank candidate models in exhaustive symbolic regression~\citep{Bartlett_ESR_2023,Desmond2023}; fractional Bayes factors (FBF) have been used as fitness in GPSR to reduce bloat and overfitting~\citep{Bomarito2022,Bomarito2023} and for Monte Carlo sampling of equations in non-evolutionary symbolic regression~\citep{Guimera2020,Bomarito2025}; and MDL has been used to guide search using a transformer-based neural network~\citep{yu2025_mdlformer}. However, a comprehensive study of DL and FBF in GPSR is still missing, which motivates this work.
In related work, \citet{Ramlan2026} tested several model selection criteria including DL for GPSR and found inconsistent performance across several datasets with no clear winner. \citet{Soltani2026} compared similar model selection criteria, including DL for synthetic data and a small number of candidate models, but did not test the effects of using those model selection criteria to guide GPSR.

This work makes four concrete contributions beyond prior studies. First, we provide a controlled empirical comparison of AIC, BIC, DL, and FBF for symbolic regression across multiple datasets and noise regimes. Second, we implement an adaptation to DL as described by \citet{Bartlett_ESR_2023} that encodes parameter precision via the observed Fisher information and a rotation toward approximately independent parameter directions, improving stability in GPSR. Third, we disentangle search effects from selection effects by evaluating three strategy families (multi-objective length-first, multi-objective DL-as-objective, and single-objective DL/FBF fitness), showing where DL/FBF helps and where it can hurt. Fourth, we provide practical recommendations that reduce bloat and overfitting without requiring a separate validation set.

\section{Methods}\label{sec:methods}
\noindent
In all experiments, we assume i.i.d. Gaussian errors with unknown noise variance $\sigma^2_\text{err}$ and use the negative log-likelihood (NLL)
\begin{multline}
  -\log \mathcal{L}(x,y|\hat{\theta},\hat\sigma^2_\text{err}) = \frac{m}{2}\log (2\, \pi\, \sigma^2_\text{err}) \\
  + \frac{1}{2\, \sigma^2_\text{err}} \sum_{i=1}^m \left(y_i - f(x_i, \theta) \right)^2
  \label{eqn:nll}
\end{multline}
for a regression function $f(x, \theta)$ with fitting parameters $\theta$ and $\sigma^2_\text{err}$ and a dataset with $m$ observations $(x_i, y_i)_{i=1\ldots m}$. Throughout this paper, $\log$ refers to natural logarithms. When interpreting description lengths as bits, we divide by $\ln 2$; this does not affect model selection because it rescales all description-length terms by a constant. The regression parameters $\theta$ are optimised via the Levenberg-Marquardt algorithm for nonlinear least squares  \citep{Levenberg_1944,Marquardt_1963}.
We do not assume known noise variance. Instead, for each candidate model we estimate $\sigma^2_\text{err}$ from residuals, as \citet{Guimera2020}, using the model MSE $\left(\frac{1}{m} \sum_{i=1}^m \left(y_i - f(x_i, \theta) \right)^2\right)$ at a given $\theta$. This is equivalent to profiling over $\sigma^2_\text{err}$ and yields the same optimum as joint optimisation over $\sigma^2_\text{err}$ and $\theta$.
This gives the maximum-likelihood estimates (MLEs) for the $p = p_\theta + 1$ parameters $(\hat\theta_1, \ldots, \hat\theta_{p_\theta}, \hat\sigma^2_\text{err})$, i.e., the $p_\theta$ regression parameters plus the noise variance. Throughout this paper, $p$ denotes the total parameter count \emph{including} the noise variance (used in AIC, BIC, BIC$_\text{SR}$, FBF, and DL), whereas $p_\theta = p - 1$ denotes the number of regression parameters only. We collect all fitted parameters in the vector $\hat\vartheta = (\hat\theta_1, \ldots, \hat\theta_{p_\theta}, \hat\sigma^2_\text{err})$. The Fisher information matrix used for the parameter complexity of DL is the observed Fisher information with respect to all $p$ elements of $\hat\vartheta$, i.e., it includes the profiled noise variance. We evaluate NLL at these MLEs for all model-selection criteria.

\subsection{Model selection criteria}
\noindent
All model-selection criteria studied here can be combined with any likelihood, but they cannot be applied directly to non-likelihood losses. This is expected: a likelihood is required to specify the data-generating process. For general losses, one can use a validation set for model selection when sufficient training data are available, but this leaves parameters unconditioned on all observations.

Classic model selection criteria grounded in information theory are Akaike's information criterion (AIC)~\citep{Akaike1974} and the Bayesian information criterion (BIC)~\citep{Schwarz1978}. Both use only the number of parameters $p$ in the complexity penalty and neglect the structural complexity of expressions as well as parameter stiffness.
The AIC is defined to be
\begin{equation}
  \mathrm{AIC} = -2\, \log \mathcal{L}(x,y|\hat{\theta},\hat\sigma^2_\text{err}) + 2\, p,
  \label{eqn:AIC}
\end{equation}
and can compare parametric models when at least one model can approximate the true data-generating process in the limit $m \rightarrow \infty$. It can also be used for regularised nonparametric model families when an effective number of parameters is defined~\citep{Hastie2001}.

BIC uses a stronger penalty than AIC by replacing the factor 2 with $\log m$. Adding one parameter increases the penalty by $\log m$, so BIC improves only if NLL decreases accordingly across the dataset\footnote{For i.i.d Gaussian errors with variance $\sigma_\text{err}^2$, adding one parameter is matched when the sum of squared residuals decreases by $\sigma_\text{err}^{-2}\, \log m$.}.
\begin{equation}
  \mathrm{BIC} = -2\, \log \mathcal{L}(x,y|\hat{\theta},\hat\sigma^2_\text{err}) + p\, \log m,
  \label{eqn:bic}
\end{equation}
However, the BIC penalty still depends only on parameter count. Any NLL reduction is preferred when $p$ is unchanged. This is problematic in GPSR because it can increase structural complexity (e.g., by adding unary functions or input variables) without introducing new fitted parameters. BIC cannot detect this form of overfitting.

For Bayesian model selection for SR, we may consider a hierarchical model with priors over candidate models $\{f_1, f_2, \ldots, f_M\}$ and their parameters $\{\ \boldsymbol{\theta}_1, \boldsymbol{\theta}_2, \ldots, \boldsymbol{\theta}_M \}$ and calculate the probability of one of the models under data using Bayes' theorem
\begin{multline}
  P(f_i | D) = \frac{1}{P(D)} \int P(D | f_i, \theta_i) P(\theta_i | f_i) P(f_i) d\theta_i  \\
  \equiv \frac{P(f_i)}{P(D)} \mathcal{Z}(D|f_i),
\end{multline}
where $P(D)$ is the probability of the data, $P(f_i)$ is the prior probability of model $f_i$, $P(\theta_i | f_i)$ is the prior over parameters of model $f_i$, and $\mathcal{Z}(D|f_i)$ is the Bayesian evidence for model $f_i$. The model with the highest posterior probability is selected. When comparing multiple models on the same data, we may ignore the constant $P(D)$ and select the model with minimal
\begin{equation}
  -\log P(f_i|D) = -\log P(f_i) - \log \mathcal{Z}(D|f_i) .
\end{equation}

The evidence is often approximated using the Laplace approximation about the maximum a posteriori point $\hat{\mathbf{\theta}}_i$, which gives
\begin{multline}
  \log \mathcal{Z}(D|f_i) \approx \log P(D | f_i, \hat{\mathbf{\theta}}_i) + \log P(\hat{\mathbf{\theta}}_i | f_i) \\
  + \frac{p}{2} \log (2 \pi) - \frac{1}{2} \log |\mathrm{FIM}(\hat{\mathbf{\theta}}_i)|
\end{multline}
where $|\mathrm{FIM}(\hat{\mathbf{\theta}}_i)|$ is the determinant of the observed Fisher information matrix evaluated at the maximum likelihood estimate (MLE) for the parameters $\hat{\mathbf{\theta}}_i$ for model $f_i$.
BIC is a large-sample approximation to Bayesian evidence under the Laplace approximation. In the limit $m \rightarrow \infty$, the likelihood term dominates the evidence, the prior over parameters can be neglected, the final term can be approximated by $-p/2 \log m$, and the term proportional to $p$ is dropped. This gives $\log \mathcal{Z}(D|f_i) \approx \log P(D | f_i, \hat{\mathbf{\theta}}_i) - \frac{p}{2} \log m$ which is equivalent to $1/2$ BIC.

DL and FBF as defined by \citet{Bartlett_ESR_2023} are other approximations to the Bayesian evidence under particular choices of priors over models and parameters, which we discuss in more detail below.

The minimal description length (MDL) principle~\citep{Rissanen1978,Grunwald2007} prefers models that yield the shortest compressed representation of the data. The description length is $DL = L(D|M) + L(M)$, i.e., data codelength under the model plus model codelength. More compact data encoding and shorter models are preferred.

\citet{Udrescu2020} included a structural complexity penalty into the DL criterion used for AI Feynman 2.0 to account for expression structure and constants.
This is extended by \citet{Bartlett_ESR_2023} to include codelength for fitted parameter values. Intuitively, parameter complexity is determined by required precision: stiff parameters that must lie close to their maximum likelihood values to maintain high likelihood are complex and require a longer codelength. Sloppy parameters are model parameters that can be changed with negligible effect on the predictions of the model and require a shorter codelength. 

This formulation of DL has been used to rank expressions for exhaustive symbolic regression.
In GPSR, crossover can produce unnecessarily complex and overparameterised expressions~\citep{Kronberger2025} which leads to highly correlated parameter estimates. \citet{Bartlett_ESR_2023} used only the diagonal of the FIM which cannot detect highly correlated or redundant parameters. We therefore modify the parameter complexity term of DL and use the full Fisher information matrix (FIM) to evaluate parameter codelength in a rotated rectangular lattice calculated via singular value decomposition (SVD). This allows automatic detection of highly correlated parameters and reduced codelengths. The eigenvectors of the FIM represent orthogonal directions in a rotated parameter space, where vectors with large eigenvalues represent \emph{stiff directions} and vectors with small eigenvalues represent \emph{sloppy} directions~\citep{Quinn2022}.

\begin{figure}
  \includegraphics[]{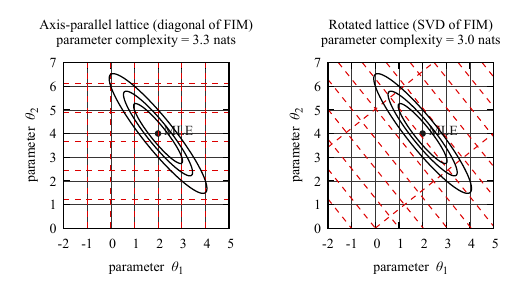}
  \caption{Visualisation of axis-parallel and rotated lattices used for calculating the parameter complexity based on the Fisher information matrix.}\label{fig:lattice-rotation}
\end{figure}

The concept is visualised in \Cref{fig:lattice-rotation} for an estimated parameter vector $\theta=(2,4)$ with high correlation between $\theta_1$ and $\theta_2$. For the axis-parallel rectangular lattice, the precision for both parameters is similar, and the MLE is mapped to the lattice values $(2.0,3.3)$ with parameter complexity $\approx 3.3$ nats. For the rotated lattice, the parameter vector is mapped to lattice value $(5.1, 0.5)$ with complexity $\approx 3.0$ nats; one parameter direction is stiff and has to be communicated precisely, while  the second direction is sloppy and costs only $\log 2$ for its sign.

Under this scheme, the DL can be written as
\begin{multline}
  \mathrm{DL} = -\log \mathcal{L}(x,y|\hat{\theta},\hat\sigma^2_\text{err}) + \underbrace{k \log n + \sum_{c_i} \log (|c_i| + 1)}_{\text{function complexity}} \\
  + \underbrace{p \log 2 +{} \sum_{i=1}^{p} \mathrm{max} \left(0, \tfrac{1}{2}\log |\Sigma_{ii}| + \log |\hat{\vartheta}^\mathrm{rot}_i| - \tfrac{1}{2}\log 12  \right)}_{\text{parameter complexity}}
  \label{eqn:dl}
\end{multline}
where $k$ is the number of expression-tree nodes that are operators, functions, or variables (nodes for numerical constants and fitted parameters are not counted, see below); $n$ is the number of distinct symbols among these nodes (each variable $x_1, x_2, \ldots$ is counted separately); $c_i$ are the numerical constants in the expression; $p$ is the parameter count including the noise variance; $\mathrm{FIM}(\hat{\vartheta}) = U\Sigma V^\top$ is the factorisation from SVD of the Fisher information matrix of all $p$ parameters (including the profiled $\sigma^2_\text{err}$); and $\hat{\vartheta}^\mathrm{rot} = V^\top \hat{\vartheta}$. Each rotated parameter is encoded by its sign ($\log 2$) and the magnitude $|\hat{\vartheta}^\mathrm{rot}_i|$ on a lattice with spacing $\sqrt{12/\Sigma_{ii}}$. The magnitude term is clamped at zero, so a sloppy direction whose value is smaller than the lattice spacing costs only $\log 2$ for its sign\footnote{This variant of parameter complexity requires the full FIM, which costs $O(m\,p^2)$ for an expression with $O(p)$ nodes. The SVD additionally costs $O(p^3)$, but it does not depend on the number of observations $m$. In our experiments, the number of parameters of each evaluated model is indirectly limited (up to $\approx 50$) through the expression size limit, and the SVD takes a median 3~\% of the runtime of MO-DL, which calculates the DL of every evaluated expression (up to 27~\% for the smallest dataset, see \Cref{sec:appendix-runtime-profile} of the supplementary material).}.

The function complexity term $k \log n + \sum_{c_i} \log (|c_i| + 1)$ represents the codelength for the node sequence of the expression tree, i.e., a string of length $k$ over $n$ distinct symbols (each node costs $\log n$ nats), plus the codelength $\log (|c_i| + 1)$ of each integer constant $c_i$ (the offset of one keeps the codelength well-defined and non-negative for $c_i = 0$; constants are encoded separately and are therefore not part of $k$ and $n$; fitted parameters are accounted for only in the parameter complexity term) and therefore accounts for the structural complexity of expressions \citep{Udrescu2020,Bartlett_ESR_2023}. For example, the expression $f(x) = m \times v \times v / 2$ has length $k=6$ and $n=4$ distinct symbols ($\times, v, m, /$) \citep{Udrescu2020}. Together with the $\log(2+1)$ term for the constant $2$, the codelength of this expression is $6 \log 4 + \log 3 \approx 9.4$ nats\footnote{Note that unlike in~\citet{Bartlett_Priors_2023,Bartlett_ESR_2023}, we do \emph{not} count constant or parameter symbols in the $k \log n$ term. Thus, for example, $f(x)=\theta_1 \times (1 - \exp(\theta_2 \times x))$ has $k = 5 \: (\times, -, \exp, \times, x)$, $n = 4 \: (\times, -, \exp, x)$, and function complexity $5\log 4 + \log(1+1) = 7.6$ nats (variables are counted as symbols, the constant $1$ contributes $\log 2$).}.

As discussed by \citet{Bartlett_Priors_2023}, DL can be interpreted as the negative log of Bayesian model probability under a specific prior choice.
The parameter-complexity term corresponds to a particular choice of parameter prior which favours small values.

As an alternative, fractional Bayes factors (FBF) have been proposed for Bayesian model selection in combination with vague or improper parameter priors. For SR, FBF for model selection have been discussed by \citet{Bartlett_Priors_2023} and \citet{Bomarito2025}, and FBF have been used as fitness in GPSR~\citep{Bomarito2022,Bomarito2023}.
Under FBF, a fraction $b$ of the data is used as a calibration sample to derive a parameter prior, and model comparison is performed on the remaining data. As described by \citet{Bartlett_Priors_2023}, this gives an approximation of the Bayesian evidence of $p/2 \log b - (1-b)\mathrm{NLL}$. Adding the negative log prior on function structure yields the FBF loss \citep[Eq. 7, with $\log\nu_p = 1 - \log3$ corresponding to a rectangular lattice]{Bartlett_Priors_2023}:
\begin{multline}
  \mathrm{FBF} = -(1-b)\, \log \mathcal{L}(x,y|\hat{\theta},\hat\sigma^2_\text{err}) + \text{func}_\text{compl} \\
  +\frac{p}{2} (\log(2\,\pi\, e^{1-\log 3}) - \log b),
  \label{eqn:fbf}
\end{multline}
where $\text{func}_\text{compl} \equiv k \log n + \sum_{c_i} \log (|c_i| + 1)$ and $p$ includes the noise variance as above, i.e., FBF and BIC$_\text{SR}$ count the same parameters as DL.
Following \citet{OHagan1995}, we choose $b=m^{-1/2}$, where $m$ is the number of observations.
This FBF formulation is based on DL~(\Cref{eqn:dl}) but does not require the improper prior induced by integer parameter discretisation. The remaining structural prior, $\text{func}_\text{compl}$, is the same compression-motivated prior used in DL.

Similarly, the BIC criterion given above can simply be extended with the same function complexity term  for GPSR:
\begin{equation}
  \mathrm{BIC}_\mathrm{SR} = -2\, \log \mathcal{L}(x,y|\hat{\theta},\hat\sigma^2_\text{err})  + 2\, \text{func}_\text{compl} + p \log m,
  \label{eqn:BICSR}
\end{equation}
where $m$ is the number of observations (not to be confused with the number $n$ of distinct symbols in $\text{func}_\text{compl}$), providing three approximations for $\mathcal{Z}(D|f_i)$ which differ entirely in the parameter complexity penalty.

In the large sample limit, all three criteria converge to the same model selection result, but they differ for smaller dataset sizes. In particular, the parameter complexity term of DL is expected to be more accurate than the $p \log m$ term of BIC$_\text{SR}$, because it accounts for parameter stiffness and correlations. Through this mechanism, DL can distinguish two models with the same structural complexity and the same number of parameters based on parameter stiffness. Models which are more sensitive to parameter variations, have higher parameter complexity, and thus are penalised more heavily. FBF lies between BIC$_\text{SR}$ and DL, and uses a different, dataset-size dependent, trade-off between NLL and number of parameters than BIC$_\text{SR}$, but does not account for parameter stiffness.

In our synthetic experiments, the data-generating process of each ground-truth function is fixed (fixed parameters and i.i.d.\ noise), and GPSR is run on 100 datasets sampled from it, each split into training and test folds. This setup does not assess whether the structural or parameter priors implicit in DL, FBF, or BIC match any generative prior over functions; rather, it probes their finite-sample predictive behaviour conditional on a fixed truth. Concretely, each metric induces a different effective regularisation strength through its complexity penalty, and the experiment measures how well that penalty approximates the oracle bias-variance trade-off for the given function, noise level, and sample size. The test error therefore serves as a frequentist estimate of expected generalisation risk under repeated sampling, allowing us to compare the selection criteria as regularisation rules applied to the same candidate models discovered by the SR search.

\subsection{Algorithms}
\noindent
We compare three tree-based GP variants with model selection by AIC, BIC, BIC$_\text{SR}$, DL, and FBF:
\textbf{MO-Len:} multi-objective GP with NLL and program length as objectives. Here, the selection criteria are used only to select a single model from the Pareto front in each generation.
\textbf{MO-DL:} multi-objective GP using NLL and the DL complexity penalty as objectives and returning the individual with best DL from the final Pareto front.
\textbf{SO:} single-objective GP using AIC, BIC, BIC$_\text{SR}$, DL, or FBF directly as fitness and returning the best-fitness solution.
\Cref{tab:methods-table} gives an overview of the multi-objective algorithm variants, which we explain in more detail in the following.
\begin{table}[t!]
\centering

\caption{Overview of tested multi-objective GP variants.}\label{tab:methods-table}
\begin{tabular}{llll}
  Identifier & Obj. 1 & Obj. 2 & Selection \\
  \hline
  MO-Len-AIC               & NLL   & length & AIC (\cref{eqn:AIC}) \\
  MO-Len-BIC               & NLL   & length & BIC (\cref{eqn:bic}) \\
  MO-Len-DL                & NLL   & length & DL (\cref{eqn:dl}) \\
  MO-Len-FBF               & NLL   & length & FBF (\cref{eqn:fbf}) \\
  MO-Len-BIC$_\text{SR}$   & NLL   & length & BIC$_\text{SR}$ (\cref{eqn:BICSR}) \\
  MO-DL                       & NLL   & $\text{func}_\text{compl} +\text{param}_\text{compl}$ & DL (\cref{eqn:dl}) \\
\end{tabular}

\end{table}

\subsubsection{Multi-objective GPSR (MO-Len)}
For the first experiment, we use a multi-objective (MO) algorithm that optimises NLL and program length via non-dominated sorting. This approach is implemented, for example, in pyoperon~\citep{Burlacu2020operon}. Our implementation is based on NSGA-II~\citep{Deb2002} with minor adaptations and is shown in \Cref{alg:multi-objective} (in the appendix \ref{sec:appendix-algorithms}). The adaptations, namely the handling of expressions with identical objective values (duplicates) and the rounding of the NLL to two decimal places, are described in \Cref{sec:appendix-population-update} of the supplementary material. Hyperparameters that are set by the user are: population size (popSize), the maximum number of generations (maxGen), the maximum program length (maxLen), the crossover rate, and the sets of function symbols $\mathcal{F}$ and terminal symbols  $\mathcal{T}$. The hyperparameter values used for all algorithm variants are listed in \Cref{tab:common-hyperparameters}. All algorithm variants evaluate popSize expressions in each generation and therefore the same number of expressions in a run, and the parameters of each evaluated expression are fitted with at most 50 residual evaluations (\Cref{sec:appendix-numerical} of the supplementary material).

Fitting parameters of solution candidates are optimised whenever a solution candidate is evaluated as shown in \Cref{alg:eval} (in the appendix \ref{sec:appendix-algorithms}), whereby the starting values are extracted from the individual and the individual is updated with the optimised values afterwards~\citep{Kommenda2019} (memetic optimisation of parameters). The same parameter optimisation procedure is used in all algorithm variants. The second objective in MO-Len is program length, which is the number of symbols in the expression including operators, variables, and fitting parameters or equivalently the number of nodes of the corresponding expression tree.

In NSGA-II, the population is partitioned into Pareto fronts by non-dominated sorting. Tournament selection returns the candidate with the smallest front rank and breaks ties by crowding distance, which prefers candidates in sparsely populated regions of the objective space~\citep{Deb2002}. In each generation, offspring are created by tournament selection, subtree crossover, and mutation, and the merged population and offspring are re-ranked and truncated to the population size (elitist generational replacement).

Multi-objective GPSR returns a Pareto set of expressions from which a user can select a final model, either manually (e.g., by structural simplicity or asymptotic behaviour) or automatically via one of the model selection criteria. Because MO-Len always optimises NLL and program length, we can compare multiple selection criteria based on the same runs. Because the criteria are computed only for the first Pareto front, the additional DL cost (full FIM + SVD) is small relative to parameter fitting, which is required for all candidates and dominates runtime (DL selection takes about 3~\% and parameter fitting 94~\% of the runtime of MO-Len-DL).

\subsubsection{Multi-objective optimisation of DL (MO-DL)}
In the second experiment, we use the same algorithm (\Cref{alg:multi-objective}) but replace program length with the DL penalty (the sum of the function complexity and the parameter complexity) as the second objective. All tested criteria are weighted sums of NLL and a penalty term, so MO-DL minimises DL directly, which may improve the selected DL values and test errors relative to MO-Len. As before, the final model is selected from the Pareto front by minimum DL.

\subsubsection{Single-objective GPSR (SO)}
In the single-objective approach, we assign AIC, BIC, BIC$_\text{SR}$, DL, and FBF as fitness values of candidate solutions and minimise them directly.
The algorithm shown in \Cref{alg:basic} (in the appendix \ref{sec:appendix-algorithms}) is close to a Koza-style GP with generational replacement and a single elite individual. We use the same operators as for the multi-objective algorithm. The only change is that fitness for tournament selection is based on the model selection criterion.

\begin{table}
\centering
\caption{Common hyperparameter values for tree-based GPSR. For the datasets from the \emph{Engineering} group, we allowed larger length limits and more generations because initial runs with the length limit 100 showed no signs of overfitting.
}\label{tab:common-hyperparameters}
\small
\begin{tabular}{ll}
  \hline
  Parameter & Value \\
  \hline
Population size & 1000 \\
Selection & Tournament (group size 2) \\
Max. generations & 200 (Synthetic, SRbench, Nikuradse) \\
                 & 200 (Chem. comp.) \\
                 & 400 (Friction, Tower) \\
Crossover prob. & 90\% \\
Mutation prob. & 10\% of indiv. and 5\% for each node \\
Initialisation & Grow with depth limit 5 \\
Max. length   & 100 (Synthetic, SRbench) \\
                   & 200 (Nikuradse, Friction) \\
                   & 300 (Tower, Chem. comp.) \\
Function set & $\bigl\{+, -, \times, \div, x^2, \sin, \cos, \exp, \log |x|, $ \\
             & $\sqrt{|x|}, |x|^y \bigr\}$ \\
Terminal set  & $\{\text{var}, \text{param}, \text{var} \times \text{param}\}$ \\
\hline
\end{tabular}
\end{table}

\subsection{Datasets}
\noindent
We test on three groups of datasets shown in \Cref{tab:datasets}, (i) synthetic datasets with known ground truth, (ii) a subset of datasets from SRBench, a community-driven benchmark for symbolic regression~\citep{SRBench_LaCava}, which was recommended by~\citet{SRBench_Aldeia}, and (iii) a set of real-world datasets from~\citet{Kronberger2024}. The generating functions, the sampling of the inputs for the training and test sets, and the noise levels of the synthetic datasets are given in \Cref{tab:synthetic-datasets} in the supplementary material.
For synthetic datasets, the likelihood is consistent with the data-generating process and they allow controlled tests of noise level and sample-size effects. This is not necessarily the case for SRBench and real-world datasets, where the true data-generating process is unknown and may not be well-described by a Gaussian likelihood. We nevertheless use Gaussian likelihoods for the real-world datasets, which is common in GPSR, and evaluate performance on held-out test data which enables comparison to prior work. If the true data-generating process is not Gaussian, the likelihood is misspecified and the model selection criteria would no longer select models with optimal accuracy/complexity trade-off and would be expected to perform worse on the test data. We discuss the implications of these assumptions in \Cref{sec:limitations}. 

For the engineering datasets, the training and testing sets are predefined. The two \emph{Chemical} datasets both stem from industrial processes. The \emph{Chemical Tower} dataset has been used for modelling gas chromatography measurements of the composition of a distillation tower \citep{Vladislavleva2009}. It has 25 input variables which are temperatures, flows, and pressures related to the distillation tower and a single target, which is the propylene concentration at the top of the distillation tower. The dataset published by \citet{Vladislavleva2009} seems to be preprocessed to assign similar time-series fragments to training and testing sets. The \emph{Chemical Competition} dataset is a multivariate time-series with 57 input variables and a single target.

The two \emph{Nikuradse} datasets were produced by  \citet{nikuradse1950} by lab measurements of the friction of fluids in rough pipes. We use both datasets extracted from the tables given in the original paper: \emph{Nikuradse (1d)} contains the data for Prandtl's collapse where the goal is to find  the relationship between turbulent friction $\lambda$ as a function of scaled roughness; \emph{Nikuradse (2d)} is the dataset with two controlled input variables: Reynolds number ($\log$ Re) and the roughness factor ($r/k$), as well as the turbulent friction as the target \citet{nikuradse1950}. The dataset has been studied by \citet{Reichardt2020,Guimera2020} and \citet{Reichardt2020} highlighted its importance as the functional dependence of friction on the Reynolds number and relative roughness is still unknown.

The friction datasets stem from performance testing of industrially produced friction parts whereby the categorical variable for friction material types has been one-hot encoded~\citep{Kronberger2024}. The goal is to predict the friction coefficient for each material from the input variables pressure, temperature, and sliding velocity for dynamic friction, or only pressure and temperature for static friction.

For the SRBench datasets, we generated 100 random splits into training ($2/3$) and testing ($1/3$) sets, except for the large datasets \emph{1193 BNG lowbwt} and \emph{1199 BNG echo months}, for which we randomly sampled 100 datasets each with 1000 observations for training and used the remaining rows for testing.

\begin{table}
  \centering
 \small
 \caption{Datasets used in this study. Number of observations in training and testing sets are given as $(m_\text{train}, m_\text{test})$. }
 \label{tab:datasets}
 \begin{tabular}{l|rrr}
   Identifier & \textbf{Observations} & \textbf{Vars.} & \textbf{Ref.} \\
   \hline
   \textbf{Synthetic} & & & \\
   Friedman-1 (additive) & (50 -- 1000, 1000) & 10 &\citep{FriedmanMARS} \\
   Friedman-2 (interactions) & (100, 1000) & 10 & \citep{FriedmanMARS} \\
   Kotanchek & (100, 2025) & 2 & \citep{Vladislavleva2009} \\
   RatPol2d & (50, 1156) & 2 & \citep{Vladislavleva2009} \\
   RatPol3d & (300, 2700) & 3 & \citep{Vladislavleva2009} \\
   Ripple & (300, 1000) & 2 & \citep{Vladislavleva2009} \\
   Salustowicz & (100, 220) & 1 & \citep{Salustowicz1997} \\
   Salustowicz2d & (606, 2553) & 2 & \citep{Vladislavleva2009} \\
   \hline
   \textbf{Engineering} & & & \\
   Nikuradse (1d) & $(200, 162)$ & $1$  & \citep{nikuradse1950,Guimera2020} \\
   Nikuradse (2d) & $(230, 132)$ & $2$  & \citep{nikuradse1950,Guimera2020} \\
   Friction dyn. & $(1009, 1007)$ & $17$ & \citep{Kronberger2018} \\
   Friction stat. &  $(1009, 1007)$ & $16$ & \citep{Kronberger2018} \\
   Chemical - Tower & $(3136,  1863)$ & $25$ & \citep{Vladislavleva2009} \\
   Chemical - Competition & $(711, 355)$ & $57$ & \citep{Kordon2008} \\
   \hline
   \textbf{SRBench selection} &  & & \\
   192 Vineyard & $(31,21)$ & $2$ & \citep{vineyard,olson2017pmlb} \\
   210 Cloud & $(64,44)$ & $5$  & \citep{olson2017pmlb} \\
   522 PM10 &  $(300,200)$ & $7$ & \citep{olson2017pmlb}  \\
   557 Analcat Apnea1 &  $(285,190)$ & $3$ & \citep{olson2017pmlb,simonoff2003analyzing} \\
   579 Fri. c0 250 5 & $(150,100)$ & $5$ & \citep{olson2017pmlb,friedman2002stochastic} \\
   606 Fri. c2 1000 10 & $(600,400)$ & $10$ & \citep{olson2017pmlb,friedman2002stochastic} \\
   650 Fri. c0 500 50 & $(300,200)$ & $50$ &  \citep{olson2017pmlb,friedman2002stochastic} \\
   678 Vis. env & $(66,45)$ & $3$ & \citep{olson2017pmlb,cleveland1993visualizing} \\
   1028 SWD & $(600,400)$ & $10$ & \citep{olson2017pmlb}  \\
   1089 USCrime & $(28,19)$ & $13$ & \citep{olson2017pmlb}  \\
   1193 BNG lowbwt & $(1000,10368)$ & $9$ & \citep{olson2017pmlb,kilpatrick1998numeric} \\
   1199 BNG echo months & $(1000,5832)$ & $9$ & \citep{olson2017pmlb,kilpatrick1998numeric} \\
   \hline
 \end{tabular}
  \end{table}

\subsection{Statistical comparisons}\label{sec:statistical-comparisons}
\noindent
Instead of null-hypothesis significance testing, we use the Bayesian Bradley-Terry (BBT) model for the comparison of multiple methods over multiple datasets as recommended by~\citet{Wainer2023}. The BBT model estimates the capability of each method (``team'') from observed results of competitions between pairs.
The result of a match is either a win for team A, a tie, or a win for team B. \citet{Wainer2023} recommends BBT over the Bayesian Wilcoxon signed-rank test proposed by ~\citet{Benavoli2014,Benavoli2017}, as it can be used for accuracy metrics that are not comparable over datasets (e.g., RMSE, DL), and does not require to set a global region of practical equivalence (ROPE).

We compare the test RMSE and DL of the expressions selected by each possible selection-criterion pairing. In each run, the method which selects the better value wins. If both methods select the same expression it is a tie, which is counted as 1/2 win for both methods. \Cref{tab:wins} and \Cref{tab:ties} in the supplementary material list the win and tie percentages, respectively, for test RMSE over all datasets for all pairings.
All compared methods are evaluated on the same 100 runs (training sets and random seeds) per dataset, so the comparison is paired. However, the runs are not independent: the 100 runs on a dataset share the generating function (synthetic datasets), overlapping training sets of random splits of the same data (SRBench datasets), or the same training set (engineering datasets), and the criteria compared for MO-Len select from the same Pareto front. We therefore use the 34 datasets as the independent units and count one match per dataset for each pair of methods. Method A wins the match if it is better than method B in more than 55~\% of the runs on the dataset (ties counted as 1/2), B wins if A is better in less than 45~\% of the runs, and otherwise the match is a tie. This is the same band as the ROPE for $P(A > B)$ defined below. \Cref{tab:bbt-sensitivity} in the supplementary material shows that the results are not sensitive to this band, and that the local ROPE of~\citet{Wainer2023} for the log test RMSE leads to similar results.
In the BBT model, a single parameter for each criterion expresses the capability of the criterion. The result is a posterior distribution for the capabilities which can be used to rank the criteria. The probability  that method A wins over method $B$ $P(A > B)$ can then be determined. As suggested by \citet{Wainer2023}, we define the region of practical equivalence (ROPE) for $P(A > B)$ as $45~\%$ to $55~\%$. We report the median value for $P(A>B)$
and mark  $\dagger$ results with more than 99~\% probability mass outside of ROPE .

\section{Main results}\label{sec:results}
\noindent
\label{sec:results-multi-objective}

To illustrate qualitative differences between the selection criteria, we first show one representative synthetic example (line charts for all datasets in \Cref{sec:appendix-multi-objective-linecharts}) and then summarise the results for all datasets.

\begin{figure*}[!t]
  \centering
  \includegraphics{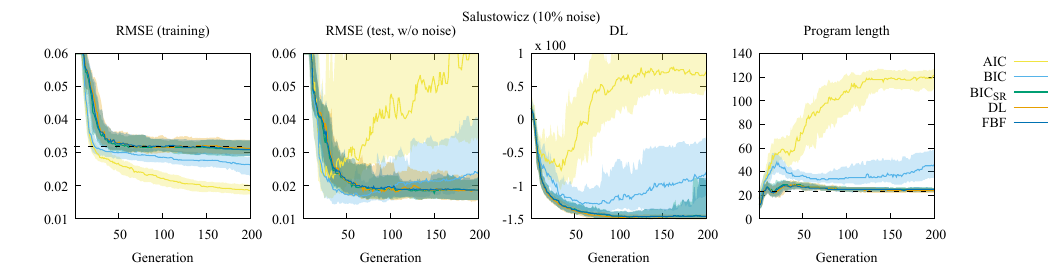}
  \caption{Comparison of selected model metrics for MO-Len on the Salustowicz dataset. The panels show RMSE on the training and test sets, description length, and program length for the five model selection criteria (AIC, BIC, BIC$_\text{SR}$, DL, and FBF). Across 100 MO-Len runs, the different model selection criteria are used to select a single model from the Pareto front in each generation. The lines show the median values, and the shaded areas show the 25\% and 75\% percentiles over generations. The dashed line in the first panel shows $\sigma_\text{err}$, in the last panel the length of the generating expression.
  }
  \label{fig1}
\end{figure*}

\Cref{fig1} shows results from 100 MO-Len runs on the Salustowicz dataset \citep{Salustowicz1997}, a synthetic benchmark that has long been used to test GPSR implementations. The generating expression $f(x)=x^3\, e^{-x}\, \cos(x)\, \sin(x)\, (\sin^2(x)\, \cos(x) - 1)$ for this problem defines a univariate nonlinear function for which it is necessary to find a complex fitting expression; the training set consists of 100 $x$ values on a regular grid in the range [$0.05$, $9.95$], the test set spans a larger range of 220 points [$-0.5$, $10.45$]. We use 10\% noise on the training set and sample 100 random noise realisations. The test sample is generated without noise.

Several patterns are clear. First, FBF, DL, and BIC$_\text{SR}$ selection produce similar results.
The same trend appears on many datasets (\Cref{sec:appendix-multi-objective-linecharts}). 
The first panel demonstrates that the training RMSE of models selected by DL,BIC$_\text{SR}$, and FBF converges to the noise standard deviation $\sigma_\text{err}$, which in this example is known and shown as a dashed line. AIC starts to select overfit models  after 10 to 20 generations because its penalty is too weak and tends to favour the lowest training NLL. BIC starts to overfit later, after about 100 generations.

The third panel shows DL and the fourth program length. Although  FBF ignores parameter stiffness, it still selects models close to minimum DL.

\subsection{Summary of aggregated results}

\noindent
\begin{table}

\caption{Table of aggregated results compared to MO-Len-DL as the reference sorted by relative length difference within each group. The difference between MO-Len with DL, FBF, and BIC$_\text{SR}$ selection is small. SO with selection criteria including the function complexity penalty underfits. SO and MO with AIC or plain BIC overfits. MO-DL has similar test error and runtime . The runtime differences are averaged over the subset of 19 datasets in \Cref{tab:median-runtime-all-datasets-relative}.}\label{tab:result-summary}
\begin{tabular}{l|r|r|r}
Algorithm & Avg. $\Delta$ len. \% & Wins \% & Avg. $\Delta$ time \% \\
\hline
SO-BIC$_\text{SR}$             & $-15.3$ & $18.7^\dagger$ & $-72$ \\
SO-FBF                         & $-15.2$ & $16.4^\dagger$ & $-71$ \\
SO-DL                          & $-1.8$ & $21.7^\dagger$ & $-51$ \\
\hline
\textbf{MO-Len-DL (Ref.)}      & $0.0$ & - & $  0$ \\
MO-Len-BIC$_\text{SR}$         & $-1.5$ & $46.9$ & $ -3$ \\
MO-Len-FBF                     & $-3.1$ & $49.1$ & $ -3$ \\
\hline
MO-DL                          & $22.0$ & $49.9$ & $ -2$ \\
SO-BIC                         & $153.8$ & $38.6$ & $-29$ \\
MO-Len-BIC                     & $340.3$ & $19.8^\dagger$ & $ -3$ \\
SO-AIC                         & $498.7$ & $14.8^\dagger$ & $ 18$ \\
MO-Len-AIC                     & $664.3$ & $5.8^\dagger$ & $ -3$ \\
\end{tabular}

\end{table}

\Cref{tab:result-summary} shows the summary of aggregated results of all algorithm variants compared to MO-Len-DL. The win rate is based on test RMSE. Length is based on the relative differences of median program lengths over 100 repetitions for each dataset. Similarly, median runtimes\footnote{Apple M2 Pro, 32GB RAM, macOS Sequoia, Julia v1.13, single-threaded runs for the runtime measurements (10 runs per dataset and method, 100 generations). The average length differences exclude the dataset 1089 USCrime (smallest training set) on which MO-Len-BIC$_\text{SR}$ and MO-Len-FBF select much longer, overfit expressions.} are calculated for each dataset and then we average the relative differences to the reference algorithm.

Single-objective GP combined with the selection criteria which include the function complexity penalty (BIC$_\text{SR}$, FBF, and DL) leads to underfitting with shorter expressions and worse test RMSE.
MO-Len with DL, FBF, and BIC$_\text{SR}$ are the best, select similar expressions, and have similar runtimes. The additional cost of the DL calculation for the expressions in the Pareto-front is ($\approx 3~\%$).
AIC and plain BIC select longer expressions leading to overfitting and worse test RMSE for single-objective and multi-objective GP.

Runtime is determined mainly by parameter fitting, whose cost grows with the length of expressions in the population. 
MO-DL calculates DL for every evaluated individual. The FIM and SVD take about 38~\% of its runtime, but the expressions in the MO-DL population are on average shorter and have fewer parameters than in MO-Len-DL (average length 31.5 vs.\ 44.3), so that parameter fitting takes only about half as long. This does not contradict \Cref{tab:result-summary}, which compares only the final expressions selected by DL. On average, MO-DL is therefore as fast as MO-Len-DL ($-2~\%$), with large differences between datasets ($-52~\%$ to $+55~\%$, runtime profile in \Cref{sec:appendix-runtime-profile} of the supplementary material).
SO-DL also requires evaluation of the FIM in all fitness evaluations, but the algorithm produces much smaller expressions on average reducing overall costs.

The criteria that include the function complexity penalty (BIC$_\text{SR}$, DL, FBF) select similar expressions with the smallest program lengths, and BIC selects longer expressions (boxplots for all datasets in \Cref{fig:boxplot-size} in the supplementary material).

The median test $R^2$ for MO-Len under each selection criterion is given for all datasets in \Cref{tab:MO-Len-r2} in the supplementary material. DL has the best median test $R^2$ for 17 of the 34 datasets, BIC$_\text{SR}$ and FBF for 13 each, BIC for 5, and AIC for 1 (ties counted for each criterion). We find no difference between DL, BIC$_\text{SR}$, and FBF  ($P(\mathrm{DL}>\mathrm{FBF}) = 50.9~\%$, $P(\mathrm{DL} > \mathrm{BIC}_\text{SR}) = 53.1~\%$; 17 and 19 of the 34 matches are ties). With 34 datasets, the posterior is too wide to establish practical equivalence (60~\% and 53~\% of the probability mass in the ROPE). BIC and AIC are worse ($P(\mathrm{DL} > \mathrm{BIC}) = 80.2~\%^\dagger$, $P(\mathrm{DL} > \mathrm{AIC}) = 94.2~\%^\dagger$).

An ablation of the parameter complexity in \Cref{sec:appendix-ablation} of the supplementary material shows that the axis-parallel lattice of \citet{Bartlett_ESR_2023} (diagonal of the FIM), the lattice based on the determinant of the FIM~\citep{Bartlett_Priors_2023}, and DL without parameter complexity also lead to similar test RMSE ($P(\mathrm{DL}>\mathrm{DL}_\text{diag}) = 55.5~\%$, $P(\mathrm{DL}>\mathrm{DL}_\text{det}) = 48.5~\%$, $P(\mathrm{DL}>\text{NLL + func.\ compl.}) = 37.8~\%$). Without parameter complexity, the selected expressions are 25~\% longer on average, and the test RMSE is slightly lower for many datasets but much higher for some datasets on which overfit expressions are selected.

Multi-objective direct optimisation of DL (MO-DL) selects longer expressions than MO-Len-DL (\Cref{tab:result-summary}) with a smaller DL ($P(\text{MO-DL}>\text{MO-Len-DL})=86.2~\%^\dagger$, \Cref{fig:boxplot-dl-mo-vs-modl} in \Cref{sec:appendix-mo-dl}).
Even though DL is lower, the test RMSE under MO-DL is similar  to that under MO-Len-DL selection ($P(\text{MO-DL}>\text{MO-Len-DL})=49.9~\%$ HPD: $34.4~\% - 65.8~\%$, compare \Cref{fig:boxplot-rmse-mo-vs-modl} in \Cref{sec:appendix-mo-dl}). Intuitively, one might expect that the direct optimisation of DL would yield better test RMSE, but this is not the case here. Detailed analysis of the MO-DL and MO-Len-DL runs shows that the lower DL comes from expressions with a lower parameter complexity (median difference $-2.8$ nats), while the fit on the training set is practically the same (median NLL difference $0.0$ nats). A lower parameter complexity means that the parameters can be encoded with fewer nats, but it does not change the predictions and therefore does not imply a better test error for the same fit on the training set. The NLL part of the DL difference correlates with the difference in test RMSE (Spearman's $\rho = 0.26$). MO-DL finds expressions with a lower test RMSE more often in runs in which its DL is substantially lower, but this is offset by runs in which MO-DL ends with a higher DL than MO-Len-DL (\Cref{tab:modl-dl-bins} in \Cref{sec:appendix-mo-dl}).

Comparing the median test $R^2$ of SO with all criteria and of MO-Len-DL for all datasets (\Cref{tab:r2-so} in the supplementary material) shows that MO-Len-DL has the best median test $R^2$ for 17 of the 34 datasets, SO-BIC for 8, SO-FBF for 5, SO-BIC$_\text{SR}$ and SO-DL for 3 each, and SO-AIC for 1. MO-Len-DL is better than SO-AIC ($85.2~\%^\dagger$), SO-BIC$_\text{SR}$ ($81.3~\%^\dagger$), SO-FBF ($83.6~\%^\dagger$), and SO-DL ($78.3~\%^\dagger$).

Convergence line charts (\Cref{sec:appendix-single-objective}) suggest that directly including a function complexity penalty (BIC$_\text{SR}$, DL, FBF) into the fitness function creates strong simplicity pressure, rapidly reducing diversity and causing premature convergence to short underfit expressions. The $k \log n$ complexity requires a matching linear improvement in the log-likelihood with growing expression length $k$. Among individuals with similar training errors, the shorter ones are selected consistently, even with the low selection pressure of tournaments of size two, which limits the growth of program length. In contrast, multi-objective GP keeps expressions of all lengths in the population and thus preserves genetic diversity. 
SO-BIC produces results that are closest to MO-Len ($P(\mathrm{DL}>\text{SO-BIC})=61.4~\%$, 12~\% of the probability mass in the ROPE) because it allows longer expressions  but it leads to overfitting in several datasets (cf. \Cref{sec:appendix-single-objective}).

\section{Detailed Analysis}

\subsection{Stability of predictions and extrapolation}
\noindent

\begin{figure*}
  \center
  \includegraphics{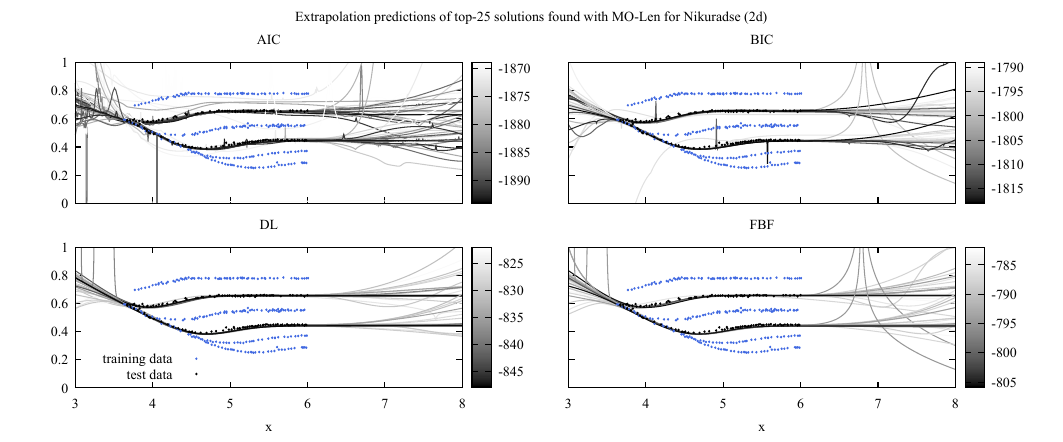}
  \caption{Test predictions of the top-25 models found with MO-Len and selected by AIC, BIC, DL, and FBF on the \emph{Nikuradse (2d)} dataset. AIC overfits extremely, the best BIC models are also overfit with unreliable extrapolation behaviour. The best DL/FBF models produce no poles for in-sample predictions and better extrapolation behaviour. Colours correspond to selection criterion values.
  The blue dots represent training observations, black dots are test observations. The dataset has two variables, of which only the first is shown. The second variable has six distinct values, roughly visible in the plots as lines.}
  \label{fig2:niku1}
\end{figure*}

The example shown in \Cref{fig2:niku1} demonstrates that MDL selection can produce qualitatively good predictions. It shows predictions of the top-25 MO-Len models selected by AIC, BIC, DL, and FBF for the \emph{Nikuradse (2d)} dataset. The test observations require interpolation and we plot the predictions over a larger range to highlight the differences in extrapolation.
AIC selects overfit models. BIC is better but also selects overfit models. DL and FBF models fit test observations and extrapolate reasonably well. Physically, behaviour should approach a constant as $x\rightarrow \infty$ and a decreasing linear function as $x\rightarrow 0$~\citep{nikuradse1950,Guimera2020}.

\subsection{Effect of training set size and noise level}\label{sec:effect-size-noise}
\noindent
\Cref{fig:sigma-effect} (top row) shows the effect of different noise levels $\sigma_\text{err}$ on Friedman-2
for MO-Len with MDL selection. This synthetic benchmark of \citet{FriedmanMARS} tests the identification of relevant variables, variable interactions, and nonlinearity. The generating expression is $y =  10\, \sin(\pi x_1\, x_2) + 20\, (x_3 - 1/2)^2 + 10\, x_4 + 5\, x_5 + \varepsilon $ with a training set of 100 observations of 10-dimensional $x_{1..10} \sim U(0,1)$ and Gaussian noise $\varepsilon \sim N(0, \sigma^2_\text{err})$. The generating function accounts for 96\% of the variance of $y$ for $\sigma_\text{err}=1$. Training RMSE (first panel) converges to the noise level (black dashed lines) in most runs, indicating that MO-Len-DL can infer an appropriate noise threshold. Lower noise settings yield models closer to ground truth on the noiseless test set (second panel). DL is also lower at low noise, mainly because NLL is lower (third panel). Most importantly, MDL adapts model size to noise: longer expressions at low noise, shorter expressions at high noise (right panel).
Overfitting does not occur at any noise level. Even though the length limit allowed expressions of up to 100 symbols, MDL automatically selected expressions with a length around 20 to 30 symbols for the lower noise levels. The generating expression for this dataset has a length of 22 symbols (dashed black line).

The bottom row of \Cref{fig:sigma-effect} shows the same linecharts for SO-DL which exhibit premature convergence and underfitting when DL is used directly as fitness.
\begin{figure*}
  \center
  \includegraphics[page=4]{noise-effect.pdf}
  \includegraphics[page=4]{noise-effect-so.pdf}

  \caption{MO-Len-DL (top row) automatically adjusts to the noise level. No overfitting occurs, as the training RMSE of MDL models converges to the irreducible error $\sigma_\text{err}$ and the test error approaches the ground truth. MDL automatically selects longer expressions for low noise levels and shorter expressions for high noise levels. SO-DL (bottom row) leads to premature convergence and underfitting.
  Horizontal dashed lines show the irreducible error $\sigma_\text{err}$ (left panel), and the length of the generating expression (22, right panel).
  }
  \label{fig:sigma-effect}
\end{figure*}

\Cref{fig:n-effect} (top row) shows the effect of training-set size on Friedman-1 under MO-Len-DL. This benchmark is also from \citet{FriedmanMARS}. The generating expression is $\exp(4\,x_1)/10 + 4 / (1 + e^{-20 \, x_2 + 10})+ 3\, x_3 + 2\, x_4 + x_5 + \varepsilon$ with $x_{1..10} \sim U(0,1)$ and Gaussian noise $\varepsilon \sim N(0,\sigma^2_\text{err})$ with $\sigma_\text{err}=1$. The length of the generating expression is 24 symbols. We keep $\sigma_\text{err}$ fixed and vary the training set size $m$ from 50 to 1000. For this dataset, $m \geq 100$ is needed for MO-Len-DL to reliably reach training RMSE near the irreducible error (first panel). SO-DL produces slightly worse training and test RMSE.
With smaller $m$, MDL selects shorter expressions. With larger $m$, selected lengths approach the generating expression length. MDL suppresses the bloating tendency of GP; the training error approaches the irreducible error and program length does not grow unnecessarily. The third panel shows DL per observation; it is mainly driven by the NLL, so models with larger training error have larger DL.

\begin{figure*}
  \center
  \includegraphics[page=4]{size-effect.pdf}
  \includegraphics[page=4]{size-effect-so.pdf}
  \caption{MO-Len-DL (top row) automatically adjusts to the number of observations in the training set ($m$). Smaller $m$ leads to selection of shorter expressions with larger training and test errors. SO-DL (bottom row) produces slightly worse results. Horizontal dashed lines show the irreducible error $\sigma_\text{err}$ (left panel), and the length of the generating expression (24, right panel)}
  \label{fig:n-effect}
\end{figure*}

\subsection{Program length preference of model selection criteria}
\noindent
The results above indicate that BIC$_\text{SR}$, DL, and FBF favour short expressions, which can hurt SO convergence on smaller datasets. To quantify size preference, we analyse final MO-Len Pareto fronts using the best-test-error model as reference and compare criterion-selected models relative to it.

Over all datasets, BIC$_\text{SR}$, DL, and FBF often select slightly shorter expressions than the best-test-error model, whereas BIC usually selects longer expressions, has larger variance, and frequently selects overfit expressions (\Cref{fig:boxplot-relative-size} in the supplementary material).

\begin{figure*}
  \center
  \includegraphics[]{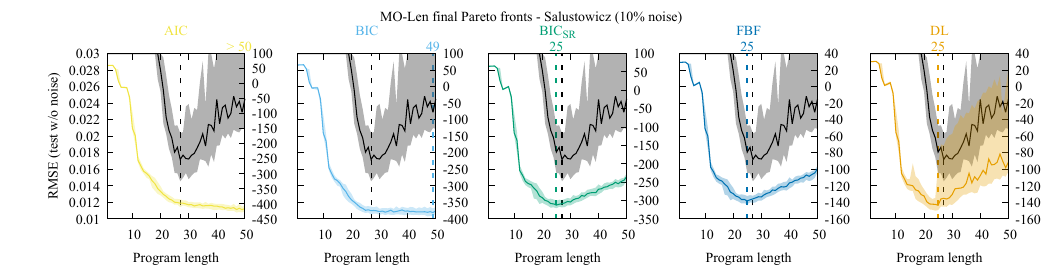}
  \caption{Test errors of the models in the final Pareto front for the different model selection criteria. The black line shows the median RMSE on the test set. The coloured lines show the median selection criterion value. The function complexity used in BIC$_\text{SR}$, DL, and FBF produces a clear minimum for the criteria around program length 25, which is close to the length of the generating expression (23), but is slightly shorter than the average length of the best test error expressions.}
  \label{fig3}
\end{figure*}

A detailed view of one dataset (Salustowicz, \Cref{fig3}) clarifies this effect. All five panels show the same distribution of test RMSE over expression length in the final Pareto fronts from 100 MO-Len runs (black). The black line is the median RMSE per length bin; the grey band is the interquartile range. The minimum test error occurs near length 27. The generating expression has length 23. Colored curves show each selection criterion (right axis), computed from noisy training data only. BIC$_\text{SR}$, FBF, and DL show clear minima, whereas the BIC is nearly flat at larger lengths, producing higher length variance. AIC shows a monotonic downward trend with no localised minimum.
The minimum median criterion value is marked by a vertical dashed line. The criteria including the function complexity penalty are more conservative and select shorter expressions on average than AIC or BIC.

\section{Discussion}\label{sec:discussion}
\noindent
The relative performance differences observed across DL, FBF, and BIC reflect regime-dependent regularisation effects rather than universal superiority of one metric. When the signal-to-noise ratio is low or the sample size is limited, stronger structural penalties (BIC$_\text{SR}$, DL, FBF) reduce variance and tend to yield better predictive stability. This is consistent with \citet{FajardoFontiveros2023}, who showed that above a certain noise level the true model cannot be identified from the data by any method, while probabilistic model selection still yields optimal generalisation. As sample size increases, differences between the criteria diminish as the likelihood term dominates and all consistent metrics concentrate on the true model. Thus, the results primarily characterise how each scoring rule navigates the bias–variance trade-off across regimes, rather than establishing a single metric as uniformly optimal.

In our experiments, multi-objective optimisation has a clear advantage over single-objective optimisation possibly because it preserves diversity and mitigates premature convergence when used with BIC$_\text{SR}$, DL, or FBF. Future work should explore direct search for MDL-optimal expressions without this failure mode. Alternatives to non-dominated sorting on NLL and length include Pareto tournaments~\citep{Haut2025,Smits2005}, diversity-preservation schemes such as novelty search~\citep{Lehman2011NoveltySearch}, and quality-diversity methods such as MAP-Elites~\citep{mouret_mapelites}.

The rotated-lattice parameter complexity does not penalize redundant parameters and can therefore accept such expressions that would be penalized more strongly by the original formula which requires only the diagonal of the FIM. When we compared the different parameter complexity variants we found no difference, and we do not claim that one of the variants is superior for GPSR.
Additionally, \citet{Leser2024} compared the Laplace approximation with sequential Monte Carlo estimates of the marginal likelihood for SR and found that they can differ substantially, which can change the expressions selected in GPSR.

Several related methods can help choose hyperparameters that affect expression complexity in GPSR, but they do not directly provide model-level selection scores and thus cannot guide search in the same way. Cross-validation (CV) can estimate the expected prediction error for settings such as length limits, function sets, or parsimony coefficients, but it requires repeated runs and cannot directly score individual candidate models generated within a run.

Structural risk minimisation~\citep{Vapnik1995} is another route for selecting among model classes, but it requires VC-dimension (Vapnik–Chervonenkis-dimension) estimates. Despite proposed approximations~\citep{Borges2010,Chen2016,Chen2019}, the VC dimension of SR expression trees with nonlinear operators and fitted parameters remains unresolved, and estimates may yield very loose bounds.

Finally, countable-hypothesis bounds and PAC-Bayes bounds~\citep{McAllester1999,Alquier2024} upper-bound expected test error using training error and a complexity penalty. These bounds are linked to Bayesian marginal likelihoods~\citep{Germain2016} and can support model selection, but for GPSR the basic bounds are usually vacuous and tighter bounds are not yet available. \citet{Nomura2026} recently derived a Rademacher-complexity bound for GPSR that, similar to DL, separates into training error, structural complexity, and parameter complexity terms; however, the bound is purely theoretical and requires calibration parameters. Deriving tighter GPSR-specific bounds and comparing them with MDL/FBF selection is an important future direction.

SRBench++ datasets~\citep{SRBench_Aldeia} may be less representative for benchmarking SR systems that target complex expressions, because on many of these datasets, all but the shortest expressions overfit. In our tests, MDL and FBF effectively prevented overfitting and typically selected only short expressions of around 10 symbols. The main exceptions are Friedman instances 579, 606, and 650, which require larger models. We therefore do not recommend this subset alone for assessing the ability of SR algorithms to recover complex multivariate nonlinear relations. The Tower dataset is another exception: AIC gave the best test results there, and we could not induce overfitting even with larger length limits and more generations. A likely reason is that this dataset originates from a multivariate time series that was shuffled before train/test splitting, violating the i.i.d.-error likelihood assumption used here.

\subsection{Limitations}\label{sec:limitations}
\noindent
While all selection criteria tested in this paper can be used with any likelihood function, we only performed empirical tests with the Gaussian likelihood. It would be interesting to test other likelihoods, for example for classification problems.
For the real-world datasets, we assumed a homoscedastic Gaussian likelihood with i.i.d.\ errors, a common assumption in GPSR, because no information about the error distributions is available. However, industrial datasets can have heteroscedastic or long-tailed errors (outliers), errors in the input variables, or temporally correlated errors, and the data-generating process can be non-stationary. If the likelihood is misspecified, the selection criteria do not necessarily select the expression with the best trade-off between accuracy and complexity. The results for the Tower dataset discussed above are a likely example.
Because all tested criteria are based on the likelihood, such assumptions are explicit and can be adapted by replacing the likelihood, without changing the criteria or the GP algorithms. Our implementation provides likelihoods for known heteroscedastic measurement errors, long-tailed (Laplace) errors, errors in both the inputs and the target, and correlated errors with a given covariance structure, e.g., for time series. Robustness experiments with these likelihoods and with time-ordered splits for non-stationary time series are left for future work.

We used only a simple single-objective GP algorithm and only a single set of hyperparameters without optimisation for our comparisons. One could experiment with larger population sizes, different selection pressure, or mutation rates to improve the performance of single-objective GP with BIC$_\text{SR}$, DL, or FBF selection. In a small study, we found larger populations and tournaments lower the training error noticeably, at 10–20× the evaluations, but the median stays clearly above $\sigma_\text{err}$ (details in the appendix).

We found that the exact calculation of function complexity in MDL can have an effect on the quality of selected expressions. The $k \log n$ penalty for expressions of length $k$ consisting of $n$ distinct symbols excluding constants and fitting parameters works well on average over all datasets. However, on individual datasets we found that the penalty was often slightly too strong, which led to selection of shorter expressions compared to the best test error expression (\Cref{fig:boxplot-relative-size} in the supplementary material). In our experiments, the function complexity was consistently larger than parameter complexity and therefore had a stronger effect on model selection. The function complexity in DL may depend on characteristics of the problem. How such problem-specific characteristics could be included in DL is an interesting topic for future research.

\section{Conclusion}\label{sec:conclusions}
\noindent
In summary, we recommend DL, BIC$_\text{SR}$, or FBF for model selection together with multi-objective GPSR that optimises accuracy and expression length. Across all datasets, the three criteria selected compact and accurate models. Differences between DL, BIC$_\text{SR}$, and FBF were small, but BIC$_\text{SR}$ and FBF are computationally slightly cheaper than DL. Because many GPSR systems already optimise accuracy and complexity jointly, this recommendation is straightforward to implement by selecting from the final Pareto set or archive. The additional overhead is modest, since DL is only needed for Pareto-front candidates.

We did not find a clear advantage in replacing expression length with the DL penalty as the second optimisation objective. The final DL was shorter but it did not change test errors. The runtime was similar, because the cost of the FIM and SVD for all candidates (about 38~\% of the runtime) was offset by the shorter expressions in the population, which are cheaper to fit.

We do not recommend using BIC$_\text{SR}$, DL, or FBF directly as single-objective fitness. In our experiments, this frequently caused premature convergence and underfitting.

We also note that BIC$_\text{SR}$, DL, and FBF are merely approximations to the Bayesian probability of a function given the data under given choices of parameter and function priors. Because evaluating the DL for model selection is cheap compared to the optimisation, future work should use more sophisticated (and perhaps more expensive) methods, such as nested sampling, bridge sampling, or the harmonic estimator, to obtain a better estimate of this probability.

These conclusions are not specific to GP-based symbolic regression. They should transfer to any equation-discovery workflow that produces candidate models and defines a likelihood describing the hypothesised data-generating process. The same principles can also extend beyond regression to classification with an appropriate likelihood.

\section*{Code availability}
\noindent
The GP system used for all experiments is available as the Julia package NeoGP.jl at \url{https://github.com/heal-research/NeoGP.jl}. Python bindings for NeoGP.jl are available at \url{https://github.com/heal-research/pyNeoGP}.

\section*{Acknowledgements}
\noindent
G.K. acknowledges support by the Austrian Federal Ministry for Climate Action, Environment, Energy, Mobility, Innovation and Technology, the Federal Ministry for Labour and Economy, and the regional government of Upper Austria within the COMET project ProMetHeus (904919) supported by the Austrian Research Promotion Agency (FFG).
D.J.B. acknowledges that support was provided by Schmidt Sciences, LLC. H.D. is supported by a Royal Society University Research Fellowship (grant no. 211046).
F.O.F. is supported by Conselho Nacional de Desenvolvimento Cient\'{i}fico e Tecnol\'{o}gico (CNPq) grant 301596/2022-0.

We used GNU parallel \citep{gnuparallel} for running experiments and data processing.

\small
\bibliographystyle{IEEEtranN}
\bibliography{symreg}

\clearpage
\onecolumn
\normalsize
\setlength{\parskip}{0.5\baselineskip plus 2pt minus 1pt}
\setcounter{table}{0}
\setcounter{figure}{0}
\setcounter{algorithm}{0}
\renewcommand{\thetable}{S\Roman{table}}
\renewcommand{\thefigure}{S\arabic{figure}}
\renewcommand{\thealgorithm}{S\arabic{algorithm}}
\renewcommand{\theHtable}{S\arabic{table}}
\renewcommand{\theHfigure}{S\arabic{figure}}
\renewcommand{\theHalgorithm}{S\arabic{algorithm}}

\appendix[Supplementary Material]
\normalsize

\subsection{Algorithms} \label{sec:appendix-algorithms}
\begin{algorithm}[!ht]
  \caption{Multi-objective GPSR using nondominated sorting  (MO). $P_g$ is the population of expressions at generation $g$, $Q_g$ are the candidate expressions, Obj$_g$ is the vector of objective value tuples, rank$_g$ is the vector of Pareto ranks, and crowd$_g$ is the vector of crowding distances. The $i$-th elements of Obj$_g$, rank$_g$, and crowd$_g$ belong to the expression $P_g[i]$; these vectors are merged, indexed, and truncated together with the population. $A \circ B$ denotes the concatenation of the vectors $A$ and $B$, and $A[I]$ the elements of $A$ at the indices $I$.
  }
  \label{alg:multi-objective}
    \begin{algorithmic}[1]
          \Procedure{GP}{popSize, maxGen, maxLen, crossoverRate, $\mathcal{F}$, $\mathcal{T}$}
          \State D $ := $ loadData()
          \State rand $:=$ initRandom()
          \State $g := 0$
          \State \(\triangleright\)  {Generate pop. of expressions}
          \State P$_0 :=$ initPopulation(rand, $\mathcal{F}$, $\mathcal{T}$, maxLen, popSize)
          \State Obj$_0 :=$ evaluate(P$_0$, D) \Comment{vector of objective value tuples}
          \State \(\triangleright\) {Pareto ranks and crowding distances}
          \State rank$_0$, crowd$_0 := $ nondominatedSort(Obj$_0$)
          \While {$g <$ maxGen}
          \State Q$_g := ()$ \Comment{empty vector}
          \For{$i := 1 \ldots \text{popSize}$}\Comment{generate popSize candidates}
          \State $p_1 := $ tournament(rand, P$_g$, rank$_g$, crowd$_g$, groupSize)
          \If{rand() $ < $ crossoverRate}
              \State $p_2 := $ tournament(rand, P$_g$, rank$_g$, crowd$_g$, groupSize)
              \State $c := $ subtreeCrossover(rand, p$_1$, p$_2$, maxLen)
          \Else
              \State $c :=$ mutation(rand, $\mathcal{F}, \mathcal{T}$, $p_1$)
          \EndIf
          \State Q$_{g} :=$ Q$_{g} \circ (c)$ \Comment{append}
          \EndFor
          \State R $:=$ P$_g \circ$ Q$_g$ \Comment{merge parents and offspring}
          \State Obj$_\text{R} :=$ Obj$_g \circ$ evaluate(Q$_g$, D) \Comment{only the offspring are evaluated}
          \State rank$_\text{R}$, crowd$_\text{R} := $ nondominatedSort(Obj$_\text{R}$)
          \State $I := $ truncate(rank$_\text{R}$, crowd$_\text{R}$, popSize) \Comment{indices of the popSize survivors}
          \State P$_{g+1}$, Obj$_{g+1}$, rank$_{g+1}$, crowd$_{g+1} := $ R$[I]$, Obj$_\text{R}[I]$, rank$_\text{R}[I]$, crowd$_\text{R}[I]$
          \State best$_{g+1} :=$ modelSelection(P$_{g+1}$, rank$_{g+1}$) \Comment{select from first front}
          \State $ g := g + 1$
          \EndWhile
          \State return best$_{\text{maxGen}}$
        \EndProcedure
	\end{algorithmic}
\end{algorithm}

\begin{algorithm}[!ht]
  \caption{Single-objective GPSR (SO) with model selection criteria as fitness. $P_g$ is the population of expressions at generation $g$, Obj$_g$ is the vector of fitness values for each expression, i.e., Obj$_g[i]$ is the fitness of $P_g[i]$ (smaller is better). $A \circ B$ denotes the concatenation of the vectors $A$ and $B$.}
  \label{alg:basic}
  \begin{algorithmic}[1]
          \Procedure{GP}{popSize, maxGen, maxLen, crossoverRate, $\mathcal{F}$, $\mathcal{T}$}
          \State D $ := $ loadData()
          \State rand $:=$ initRandom()
          \State $g := 0$
          \State P$_0 :=$ initPopulation(rand, $\mathcal{F}$, $\mathcal{T}$, maxLen, popSize) \Comment{Generate vector of expressions}
          \State Obj$_0 :=$ evaluate(P$_0$, D)
          \State best$_0 :=$ P$_0[\arg\min_i \text{Obj}_0[i]]$ \Comment{expression with the smallest fitness}
          \While {$g <$ maxGen}
          \State Q$_g := (\text{best}_g)$ \Comment{the elite is copied to the first position}
          \For{$i := 2 \ldots \text{popSize}$}
          \State $p_1 := $ tournament(rand, P$_g$, Obj$_g$, groupSize)
          \If{rand() $ < $ crossoverRate}
              \State $p_2 := $ tournament(rand, P$_g$, Obj$_g$, groupSize)
              \State $c := $ subtreeCrossover(rand, p$_1$, p$_2$, maxLen)
          \Else
              \State $c :=$ mutation(rand, $\mathcal{F}, \mathcal{T}, p_1$)
          \EndIf
          \State Q$_{g} :=$ Q$_{g} \circ (c)$ \Comment{append}
          \EndFor
          \State P$_{g+1} := $ Q$_g$ \Comment{generational replacement}
          \State Obj$_{g+1} :=$ evaluate(P$_{g+1}$, D) \Comment{all expressions including the elite are evaluated}
          \State best$_{g+1} :=$ P$_{g+1}[\arg\min_i \text{Obj}_{g+1}[i]]$
          \State $ g := g + 1$
          \EndWhile
          \State return best$_\text{maxGen}$
        \EndProcedure
	\end{algorithmic}
\end{algorithm}

\begin{algorithm}[!ht]
  \caption{Evaluation with memetic optimisation of parameters used by all algorithm variants. $P$ is the population of expressions, $D$ is the dataset with $m$ observations. One of the statements in lines 11 -- 17 is executed depending on the algorithm variant. round$_2$ rounds to two decimal places.}
  \label{alg:eval}
    \begin{algorithmic}[1]
        \Procedure{evaluate}{$P, D$}
            \For{$i := 1 \ldots |P|$} \Comment{for each expression in the population}
                \State expr := $P[i]$
                \State $\theta, \sigma^2_\text{err} := $ extractParam(expr) \Comment{starting values stored in the expression}
                \State $\hat{\theta}, \hat\sigma^2_\text{err} := $ optimize(negLogLik, $\theta, \text{expr}, D$) \Comment{optimisation of NLL, keeps $\theta$ if not improved}
                \State updateParam($P[i], \hat{\theta}, \hat\sigma^2_\text{err}$) \Comment{update expression with optimised parameters}
                \State nll $:=$ negLogLik($\hat{\theta}, \hat\sigma^2_\text{err}, \text{expr}, D$) \Comment{NLL is always required}
                \State $p := \text{length}(\hat{\theta}) + 1$ \Comment{total parameter count $p {}= p_\theta + 1$ includes $\hat{\sigma}^2_\text{err}$}
                \State
                \State \(\triangleright\) {Different objective values for different algorithm variants}
                \State Obj$[i]$ := (round$_2$(nll), length(expr)) \Comment{MO-Len}
                \State Obj$[i]$ := (round$_2$(nll), round$_2$(func\_compl + param\_compl)) \Comment{MO-DL}
                \State Obj$[i]$ := $2\, \text{nll} + 2\, p$ \Comment{SO with AIC}
                \State Obj$[i]$ := $2\, \text{nll} + p \log m$ \Comment{SO with BIC}
                \State Obj$[i]$ := DL(expr, $\hat{\theta}, \hat\sigma^2_\text{err}, D$) \Comment{SO with DL}
                \State Obj$[i]$ := FBF(expr, $\hat{\theta}, \hat\sigma^2_\text{err}, D$) \Comment{SO with FBF}
                \State Obj$[i]$ := $2\, \text{nll} + 2\, \text{func}_\text{compl} + p \log m$ \Comment{SO with BIC$_\text{SR}$}
                \State replace NaN elements of Obj$[i]$ by $+\infty$
            \EndFor
            \State return Obj
        \EndProcedure
	\end{algorithmic}
\end{algorithm}

\subsubsection{Population update and objective values}\label{sec:appendix-population-update}
The objective values of an expression are calculated once, when the expression is evaluated after its creation (\Cref{alg:eval}), and are stored with the expression. Whenever the population is merged, indexed, or truncated, the vectors of objective values (and, in MO, of Pareto ranks and crowding distances) are merged, indexed, or truncated in the same way. The $i$-th element of each of these vectors therefore always belongs to the $i$-th expression of the population (\Cref{alg:multi-objective,alg:basic}).

In MO (\Cref{alg:multi-objective}), the parents are not evaluated again. The offspring are evaluated and merged with the parents, and the Pareto ranks and crowding distances are calculated for the merged vector of $2\,\text{popSize}$ expressions as in NSGA-II~\citep{Deb2002}. The procedure truncate selects the survivors front by front in the order of increasing rank. From the first front that does not fit completely, the expressions with the largest crowding distances are selected. The survivors keep the ranks and crowding distances calculated for the merged vector, and these are used in the tournaments of the next generation. A tournament draws groupSize expressions uniformly at random with replacement and returns the expression with the smallest rank; ties are broken in favour of the larger crowding distance.
Our implementation differs from NSGA-II in the handling of duplicates. Expressions with equal objective tuples are duplicates. Only the first of them in the merged vector (parents precede offspring) is ranked by non-dominated sorting; all others are assigned to the last front. This prevents copies of the same expression from filling the first front. Because the NLL and the DL penalty are rounded to two decimal places (\Cref{alg:eval}), expressions with the same rounded objective values are also treated as duplicates.
In each generation, each model-selection criterion is calculated for all expressions of the first front and selects the expression with the smallest value. If several expressions have the same smallest value, the one with the smallest NLL is selected.

In SO (\Cref{alg:basic}), the population of the next generation consists of the best expression of the current generation (elite) and popSize$-1$ offspring. All of these expressions, including the elite, are evaluated. The parameters of the elite are therefore optimised again, starting from their current values. Because the optimiser only accepts parameter values that reduce the NLL (\Cref{sec:appendix-numerical}), the NLL of the elite cannot increase. A tournament draws groupSize expressions uniformly at random with replacement and returns the expression with the smallest fitness.

\subsubsection{Parameter optimisation, numerical problems, and invalid expressions}\label{sec:appendix-numerical}
\paragraph{Parameter optimisation}
The regression parameters $\theta$ are fitted with the Levenberg-Marquardt (LM) algorithm~\citep{Levenberg_1944,Marquardt_1963}, using the implementation of the Julia package GeodesicLM.jl without geodesic acceleration. The noise variance is profiled: for given $\theta$, $\hat\sigma^2_\text{err} = \max(\text{MSE}(\theta), \sigma^2_\text{min})$, so that LM solves the least-squares problem for $\theta$ only. The lower bound $\sigma_\text{min} = \epsilon\, \text{rms}(y)$ prevents an NLL of $-\infty$ for expressions that fit the training data exactly, where $\epsilon \approx 1.2 \cdot 10^{-7}$ is the machine epsilon of single-precision floating-point numbers. The Jacobian of the residuals is calculated by reverse-mode automatic differentiation of the expression. LM stops after at most 100 iterations or 50 evaluations of the residuals, or when the relative decrease of the sum of squared residuals is below $\sqrt{\epsilon} \approx 3.5 \cdot 10^{-4}$ in three consecutive iterations. Expressions are evaluated in single precision, while LM works in double precision.
The starting values are the parameter values stored in the expression~\citep{Kommenda2019}. The parameters of the expressions of the initial population are drawn from $N(0, 1)$. Offspring inherit the optimised parameter values of their parents, and mutation can add $N(0, 1)$-distributed noise to parameter values.

\paragraph{Failed parameter optimisation}
The optimised parameter values are only accepted if they reduce the NLL compared with the starting values. If LM stops because the maximum number of evaluations is reached, the best parameter values found so far are used. If the NLL is not reduced, or if the optimisation fails with an exception, e.g., because of non-finite values, the starting values are kept. In all cases, the evaluation continues with the resulting parameter values, the objective values are calculated for these values, and the expression remains in the population. A failed parameter optimisation therefore never stops a run and never increases the NLL of an expression. NaN entries of the Jacobian are set to zero.

\paragraph{Invalid expressions}
The function set contains the protected functions $\log|x|$, $\sqrt{|x|}$, and $|x|^y$, but division is not protected. Expressions can therefore produce non-finite outputs (NaN or $\pm\infty$), e.g., by division by zero, $\log|0|$, or an overflow of $\exp$. Such expressions are neither repaired nor removed. If the output of an expression is not finite for at least one training observation, its NLL is not finite and is replaced by the largest finite floating-point number ($\approx 3.4 \cdot 10^{38}$). Objective values that are NaN are replaced by $+\infty$ (line~18 of \Cref{alg:eval}). Invalid expressions are thus dominated by all valid expressions of the same or a smaller length (MO-Len) and have a worse fitness than all valid expressions (SO).

\paragraph{Numerical problems in model selection}
AIC, BIC, BIC$_\text{SR}$, and FBF depend on the parameter values only through the NLL, which is always finite. The parameter complexity of DL additionally requires the observed Fisher information matrix $\mathrm{FIM}(\hat\vartheta)$, i.e., the Hessian of the NLL with respect to all $p$ elements of $\hat\vartheta = (\hat\theta_1, \ldots, \hat\theta_{p_\theta}, \hat\sigma^2_\text{err})$. We calculate the FIM by forward-mode automatic differentiation of the analytic gradient of the NLL and symmetrise it. If the FIM contains NaN values or values larger than $10^{30}$, or if its eigendecomposition fails, the parameter complexity is set to the largest finite floating-point number. The DL of such an expression is effectively infinite, so it is not selected by DL. Negative eigenvalues of the FIM, which can occur if the optimiser stopped before reaching a minimum of the NLL, are set to zero, i.e., the corresponding parameter directions are treated as having zero precision. The magnitude term of the parameter complexity is clamped at zero for each direction, so each direction costs at least $\log 2$ for its sign and no direction has a negative codelength. When a criterion is used to select an expression from the first front, NaN values of the criterion are treated as $+\infty$.

\subsection{Synthetic datasets}\label{sec:appendix-synthetic-datasets}
\Cref{tab:synthetic-datasets} lists the generating functions and the definitions of the training and test sets of the synthetic datasets.
For each dataset, we generated 100 training sets with targets $y = f(\mathbf{x}) + \varepsilon$, where $\varepsilon \sim N(0, \sigma^2_\text{err})$ i.i.d., and a single noiseless test set with $y = f(\mathbf{x})$, which is used for all 100 training sets.
For Friedman-1 and Friedman-2, the inputs are sampled anew for each of the 100 training sets, the variables $x_6, \ldots, x_{10}$ do not affect $f$, and the noise standard deviation $\sigma_\text{err}$ is set to fixed values.
For all other datasets, the training inputs are the same for all 100 training sets and only the noise realisations differ. The noise standard deviation is 10\% of the standard deviation of the noiseless function values over the training inputs, $\sigma_\text{err} = 0.1\, \mathrm{sd}(f(\mathbf{x}))$, i.e., the noise variance $\sigma^2_\text{err}$ is 1\% of the variance of $f(\mathbf{x})$.

\begin{table}[ht]
  \centering
  
  \caption{Generating functions $f(\mathbf{x})$ and definition of the training and test sets of the synthetic datasets. $m$ is the number of observations. $U(a, b)$ denotes independent uniform sampling of each listed variable from the interval $[a, b]$. $\{a, a+h, \ldots, b\}$ denotes a regular grid with step size $h$; multivariate grids are the Cartesian product of the grids of the individual variables. Training targets contain i.i.d. Gaussian noise with standard deviation $\sigma_\text{err}$; test targets are noiseless.}
  \label{tab:synthetic-datasets}
  \small
  \renewcommand{\arraystretch}{1.5}
  \setlength{\tabcolsep}{4pt}
  \begin{tabular}{l|l|>{\raggedright\arraybackslash}p{1.6in}|>{\raggedright\arraybackslash}p{1.85in}}
    Dataset & $f(\mathbf{x})$ & Training set & Test set \\
    \hline
    Friedman-1
      & $\dfrac{e^{4 x_1}}{10} + \dfrac{4}{1 + e^{-20 x_2 + 10}} + 3 x_3 + 2 x_4 + x_5$
      & $m \in \{50, 100, 200, 500, 1000\}$ \newline $x_1, \ldots, x_{10} \sim U(0, 1)$ \newline $\sigma_\text{err} = 1$
      & $m = 1000$ \newline $x_1, \ldots, x_{10} \sim U(0, 1)$ \\
    Friedman-2
      & $10 \sin(\pi x_1 x_2) + 20 (x_3 - 1/2)^2 + 10 x_4 + 5 x_5$
      & $m = 100$ \newline $x_1, \ldots, x_{10} \sim U(0, 1)$ \newline $\sigma_\text{err} \in \{0.5, 1, 2, 4.8, 13.6\}$
      & $m = 1000$ \newline $x_1, \ldots, x_{10} \sim U(0, 1)$ \\
    Kotanchek
      & $\dfrac{e^{-(x_1 - 1)^2}}{1.2 + (x_2 - 2.5)^2}$
      & $m = 100$ \newline $x_1, x_2 \sim U(0.3, 4)$ \newline $\sigma_\text{err} = 0.0235$
      & $m = 2025$ \newline $x_1, x_2 \in \{-0.2, -0.1, \ldots, 4.2\}$ \\
    RatPol2d
      & $\dfrac{(x_1 - 3)^4 + (x_2 - 3)^3 - (x_2 - 3)}{(x_2 - 2)^4 + 10}$
      & $m = 50$ \newline $x_1, x_2 \sim U(0.05, 6.05)$ \newline $\sigma_\text{err} = 0.171$
      & $m = 1156$ \newline $x_1, x_2 \in \{-0.25, -0.05, \ldots, 6.35\}$ \\
    RatPol3d
      & $\dfrac{30 (x_1 - 1) (x_3 - 1)}{x_2^2 (x_1 - 10)}$
      & $m = 300$ \newline $x_1, x_3 \sim U(0.05, 2)$ \newline $x_2 \sim U(1, 2)$ \newline $\sigma_\text{err} = 0.0620$
      & $m = 2700$ \newline $x_1, x_3 \in \{-0.05, 0.1, \ldots, 2.05\}$ \newline $x_2 \in \{0.95, 1.05, \ldots, 2.05\}$ \\
    Ripple
      & $(x_1 - 3)(x_2 - 3) + 2 \sin\left((x_1 - 4)(x_2 - 4)\right)$
      & $m = 300$ \newline $x_1, x_2 \sim U(0.05, 6.05)$ \newline $\sigma_\text{err} = 0.348$
      & $m = 1000$ \newline $x_1, x_2 \sim U(-0.25, 6.35)$ \\
    Salustowicz
      & $x_1^3 e^{-x_1} \cos(x_1) \sin(x_1) \left(\sin^2(x_1) \cos(x_1) - 1\right)$
      & $m = 100$ \newline $x_1 \in \{0.05, 0.15, \ldots, 9.95\}$ \newline $\sigma_\text{err} = 0.0318$
      & $m = 220$ \newline $x_1 \in \{-0.5, -0.45, \ldots, 10.45\}$ \\
    Salustowicz2d
      & $f_\text{Salustowicz}(x_1)\, (x_2 - 5)$
      & $m = 606$ \newline $x_1 \in \{0, 0.1, \ldots, 10\}$ \newline $x_2 \in \{0.05, 2.05, \ldots, 10.05\}$ \newline $\sigma_\text{err} = 0.108$
      & $m = 2553$ \newline $x_1 \in \{-0.5, -0.4, \ldots, 10.5\}$ \newline $x_2 \in \{-0.5, 0, \ldots, 10.5\}$ \\
  \end{tabular}
  
\end{table}

\subsection{Detailed analysis}\label{sec:appendix-detailed-analysis}
\subsubsection{Stability of predictions}\label{sec:appendix-predictions}
\begin{figure}[h!]
  \center
  \includegraphics{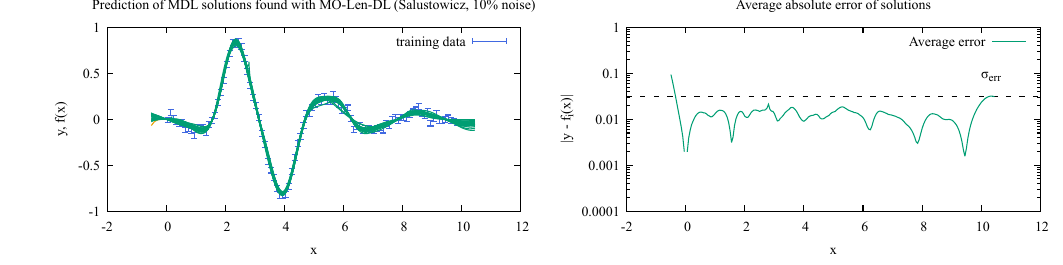}
  \caption{Predictions of the MDL models found with MO-Len on the test set. Dashed line shows $\sigma_\text{err}$, which is known for this synthetic test case. The mean absolute error is consistently below the noise level.}
  \label{fig2}
\end{figure}

\Cref{fig2} shows predictions from MDL-selected MO-Len models on Salustowicz. The right panel reports average absolute relative prediction error versus ground truth. There is no visible overfitting, and average residuals remain below the training noise threshold $\sigma_\text{err}$.

\subsubsection{Program length preference of model selection criteria}\label{sec:appendix-length-preference}
\begin{figure}[h!]
  \center
  \includegraphics[]{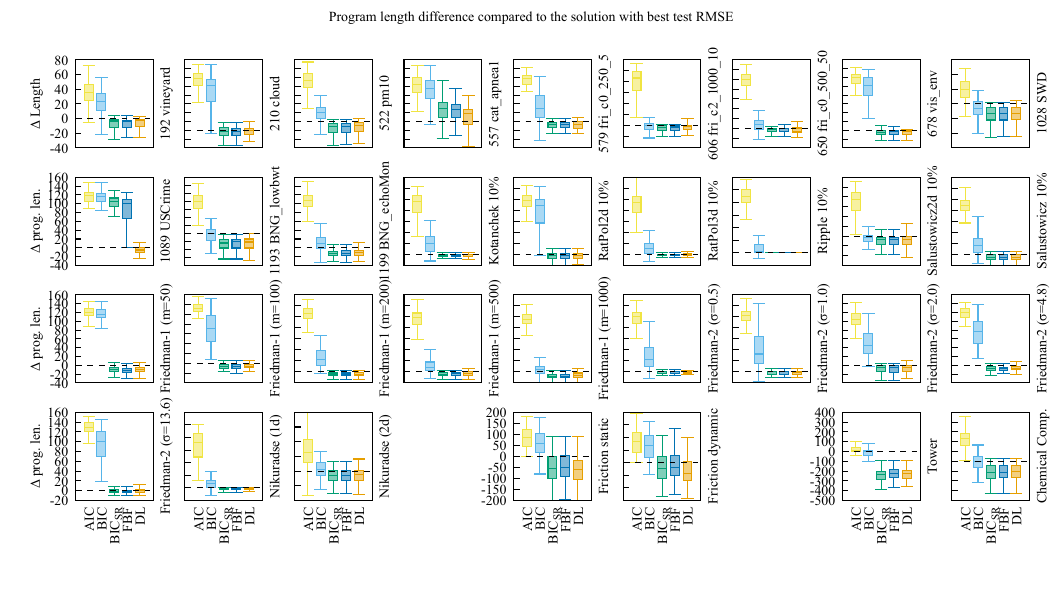}
  \caption{Program length difference of selected expressions compared to the expression with best test RMSE in each Pareto front. BIC$_\text{SR}$, DL, and FBF select expressions lengths close to the best test expressions. AIC and BIC often select models that are too long.}
  \label{fig:boxplot-relative-size}
\end{figure}

\Cref{fig:boxplot-relative-size} shows boxplots of the length differences of the selected expressions relative to the expression with the best test RMSE in each final MO-Len Pareto front. An ideal criterion would have a mean difference near zero and low variance. BIC$_\text{SR}$, DL, and FBF often select slightly shorter expressions than the best-test-error model. BIC usually selects larger expressions, has larger variance, and frequently selects overfit expressions. The corresponding differences in test RMSE are shown in \Cref{fig:boxplot-test-error}.

\begin{figure}[h!]
  \center
  \includegraphics[page=4]{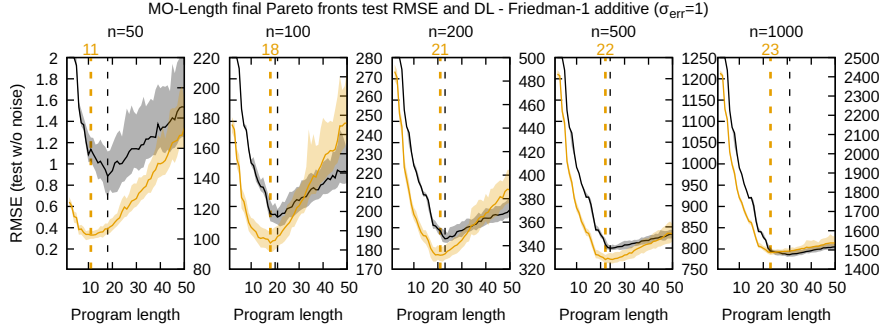}
  \caption{Test RMSE (black curve) and DL (orange curve) of expressions in the MO-Len final Pareto fronts for the Friedman-1 problem with varying training set size. The minimum of median DL and test RMSE are marked with dashed lines. The left y-axis shows the test RMSE, and the right y-axis shows the DL. With smaller datasets, DL prefers shorter expressions. The length of the generating expression for this dataset is 24.}
  \label{fig:n-effect-pareto}
\end{figure}

\Cref{fig:n-effect-pareto} compares DL across final MO-Len Pareto fronts for Friedman-1 at different training sizes. For the smallest dataset ($m=50$), DL's preference for short expressions is strongest. This explains the underfitting of MO-Len-DL at small $m$ (\Cref{sec:effect-size-noise}).

\subsection{MO-Len Results} \label{sec:appendix-multi-objective}
\noindent
\subsubsection{Win probabilities}
\noindent
We determine the number of wins of selection criterion $A$ over criterion $B$ as the number of pairings in which $A$ selected an expression with better test MSE than $B$. If both criteria select the same expression, the pairing results in a tie. We compare all possible pairs of criteria, whereby in each pairing the criteria select from the \emph{same} Pareto front, that is the MO-Len Pareto front. For each generating function, we sampled 100 datasets and executed independent GPSR runs to produce 100 pairings per generating function. The tables show the percentage of wins (\Cref{tab:wins}) and ties (\Cref{tab:ties}) over $34\times 100 \times 5 \times 4 / 2  $ (problems $\times$ samples $\times$ pairs) $= 34000$ pairings.
\begin{table}[ht]
\centering
\caption{Percentage of wins of the column criterion vs. the row criterion on the run level (BIC wins against AIC in 65.6\% of the pairings).}
\label{tab:wins}
\begin{tabular}{r|ccccc}
                    & AIC & BIC & BIC$_\text{SR}$ & FBF & DL \\
\hline
AIC                  &      & 65.6 & 84.6 & 84.7 & 86.1 \\
BIC                  & 16.9 &      & 66.3 & 66.5 & 68.4 \\
BIC$_\text{SR}$      & 14.5 & 26.9 &      &  5.6 & 27.2 \\
FBF                  & 14.6 & 27.1 &  5.1 &      & 26.9 \\
DL                   & 13.9 & 27.6 & 23.9 & 23.8 &      \\
\end{tabular}
\end{table}

\begin{table}[ht]
\centering
\caption{Percentage of ties of the column criterion vs. the row criterion on the run level. BIC$_\text{SR}$, DL, and FBF have many ties. Ties are counted as 1/2 wins for both teams.}
\label{tab:ties}
\begin{tabular}{r|ccccc}
                    & AIC & BIC & BIC$_\text{SR}$ & FBF & DL \\
\hline
AIC                  &      & 17.6 &  0.9 &  0.7 &  0.0 \\
BIC                  & 17.6 &      &  6.8 &  6.4 &  4.1 \\
BIC$_\text{SR}$      &  0.9 &  6.8 &      & 89.2 & 48.9 \\
FBF                  &  0.7 &  6.4 & 89.2 &      & 49.3 \\
DL                   &  0.0 &  4.1 & 48.9 & 49.3 &      \\
\end{tabular}
\end{table}

\Cref{tab:bbt} shows the summary statistics of the BBT posterior distributions for the pairwise win probabilities, with one match per dataset (\Cref{sec:statistical-comparisons}).
As suggested by \citet{Wainer2023}, we define the region of practical equivalence (ROPE) for $P(A > B)$ as $45\%$ to $55\%$. We report the median value for $P(A>B)$, the region of highest posterior density (HPD) which contains 95\% of probability mass, the probability mass larger 50\% ($P > 0.5$), and the probability mass within the ROPE. We interpret a high probability mass within ROPE as practically no  difference between methods.
 The rows are ordered by capability starting from the best to the worst method.

\begin{table}[ht]
\centering
\caption{Summary of BBT posterior distributions for pairwise win probabilities on the dataset level. All values are percentages. }
\label{tab:bbt}
\begin{tabular}{l|cccc}
Pairing & Median & HPD Delta & P $ > 0.5$ & in ROPE \\
\hline
$P(\text{DL}>\text{FBF})$ & $50.9$ & $23.8$ & 56 & 60 \\
$P(\text{DL}>\text{BIC$_\text{SR}$})$ & $53.1$ & $23.3$ & 69 & 53 \\
$P(\text{DL}>\text{BIC})$ & $80.2^\dagger$ & $16.2$ & 100 & 0 \\
$P(\text{DL}>\text{AIC})$ & $94.2^\dagger$ & $7.3$ & 100 & 0 \\
$P(\text{FBF}>\text{BIC$_\text{SR}$})$ & $52.2$ & $23.5$ & 64 & 56 \\
$P(\text{FBF}>\text{BIC})$ & $79.6^\dagger$ & $16.5$ & 100 & 0 \\
$P(\text{FBF}>\text{AIC})$ & $94.0^\dagger$ & $7.7$ & 100 & 0 \\
$P(\text{BIC$_\text{SR}$}>\text{BIC})$ & $78.2^\dagger$ & $17.3$ & 100 & 0 \\
$P(\text{BIC$_\text{SR}$}>\text{AIC})$ & $93.5^\dagger$ & $8.2$ & 100 & 0 \\
$P(\text{BIC}>\text{AIC})$ & $79.8^\dagger$ & $19.6$ & 100 & 0 \\
\end{tabular}
\end{table}

\Cref{tab:bbt-sensitivity} shows the BBT models with one match per dataset for all comparisons of test RMSE in the main text, and how they depend on the definition of a tie on a dataset. In the main text, method A wins the match on a dataset if it is better than method B in more than 55\% of the paired runs (ties counted as 1/2), B wins if A is better in less than 45\% of the runs, and otherwise the match is a tie, i.e., the band for a tie on a dataset is the same as the ROPE for $P(A > B)$. The table also shows the medians of $P(A>B)$ without a tie band (only a share of exactly 50\% is a tie) and with the wider band $42.1\%$--$57.9\%$. As validation, the last column shows the numbers of datasets won, tied, and lost with the local ROPE of~\citet{Wainer2023}: the match is a tie if $|d_z| \le 0.2$, where $d_z$ is the mean of the paired differences divided by their standard deviation (Eqs.~5 and~6 of~\citep{Wainer2023}), here of the log test RMSE, with non-finite test RMSE values replaced by the largest finite value of the two methods on the dataset. We do not use this definition in the main text because the test RMSE is heavy-tailed and can be infinite, so that the mean and standard deviation of the paired differences depend on how a few runs with overfit expressions are treated. For normally distributed paired differences, the share of the runs won by A is $\Phi(d_z)$, so $|d_z| \le 0.2$ corresponds to the band $42.1\%$--$57.9\%$.

All differences marked with $\dagger$ are also substantial without a tie band and with the wider band, and the local ROPE of the log test RMSE leads to similar numbers of wins, ties, and losses. For DL vs.\ BIC$_\text{SR}$ and FBF, MO-DL vs.\ MO-Len-DL, and DL vs.\ DL$_\text{diag}$ and DL$_\text{det}$, the medians are close to 50\% ($48.5\%$--$55.5\%$) and 12 to 25 of the 34 matches are ties, but with 34 datasets the posterior is too wide to establish practical equivalence (43--60\% of the probability mass in the ROPE). The test RMSE of these methods is therefore similar, but the 95\% HPD intervals extend 11 to 16 percentage points to both sides of 50\%. The largest deviation from 50\% is $P(\text{DL} > \text{NLL + func.compl.}) = 37.8\%$ (10\% of the probability mass in the ROPE): without parameter complexity, the test RMSE is slightly lower for many datasets but much higher for the three datasets on which overfit expressions are selected (\Cref{sec:appendix-ablation}). Without a tie band, these small but consistent differences also count as wins ($P(\text{DL} > \text{DL}_\text{diag}) = 66.6\%$, $P(\text{DL} > \text{NLL + func.compl.}) = 30.9\%^\dagger$). The median relative differences of the test RMSE with bootstrap confidence intervals over the datasets in \Cref{tab:rmse-rel} lead to the same conclusions.

\begin{table}[ht]
\centering
\footnotesize
\setlength{\tabcolsep}{3pt}

\caption{BBT models with one match per dataset (34 matches per pair). For the tie definition of the main text (A better in $45\%$--$55\%$ of the runs): the numbers of datasets won, tied, and lost by A (W/T/L), the median and 95\% HPD of $P(A > B)$, and the probability mass in the ROPE (in percent). Medians of $P(A > B)$ without tie band ($50\%$) and with the band $42.1\%$--$57.9\%$, and W/T/L with ties defined by $|d_z| \le 0.2$ for the log test RMSE~\citep{Wainer2023}. $\dagger$: more than 99\% of the probability mass outside the ROPE.}
\label{tab:bbt-sensitivity}
\begin{tabular}{l|cccc|cc|c}
& \multicolumn{4}{c|}{Tie: $45\%-55\%$} & \multicolumn{2}{c|}{Median, tie:} & $|d_z| \le 0.2$, \\
Pairing & W/T/L & Median & 95\% HPD & in ROPE & $50\%$ & $42.1\%-57.9\%$ & log RMSE \\
\hline
\multicolumn{8}{l}{\emph{MO-Len, test RMSE}} \\
$P(\text{DL}>\text{AIC})$ & 31/2/1 & $94.2^\dagger$ & $90.0-97.3$ & 0 & $95.3^\dagger$ & $93.2^\dagger$ & 30/3/1 \\
$P(\text{DL}>\text{BIC})$ & 26/4/4 & $80.2^\dagger$ & $71.4-87.6$ & 0 & $81.4^\dagger$ & $78.1^\dagger$ & 24/6/4 \\
$P(\text{DL}>\text{BIC$_\text{SR}$})$ & 9/19/6 & $53.1$ & $41.6-64.9$ & 53 & $51.6$ & $52.2$ & 8/19/7 \\
$P(\text{DL}>\text{FBF})$ & 9/17/8 & $50.9$ & $39.0-62.8$ & 60 & $50.1$ & $51.5$ & 7/20/7 \\
\multicolumn{8}{l}{\emph{SO vs.\ MO-Len-DL, test RMSE}} \\
$P(\text{MO-Len-DL}>\text{SO-AIC})$ & 28/4/2 & $85.2^\dagger$ & $78.8-90.6$ & 0 & $85.4^\dagger$ & $86.7^\dagger$ & 29/3/2 \\
$P(\text{MO-Len-DL}>\text{SO-BIC})$ & 20/6/8 & $61.4$ & $50.9-71.1$ & 12 & $59.5$ & $62.2$ & 18/9/7 \\
$P(\text{MO-Len-DL}>\text{SO-BIC$_\text{SR}$})$ & 27/3/4 & $81.3^\dagger$ & $74.3-87.8$ & 0 & $79.8^\dagger$ & $83.0^\dagger$ & 26/4/4 \\
$P(\text{MO-Len-DL}>\text{SO-FBF})$ & 25/3/6 & $83.6^\dagger$ & $76.9-89.6$ & 0 & $83.0^\dagger$ & $84.2^\dagger$ & 26/4/4 \\
$P(\text{MO-Len-DL}>\text{SO-DL})$ & 26/3/5 & $78.3^\dagger$ & $70.4-85.4$ & 0 & $76.2^\dagger$ & $79.5^\dagger$ & 23/8/3 \\
\multicolumn{8}{l}{\emph{MO-DL vs.\ MO-Len-DL, test RMSE}} \\
$P(\text{MO-DL}>\text{MO-Len-DL})$ & 11/12/11 & $49.9$ & $34.4-65.8$ & 47 & $52.9$ & $47.5$ & 5/20/9 \\
\multicolumn{8}{l}{\emph{Ablation, test RMSE}} \\
$P(\text{DL}>\text{DL$_\text{diag}$})$ & 7/24/3 & $55.5$ & $44.4-66.4$ & 43 & $66.6$ & $54.5$ & 8/24/2 \\
$P(\text{DL}>\text{DL$_\text{det}$})$ & 4/25/5 & $48.5$ & $37.9-59.9$ & 60 & $43.9$ & $49.9$ & 4/25/5 \\
$P(\text{DL}>\text{NLL + func.compl.})$ & 7/10/17 & $37.8$ & $26.9-48.2$ & 10 & $30.9^\dagger$ & $42.7$ & 4/17/13 \\
\end{tabular}

\end{table}

\subsubsection{Program length of selected expressions}\label{sec:appendix-program-length}
\begin{figure}[h!]
  \centering
  \includegraphics[page=1]{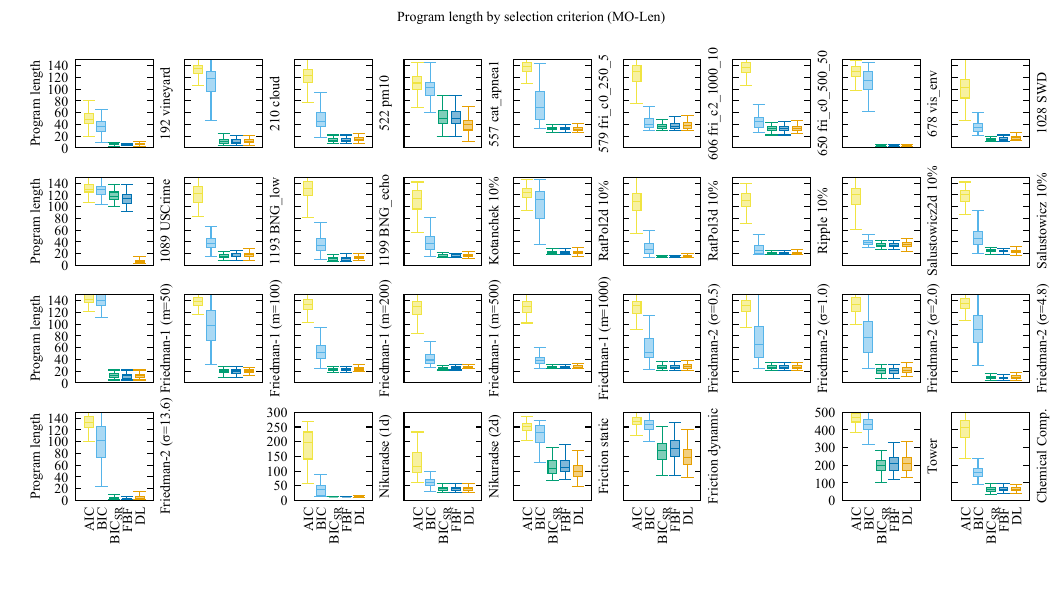}
  \caption{Boxplots of the selected program lengths for all criteria and all datasets for MO-Len. BIC$_\text{SR}$, DL, and FBF select the shortest expressions, AIC selects the longest expressions.}
  \label{fig:boxplot-size}
\end{figure}

\Cref{fig:boxplot-size} shows program length statistics for all datasets for MO-Len and all selection criteria. The key insight is that the criteria that include the function complexity penalty (BIC$_\text{SR}$, DL, FBF) select similar models, with the smallest sizes. BIC selects longer expressions, consistent with its definition.

\subsubsection{Test $R^2$ and test RMSE}
\Cref{tab:MO-Len-r2} shows the median test $R^2$ of the expressions selected from the MO-Len Pareto fronts with each criterion for all datasets, together with the BBT win probabilities of DL against each criterion. \Cref{tab:rmse-mo} shows the 10\%, 50\%, and 90\% quantiles of the test RMSE of the expressions selected with each criterion from the MO-Len Pareto fronts and of the expressions selected by MO-DL. It complements the median test $R^2$ and the win probabilities in \Cref{tab:MO-Len-r2} with the actual values of the test RMSE and their variation over the runs. \Cref{tab:rmse-rel} summarises the differences to MO-Len-DL over all datasets for all MO and SO methods: for each dataset, we calculate the relative difference of the median test RMSE of a method to the median test RMSE of MO-Len-DL, and report the median of these differences over the 34 datasets with a 95\% bootstrap confidence interval. The bootstrap resamples the datasets, i.e., each dataset is one independent unit, so that the dependence between the runs on the same dataset and between the criteria that select from the same Pareto fronts does not lead to too narrow intervals.

The median test RMSE of BIC$_\text{SR}$ and FBF is within $\pm 5\%$ of the median test RMSE of DL for 27 of the 34 datasets, and the median difference over all datasets is $0.1\%$ for both (95\% CI: $-0.5\%$ to $0.4\%$ and $-0.4\%$ to $0.7\%$). The largest difference is for 1089 USCrime (28 training observations), for which the median test RMSE of BIC$_\text{SR}$ and FBF is about twice as high as that of DL. The median test RMSE of MO-DL is within $\pm 5\%$ of that of MO-Len-DL for 26 of the 34 datasets, with a median difference of $0.4\%$ (95\% CI: $-0.5\%$ to $1.6\%$). The largest differences are for Kotanchek ($+172\%$), Friedman-2 $\sigma=1.0$ ($+49\%$), and Salustowicz2d ($-32\%$). The median test RMSE of AIC and BIC is a median $60\%$ and $26\%$ higher than that of DL (95\% CI: $30\%$ to $116\%$ and $4\%$ to $52\%$), and the 90\% quantiles show that AIC and BIC select strongly overfit expressions in a substantial fraction of the runs for many datasets.

\begin{table}[ht]
\centering
\small
\caption{Median $R^2$ values on the test set of MO-Len results with AIC, BIC, BIC$_\text{SR}$, DL, and FBF (best marked in bold). For the synthetic problem instances the test set is without noise. The last two rows give the median posterior win probabilities of DL vs. each criterion from the BBT model. Substantially higher win probabilities outside of the ROPE $45\% - 55\%$ are marked with $\dagger$ (more than 99\% of the probability mass outside the ROPE, one match per dataset, see \Cref{sec:statistical-comparisons}). The 10\%, 50\%, and 90\% quantiles of the test RMSE are given in \Cref{tab:rmse-mo}, and the median differences to MO-Len-DL with bootstrap confidence intervals in \Cref{tab:rmse-rel}.}\label{tab:MO-Len-r2}
\begin{tabular}{l|r|r|r|r|r}
Dataset & AIC & BIC & BIC$_\text{SR}$ & FBF & DL \\
\hline
Kotanchek                        & $-\infty$ & $0.8819$ & $\mathbf{0.9955}$ & $0.9952$ & $0.9938$ \\
RatPol2d                         & $-0.3880$ & $0.4823$ & $\mathbf{0.9813}$ & $\mathbf{0.9813}$ & $0.9808$ \\
RatPol3d                         & $0.9943$ & $0.9990$ & $\mathbf{0.9998}$ & $\mathbf{0.9998}$ & $0.9997$ \\
Ripple                           & $0.9966$ & $0.9996$ & $\mathbf{0.9998}$ & $\mathbf{0.9998}$ & $\mathbf{0.9998}$ \\
Salustowicz                      & $0.9515$ & $0.9939$ & $\mathbf{0.9962}$ & $\mathbf{0.9962}$ & $\mathbf{0.9962}$ \\
Salustowicz2d                    & $0.9944$ & $\mathbf{0.9983}$ & $0.9981$ & $0.9981$ & $0.9967$ \\
Friedman-1 $m=50$                & $-0.2242$ & $-0.2242$ & $\mathbf{0.7981}$ & $0.7760$ & $0.7938$ \\
Friedman-1 $m=100$               & $0.6796$ & $0.7867$ & $\mathbf{0.9436}$ & $0.9379$ & $0.9382$ \\
Friedman-1 $m=200$               & $0.8945$ & $0.9540$ & $0.9764$ & $0.9762$ & $\mathbf{0.9773}$ \\
Friedman-1 $m=500$               & $0.9575$ & $0.9873$ & $0.9918$ & $0.9920$ & $\mathbf{0.9929}$ \\
Friedman-1 $m=1000$              & $0.9801$ & $0.9949$ & $\mathbf{0.9954}$ & $\mathbf{0.9954}$ & $0.9953$ \\
Friedman-2 $\sigma=0.5$          & $0.9825$ & $0.9935$ & $\mathbf{0.9982}$ & $\mathbf{0.9982}$ & $\mathbf{0.9982}$ \\
Friedman-2 $\sigma=1.0$          & $0.9236$ & $0.9661$ & $0.9950$ & $\mathbf{0.9952}$ & $0.9948$ \\
Friedman-2 $\sigma=2.0$          & $0.6525$ & $0.8130$ & $0.9085$ & $0.9036$ & $\mathbf{0.9092}$ \\
Friedman-2 $\sigma=4.8$          & $-1.1878$ & $-0.4464$ & $0.4840$ & $0.4716$ & $\mathbf{0.5129}$ \\
Friedman-2 $\sigma=13.6$         & $-16.1230$ & $-11.2838$ & $-0.1317$ & $\mathbf{-0.0960}$ & $-0.1375$ \\
Nikuradse (2d)                   & $0.9903$ & $\mathbf{0.9948}$ & $\mathbf{0.9948}$ & $\mathbf{0.9948}$ & $\mathbf{0.9948}$ \\
Nikuradse (1d)                   & $0.9631$ & $0.9788$ & $\mathbf{0.9820}$ & $\mathbf{0.9820}$ & $0.9819$ \\
Chem. 1 Tower                    & $\mathbf{0.9541}$ & $0.9539$ & $0.9461$ & $0.9468$ & $0.9468$ \\
Chem. 2 Comp.                    & $0.4826$ & $0.7443$ & $0.7196$ & $0.7196$ & $\mathbf{0.7486}$ \\
Friction dyn.                    & $0.9199$ & $0.9295$ & $0.9339$ & $0.9349$ & $\mathbf{0.9357}$ \\
Friction stat.                   & $0.9143$ & $\mathbf{0.9156}$ & $0.9129$ & $0.9132$ & $0.9131$ \\
192 vineyard                     & $0.0791$ & $0.2011$ & $0.5494$ & $0.5369$ & $\mathbf{0.5604}$ \\
210 cloud                        & $0.3364$ & $0.3141$ & $0.7932$ & $0.7938$ & $\mathbf{0.7949}$ \\
522 pm10                         & $0.0887$ & $0.2025$ & $0.2379$ & $0.2322$ & $\mathbf{0.2557}$ \\
557 analcat apnea1               & $0.6919$ & $0.6903$ & $0.7729$ & $0.7691$ & $\mathbf{0.7803}$ \\
579 fri c0 250 5                 & $0.9166$ & $0.9372$ & $\mathbf{0.9490}$ & $\mathbf{0.9490}$ & $0.9467$ \\
606 fri c2 1000 10               & $0.9748$ & $0.9781$ & $\mathbf{0.9782}$ & $\mathbf{0.9782}$ & $0.9778$ \\
650 fri c0 500 50                & $0.9425$ & $0.9551$ & $0.9574$ & $\mathbf{0.9575}$ & $0.9562$ \\
678 vis. env.                    & $-0.5758$ & $-0.5235$ & $0.2936$ & $0.2963$ & $\mathbf{0.2971}$ \\
1028 SWD                         & $0.3658$ & $\mathbf{0.3843}$ & $0.3594$ & $0.3604$ & $0.3654$ \\
1089 USCrime                     & $-0.1414$ & $-0.0827$ & $-0.2915$ & $0.0020$ & $\mathbf{0.7212}$ \\
1193 BNG lowbwt                  & $0.5621$ & $\mathbf{0.5766}$ & $0.5683$ & $0.5689$ & $0.5699$ \\
1199 BNG echoMon                 & $0.3523$ & $0.4187$ & $0.4237$ & $0.4238$ & $\mathbf{0.4255}$ \\
\hline
Win prob. & DL$>$AIC & DL$>$BIC & DL$>$BIC$_\text{SR}$ & DL$>$FBF & \\
($34$ datasets) & $94.2\%^\dagger$ & $80.2\%^\dagger$ & $53.1\%$ & $50.9\%$ & \\
Num. best median $R^2$ & 1 & 5 & 13 & 13 & 17 \\
\end{tabular}
\end{table}

\begin{sidewaystable}
  \centering
  \footnotesize
  \setlength{\tabcolsep}{2.5pt}
  
  \caption{Quantiles of the test RMSE of the expressions selected from the MO-Len Pareto fronts with AIC, BIC, BIC$_\text{SR}$, FBF, and DL, and of the expressions selected with DL from the final MO-DL Pareto fronts. P10, P50, and P90 are the 10\%, 50\% (median), and 90\% quantiles over the 100 runs; the lowest median is marked bold. For the synthetic datasets the test set is without noise. A non-finite test RMSE (e.g., caused by a pole of the expression within the range of the test inputs) counts as $\infty$.}
  \label{tab:rmse-mo}
  \begin{tabular}{l|rrr|rrr|rrr|rrr|rrr|rrr}
Dataset & \multicolumn{3}{c|}{MO-Len-AIC} & \multicolumn{3}{c|}{MO-Len-BIC} & \multicolumn{3}{c|}{MO-Len-BIC$_\text{SR}$} & \multicolumn{3}{c|}{MO-Len-FBF} & \multicolumn{3}{c|}{MO-Len-DL} & \multicolumn{3}{c}{MO-DL} \\
 & P10 & P50 & P90 & P10 & P50 & P90 & P10 & P50 & P90 & P10 & P50 & P90 & P10 & P50 & P90 & P10 & P50 & P90 \\
\hline
Kotanchek                        & $0.0482$ & $\infty$ & $\infty$ & $0.0126$ & $0.0700$ & $\infty$ & $0.00755$ & $\mathbf{0.0137}$ & $0.0736$ & $0.00755$ & $0.0141$ & $0.0739$ & $0.00758$ & $0.0160$ & $0.0735$ & $0.0102$ & $0.0435$ & $\infty$ \\
RatPol2d                         & $0.668$ & $2.99$ & $3.3\cdot 10^{9}$ & $0.515$ & $1.83$ & $5.9\cdot 10^{8}$ & $0.229$ & $\mathbf{0.347}$ & $0.828$ & $0.229$ & $\mathbf{0.347}$ & $0.807$ & $0.218$ & $0.352$ & $0.827$ & $0.240$ & $0.413$ & $1.03$ \\
RatPol3d                         & $0.0408$ & $0.0630$ & $17.6$ & $0.0119$ & $0.0261$ & $0.0927$ & $0.00622$ & $\mathbf{0.0117}$ & $0.0362$ & $0.00622$ & $\mathbf{0.0117}$ & $0.0362$ & $0.00620$ & $0.0135$ & $0.0379$ & $0.00712$ & $0.0139$ & $0.0260$ \\
Ripple                           & $0.180$ & $0.233$ & $0.833$ & $0.0340$ & $0.0760$ & $0.172$ & $0.0306$ & $\mathbf{0.0533}$ & $0.0771$ & $0.0306$ & $\mathbf{0.0533}$ & $0.0771$ & $0.0306$ & $0.0537$ & $0.0862$ & $0.0424$ & $0.0613$ & $0.563$ \\
Salustowicz                      & $0.0273$ & $0.0668$ & $\infty$ & $0.0160$ & $0.0237$ & $0.252$ & $0.0137$ & $0.0187$ & $0.0279$ & $0.0137$ & $0.0187$ & $0.0279$ & $0.0137$ & $0.0187$ & $0.0308$ & $0.0126$ & $\mathbf{0.0179}$ & $0.0253$ \\
Salustowicz2d                    & $0.0447$ & $0.0753$ & $\infty$ & $0.0253$ & $0.0411$ & $0.124$ & $0.0293$ & $0.0433$ & $0.0944$ & $0.0282$ & $0.0434$ & $0.101$ & $0.0295$ & $0.0576$ & $0.144$ & $0.0238$ & $\mathbf{0.0393}$ & $0.0835$ \\
Friedman-1 $m=50$                & $1.43$ & $2.76$ & $2.2\cdot 10^{5}$ & $1.43$ & $2.76$ & $1.6\cdot 10^{6}$ & $0.851$ & $1.12$ & $1.53$ & $0.940$ & $1.18$ & $1.54$ & $0.879$ & $1.13$ & $1.52$ & $0.793$ & $\mathbf{1.08}$ & $1.43$ \\
Friedman-1 $m=100$               & $1.05$ & $1.41$ & $73.0$ & $0.757$ & $1.15$ & $13.8$ & $0.438$ & $\mathbf{0.593}$ & $0.942$ & $0.451$ & $0.622$ & $0.951$ & $0.436$ & $0.620$ & $0.921$ & $0.409$ & $\mathbf{0.593}$ & $0.975$ \\
Friedman-1 $m=200$               & $0.666$ & $0.811$ & $18.9$ & $0.366$ & $0.535$ & $0.810$ & $0.268$ & $0.384$ & $0.497$ & $0.277$ & $0.385$ & $0.497$ & $0.265$ & $0.376$ & $0.481$ & $0.247$ & $\mathbf{0.370}$ & $0.455$ \\
Friedman-1 $m=500$               & $0.425$ & $0.514$ & $2.90$ & $0.178$ & $0.281$ & $0.418$ & $0.176$ & $0.226$ & $0.308$ & $0.174$ & $0.223$ & $0.308$ & $0.170$ & $\mathbf{0.210}$ & $0.304$ & $0.182$ & $0.219$ & $0.319$ \\
Friedman-1 $m=1000$              & $0.290$ & $0.352$ & $0.398$ & $0.108$ & $0.177$ & $0.239$ & $0.145$ & $\mathbf{0.169}$ & $0.199$ & $0.145$ & $\mathbf{0.169}$ & $0.197$ & $0.142$ & $0.171$ & $0.199$ & $0.144$ & $0.174$ & $0.201$ \\
Friedman-2 $\sigma=0.5$          & $0.444$ & $0.673$ & $3.50$ & $0.238$ & $0.412$ & $0.704$ & $0.113$ & $\mathbf{0.216}$ & $0.310$ & $0.112$ & $\mathbf{0.216}$ & $0.307$ & $0.110$ & $\mathbf{0.216}$ & $0.286$ & $0.173$ & $0.249$ & $1.52$ \\
Friedman-2 $\sigma=1.0$          & $0.942$ & $1.41$ & $16.7$ & $0.510$ & $0.938$ & $3.02$ & $0.224$ & $0.360$ & $0.715$ & $0.224$ & $\mathbf{0.354}$ & $0.715$ & $0.232$ & $0.366$ & $0.900$ & $0.283$ & $0.544$ & $1.72$ \\
Friedman-2 $\sigma=2.0$          & $1.96$ & $3.00$ & $290$ & $1.17$ & $2.20$ & $27.9$ & $0.515$ & $1.54$ & $2.64$ & $0.519$ & $1.58$ & $2.68$ & $0.532$ & $\mathbf{1.53}$ & $2.62$ & $0.810$ & $1.71$ & $2.62$ \\
Friedman-2 $\sigma=4.8$          & $5.57$ & $7.53$ & $17108$ & $4.64$ & $6.12$ & $111$ & $2.97$ & $3.66$ & $4.40$ & $2.92$ & $3.70$ & $4.45$ & $2.86$ & $3.55$ & $4.26$ & $3.03$ & $\mathbf{3.42}$ & $4.46$ \\
Friedman-2 $\sigma=13.6$         & $13.9$ & $21.1$ & $2961$ & $12.0$ & $17.8$ & $426$ & $4.76$ & $5.42$ & $7.99$ & $4.74$ & $\mathbf{5.33}$ & $7.99$ & $4.71$ & $5.43$ & $7.74$ & $4.61$ & $5.39$ & $7.93$ \\
Nikuradse (2d)                   & $0.00670$ & $0.00982$ & $0.118$ & $0.00649$ & $0.00719$ & $0.0279$ & $0.00645$ & $0.00718$ & $0.0101$ & $0.00648$ & $0.00718$ & $0.00931$ & $0.00640$ & $\mathbf{0.00715}$ & $0.00932$ & $0.00637$ & $0.00718$ & $0.00968$ \\
Nikuradse (1d)                   & $0.0440$ & $0.0525$ & $0.0893$ & $0.0367$ & $0.0398$ & $0.0457$ & $0.0366$ & $\mathbf{0.0367}$ & $0.0382$ & $0.0366$ & $\mathbf{0.0367}$ & $0.0382$ & $0.0366$ & $0.0368$ & $0.0438$ & $0.0365$ & $0.0370$ & $0.0388$ \\
Chem. 1 Tower                    & $17.1$ & $\mathbf{17.8}$ & $18.3$ & $17.2$ & $17.9$ & $18.4$ & $18.4$ & $19.3$ & $20.4$ & $18.3$ & $19.2$ & $20.4$ & $18.4$ & $19.2$ & $20.4$ & $18.6$ & $19.7$ & $20.7$ \\
Chem. 2 Comp.                    & $0.155$ & $0.269$ & $1.31$ & $0.135$ & $0.189$ & $0.600$ & $0.142$ & $0.198$ & $0.313$ & $0.143$ & $0.198$ & $0.320$ & $0.144$ & $0.188$ & $0.300$ & $0.139$ & $\mathbf{0.184}$ & $0.431$ \\
Friction dyn.                    & $0.00245$ & $0.00337$ & $0.00722$ & $0.00248$ & $0.00316$ & $0.00593$ & $0.00248$ & $0.00306$ & $0.00546$ & $0.00245$ & $0.00304$ & $0.00541$ & $0.00254$ & $\mathbf{0.00302}$ & $0.00424$ & $0.00229$ & $0.00305$ & $0.00452$ \\
Friction stat.                   & $0.00329$ & $0.00361$ & $0.00430$ & $0.00330$ & $\mathbf{0.00358}$ & $0.00431$ & $0.00327$ & $0.00364$ & $0.00413$ & $0.00325$ & $0.00363$ & $0.00408$ & $0.00326$ & $0.00364$ & $0.00407$ & $0.00330$ & $0.00361$ & $0.00425$ \\
192 vineyard                     & $2.94$ & $3.55$ & $13.2$ & $2.67$ & $3.43$ & $6.95$ & $2.24$ & $2.82$ & $3.28$ & $2.23$ & $2.88$ & $3.78$ & $2.19$ & $2.74$ & $3.68$ & $2.09$ & $\mathbf{2.66}$ & $3.15$ \\
210 cloud                        & $0.576$ & $0.862$ & $1367$ & $0.508$ & $0.911$ & $156$ & $0.303$ & $0.517$ & $0.678$ & $0.303$ & $0.520$ & $0.722$ & $0.305$ & $\mathbf{0.507}$ & $0.648$ & $0.316$ & $0.530$ & $20.0$ \\
522 pm10                         & $0.762$ & $0.838$ & $1.07$ & $0.724$ & $0.788$ & $0.859$ & $0.715$ & $0.771$ & $0.827$ & $0.715$ & $0.772$ & $0.823$ & $0.712$ & $0.766$ & $0.821$ & $0.718$ & $\mathbf{0.763}$ & $0.816$ \\
557 analcat apnea1               & $1051$ & $1756$ & $6499$ & $1042$ & $1648$ & $5046$ & $1107$ & $1497$ & $2564$ & $1114$ & $1506$ & $2564$ & $988$ & $1478$ & $2029$ & $979$ & $\mathbf{1372}$ & $2334$ \\
579 fri c0 250 5                 & $0.255$ & $0.286$ & $0.323$ & $0.219$ & $0.251$ & $0.296$ & $0.206$ & $\mathbf{0.222}$ & $0.249$ & $0.206$ & $0.223$ & $0.249$ & $0.209$ & $0.227$ & $0.256$ & $0.204$ & $0.228$ & $0.253$ \\
606 fri c2 1000 10               & $0.151$ & $0.157$ & $0.167$ & $0.143$ & $0.148$ & $0.154$ & $0.142$ & $0.148$ & $0.154$ & $0.142$ & $\mathbf{0.147}$ & $0.155$ & $0.143$ & $0.148$ & $0.160$ & $0.144$ & $0.150$ & $0.157$ \\
650 fri c0 500 50                & $0.224$ & $0.242$ & $0.276$ & $0.197$ & $0.212$ & $0.234$ & $0.194$ & $\mathbf{0.207}$ & $0.225$ & $0.194$ & $\mathbf{0.207}$ & $0.225$ & $0.197$ & $0.210$ & $0.238$ & $0.197$ & $0.209$ & $0.235$ \\
678 vis. env.                    & $3.55$ & $4.30$ & $10.2$ & $3.44$ & $4.34$ & $6.80$ & $2.52$ & $2.87$ & $3.38$ & $2.51$ & $\mathbf{2.86}$ & $3.29$ & $2.51$ & $\mathbf{2.86}$ & $3.32$ & $2.54$ & $2.90$ & $3.39$ \\
1028 SWD                         & $0.621$ & $0.646$ & $0.667$ & $0.615$ & $\mathbf{0.639}$ & $0.662$ & $0.626$ & $0.646$ & $0.672$ & $0.625$ & $0.646$ & $0.672$ & $0.623$ & $0.645$ & $0.669$ & $0.620$ & $0.642$ & $0.664$ \\
1089 USCrime                     & $26.7$ & $40.0$ & $114$ & $26.7$ & $39.2$ & $108$ & $22.4$ & $41.8$ & $113$ & $19.9$ & $37.8$ & $113$ & $16.3$ & $\mathbf{20.4}$ & $29.0$ & $17.1$ & $21.1$ & $32.4$ \\
1193 BNG lowbwt                  & $471$ & $479$ & $491$ & $467$ & $\mathbf{471}$ & $478$ & $473$ & $476$ & $481$ & $472$ & $476$ & $480$ & $471$ & $475$ & $480$ & $469$ & $473$ & $476$ \\
1199 BNG echoMon                 & $12.4$ & $12.7$ & $86.1$ & $11.9$ & $\mathbf{12.0}$ & $12.5$ & $11.9$ & $\mathbf{12.0}$ & $12.1$ & $11.9$ & $\mathbf{12.0}$ & $12.1$ & $11.9$ & $\mathbf{12.0}$ & $12.1$ & $11.9$ & $\mathbf{12.0}$ & $12.0$ \\
  \end{tabular}
  
\end{sidewaystable}

\begin{table}[ht]
  \centering
  
  \caption{Median test RMSE of all methods relative to MO-Len-DL. For each dataset, we calculate the relative difference of the median test RMSE over the 100 runs to the median test RMSE of MO-Len-DL. The table shows the median of these differences over the 34 datasets in percent with a 95\% percentile bootstrap confidence interval (10{,}000 resamples of the datasets), and the number of datasets for which the median test RMSE is higher or lower than that of MO-Len-DL. Positive differences mean a higher test RMSE than MO-Len-DL.}
  \label{tab:rmse-rel}
  \begin{tabular}{l|r|c|r|r}
Method & Median diff. \% & 95\% CI & Higher & Lower \\
\hline
MO-Len-AIC                     & $60.1$ & $[29.8, 115.5]$ & 32 & 2 \\
MO-Len-BIC                     & $25.9$ & $[3.7, 51.6]$ & 28 & 6 \\
MO-Len-BIC$_\text{SR}$         & $0.1$ & $[-0.5, 0.4]$ & 19 & 15 \\
MO-Len-FBF                     & $0.1$ & $[-0.4, 0.7]$ & 19 & 15 \\
MO-DL                          & $0.4$ & $[-0.5, 1.6]$ & 18 & 16 \\
SO-AIC                         & $54.7$ & $[11.1, 97.2]$ & 30 & 4 \\
SO-BIC                         & $4.4$ & $[0.2, 14.9]$ & 23 & 11 \\
SO-BIC$_\text{SR}$             & $14.7$ & $[2.7, 50.1]$ & 28 & 6 \\
SO-FBF                         & $14.8$ & $[3.2, 54.4]$ & 28 & 6 \\
SO-DL                          & $16.8$ & $[0.9, 64.6]$ & 27 & 7 \\
  \end{tabular}
  
\end{table}

\subsubsection{Line charts for all datasets}\label{sec:appendix-multi-objective-linecharts}
\noindent
\begin{center}
\includegraphics[page=1]{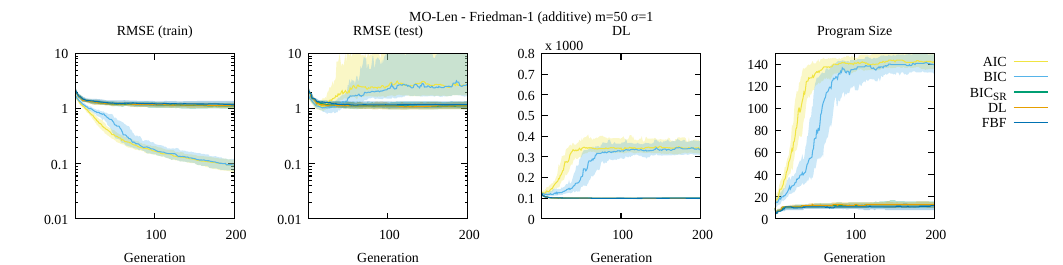}\\
\includegraphics[page=2]{neogp_linecharts_molength.pdf}\\
\includegraphics[page=3]{neogp_linecharts_molength.pdf}\\
\includegraphics[page=4]{neogp_linecharts_molength.pdf}\\
\includegraphics[page=5]{neogp_linecharts_molength.pdf}\\
\includegraphics[page=6]{neogp_linecharts_molength.pdf}\\
\includegraphics[page=7]{neogp_linecharts_molength.pdf}\\
\includegraphics[page=8]{neogp_linecharts_molength.pdf}\\
\includegraphics[page=9]{neogp_linecharts_molength.pdf}\\
\includegraphics[page=10]{neogp_linecharts_molength.pdf}\\
\includegraphics[page=11]{neogp_linecharts_molength.pdf}\\
\includegraphics[page=12]{neogp_linecharts_molength.pdf}\\
\includegraphics[page=13]{neogp_linecharts_molength.pdf}\\
\includegraphics[page=14]{neogp_linecharts_molength.pdf}\\
\includegraphics[page=15]{neogp_linecharts_molength.pdf}\\
\includegraphics[page=16]{neogp_linecharts_molength.pdf}\\
\includegraphics[page=17]{neogp_linecharts_molength.pdf}\\
\includegraphics[page=18]{neogp_linecharts_molength.pdf}\\
\includegraphics[page=19]{neogp_linecharts_molength.pdf}\\
\includegraphics[page=20]{neogp_linecharts_molength.pdf}\\
\includegraphics[page=21]{neogp_linecharts_molength.pdf}\\
\includegraphics[page=22]{neogp_linecharts_molength.pdf}\\
\includegraphics[page=23]{neogp_linecharts_molength.pdf}\\
\includegraphics[page=24]{neogp_linecharts_molength.pdf}\\
\includegraphics[page=25]{neogp_linecharts_molength.pdf}\\
\includegraphics[page=26]{neogp_linecharts_molength.pdf}\\
\includegraphics[page=27]{neogp_linecharts_molength.pdf}\\
\includegraphics[page=28]{neogp_linecharts_molength.pdf}\\
\includegraphics[page=29]{neogp_linecharts_molength.pdf}\\
\includegraphics[page=30]{neogp_linecharts_molength.pdf}\\
\includegraphics[page=31]{neogp_linecharts_molength.pdf}\\
\includegraphics[page=32]{neogp_linecharts_molength.pdf}\\
\includegraphics[page=33]{neogp_linecharts_molength.pdf}\\
\includegraphics[page=34]{neogp_linecharts_molength.pdf}\\
\end{center}

\subsubsection{Pareto fronts and selection criteria for all datasets}
\begin{center}
\includegraphics[page=1]{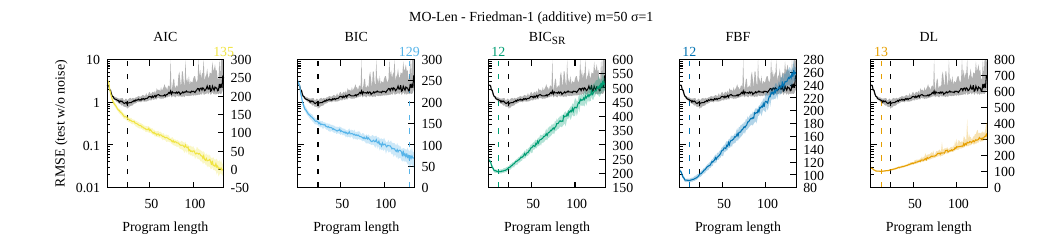}\\
\includegraphics[page=2]{neogp_pareto_fronts.pdf}\\
\includegraphics[page=3]{neogp_pareto_fronts.pdf}\\
\includegraphics[page=4]{neogp_pareto_fronts.pdf}\\
\includegraphics[page=5]{neogp_pareto_fronts.pdf}\\
\includegraphics[page=6]{neogp_pareto_fronts.pdf}\\
\includegraphics[page=7]{neogp_pareto_fronts.pdf}\\
\includegraphics[page=8]{neogp_pareto_fronts.pdf}\\
\includegraphics[page=9]{neogp_pareto_fronts.pdf}\\
\includegraphics[page=10]{neogp_pareto_fronts.pdf}\\
\includegraphics[page=11]{neogp_pareto_fronts.pdf}\\
\includegraphics[page=12]{neogp_pareto_fronts.pdf}\\
\includegraphics[page=13]{neogp_pareto_fronts.pdf}\\
\includegraphics[page=14]{neogp_pareto_fronts.pdf}\\
\includegraphics[page=15]{neogp_pareto_fronts.pdf}\\
\includegraphics[page=16]{neogp_pareto_fronts.pdf}\\
\includegraphics[page=17]{neogp_pareto_fronts.pdf}\\
\includegraphics[page=18]{neogp_pareto_fronts.pdf}\\
\includegraphics[page=19]{neogp_pareto_fronts.pdf}\\
\includegraphics[page=20]{neogp_pareto_fronts.pdf}\\
\includegraphics[page=21]{neogp_pareto_fronts.pdf}\\
\includegraphics[page=22]{neogp_pareto_fronts.pdf}\\
\includegraphics[page=23]{neogp_pareto_fronts.pdf}\\
\includegraphics[page=24]{neogp_pareto_fronts.pdf}\\
\includegraphics[page=25]{neogp_pareto_fronts.pdf}\\
\includegraphics[page=26]{neogp_pareto_fronts.pdf}\\
\includegraphics[page=27]{neogp_pareto_fronts.pdf}\\
\includegraphics[page=28]{neogp_pareto_fronts.pdf}\\
\includegraphics[page=29]{neogp_pareto_fronts.pdf}\\
\includegraphics[page=30]{neogp_pareto_fronts.pdf}\\
\includegraphics[page=31]{neogp_pareto_fronts.pdf}\\
\includegraphics[page=32]{neogp_pareto_fronts.pdf}\\
\includegraphics[page=33]{neogp_pareto_fronts.pdf}\\
\includegraphics[page=34]{neogp_pareto_fronts.pdf}\\
\end{center}

\subsubsection{MO-Len with BIC$_\text{SR}$ vs. DL vs. FBF}
\label{sec:boxplot-dl-vs-fbf}
\begin{figure}[h!]
  \center
  \includegraphics[]{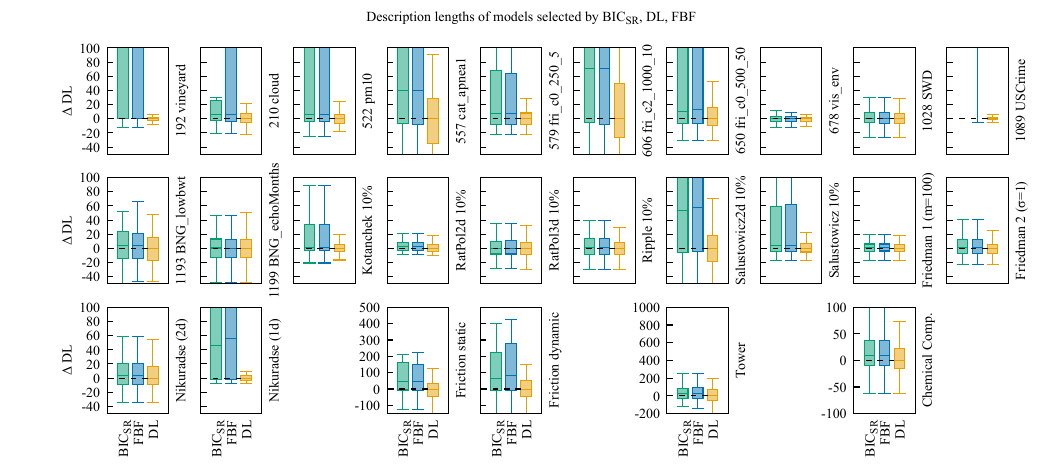}
  \caption{Boxplots of the differences of the DL of expressions selected by BIC$_\text{SR}$, FBF, and DL with MO-Len. Values are centred on the median DL found by MO-Len-DL for each dataset. }
  \label{fig:boxplot-dl-vs-fbf}
\end{figure}

\Cref{fig:boxplot-dl-vs-fbf} shows a boxplot of the DL values of expressions found by MO-Len with BIC$_\text{SR}$, DL, and FBF model selection.
MO-Len-DL finds expressions with shortest DL ($P(\text{MO-Len-DL} > \text{MO-Len-FBF}) = 99.0~\%^\dagger$, $P(\text{MO-Len-DL} > \text{MO-Len-BIC}_\text{SR}) = 99.1~\%^\dagger$; MO-Len-DL wins on all 34 datasets). MO-Len-FBF and MO-Len-BIC$_\text{SR}$ select the same expression in 90\% of the runs  ($P(\text{MO-Len-FBF} > \text{MO-Len-BIC}_\text{SR}) = 52.9~\%$; 32 of the 34 matches are ties).
Since BIC$_\text{SR}$ and FBF selection do not penalise large parameter complexities resulting from stiff parameters, these selection criteria select expressions with longer DL. Nevertheless, the test RMSE values of selected expressions are similar .

\subsubsection{Statistics on selected test RMSE relative to minimal test RMSE of models in the Pareto front}

\begin{figure}[h!]
  \center
  \includegraphics[]{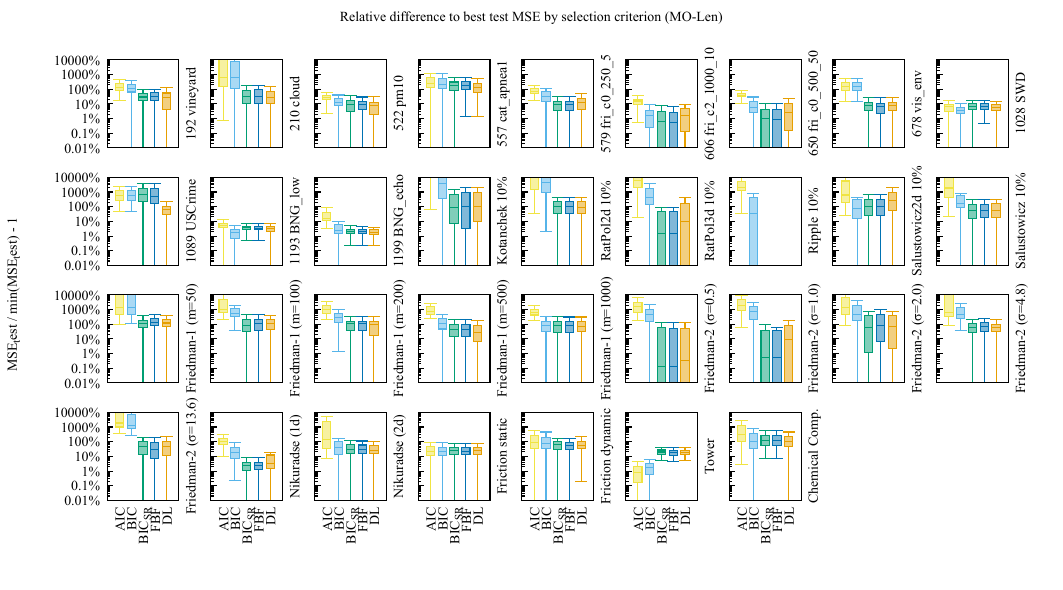}
  \caption{Boxplots of the RMSE on test sets of the expressions found with MO-Len selected by each of the criteria for all datasets. The values are relative to the best test RMSE found in each of the final Pareto fronts. BIC$_\text{SR}$, DL, and FBF produce similar results, BIC is often worse. AIC is worst (except for the Tower dataset).}
  \label{fig:boxplot-test-error}
\end{figure}

\Cref{fig:boxplot-test-error} shows boxplots for the distributions of test RMSE of the expressions found by MO-Len with each model selection criterion. The values are shown relative to the best test RMSE in each Pareto front, a value of 1\% means that the selected expression has a 1\% worse RMSE than the minimal RMSE value in the Pareto front. The best possible value is 0\% when the minimum is selected. The selection is from the same Pareto fronts, only the criterion by which an expression is selected is changed. The plot visualises how much worse a selection criterion is compared to an oracle that would be able to select the expression with best test RMSE. The boxplots for DL and FBF are similar for most datasets. BIC is sometimes worse but also sometimes better than DL.

\clearpage

\subsection{Multi-objective optimisation of DL (MO-DL) Results}\label{sec:appendix-mo-dl}
\begin{figure}[h!]
  \center
  \includegraphics[page=1]{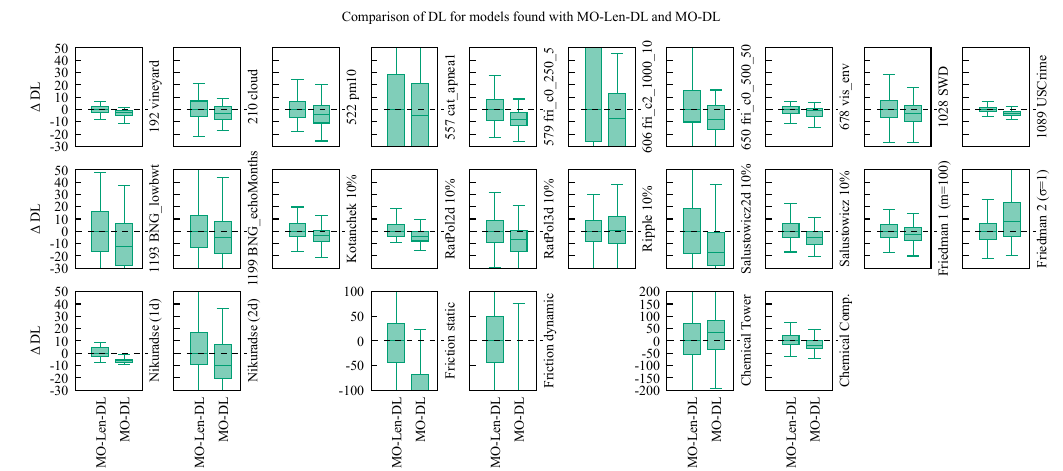}
  \caption{Boxplots of the DL of selected expressions in the final generation when using length vs. complexity as the second objective (MO-Len vs. MO-DL). $\Delta$ DL is the difference of expression DL and median DL for the MO-Len runs. Shorter expressions have negative $\Delta$ DL. MO-DL produces final solutions with better DL ($P(\text{MO-DL}>\text{MO-Len-DL})=86.2~\%^\dagger$).}
  \label{fig:boxplot-dl-mo-vs-modl}
\end{figure}

\begin{figure}[h!]
  \center
  \includegraphics[page=2]{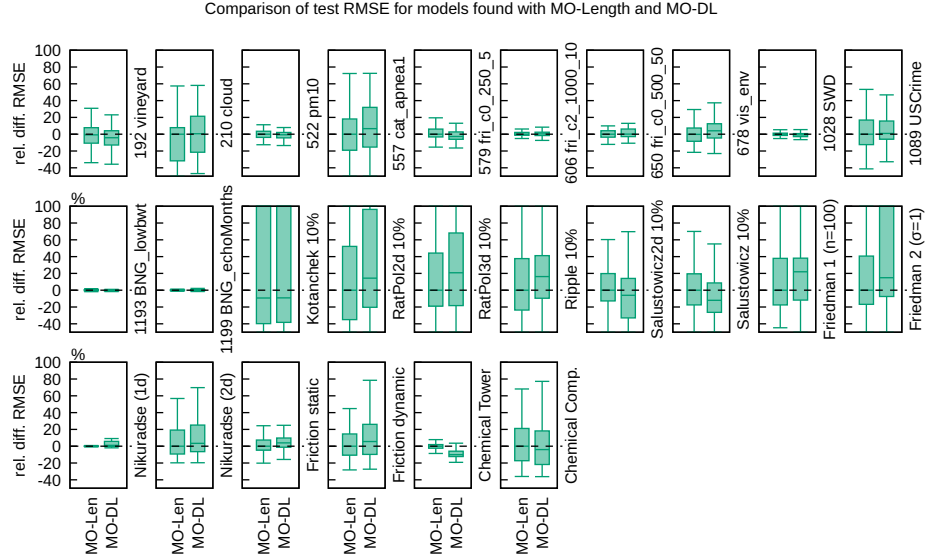}
  \caption{Boxplots of relative difference in test RMSE of MDL expressions found with MO-Len-DL and MO-DL. The reference value is the median test RMSE for MO-Len. Y-axis shows the relative difference to the reference value in percent. The test RMSE of expressions produced by MO-Len-DL and MO-DL is similar  ($P(\text{MO-DL}>\text{MO-Len-DL})=49.9~\%$ HPD: $34.4~\% - 65.8~\%$).}
  \label{fig:boxplot-rmse-mo-vs-modl}
\end{figure}

\Cref{fig:boxplot-dl-mo-vs-modl} compares the DL of the expressions selected in the final generation by MO-Len-DL and MO-DL. MO-DL yields a smaller DL ($P(\text{MO-DL}>\text{MO-Len-DL})=86.2~\%^\dagger$). \Cref{fig:boxplot-rmse-mo-vs-modl} compares the test RMSE of the same expressions.

To analyse why the lower DL of MO-DL does not lead to a lower test RMSE, we paired the expressions selected by MO-DL and by MO-Len-DL in the same run, i.e., for the same training set and random seed. Pairs with a non-finite DL or test RMSE are excluded, which leaves 3381 of the 3400 pairs. \Cref{tab:modl-dl-bins} groups the pairs by the difference $\Delta\text{DL} = \text{DL}(\text{MO-DL}) - \text{DL}(\text{MO-Len-DL})$. MO-DL finds the expression with the lower DL in 78\% of the pairs. For small differences ($-5 < \Delta\text{DL} \le 2$ nats, 43\% of the pairs), the NLL is practically the same, the difference in DL comes from the penalty, and MO-DL has the lower test RMSE in only 48\% to 54\% of the pairs. If MO-DL finds an expression with a substantially lower DL ($\Delta\text{DL} \le -5$ nats, 40\% of the pairs), it has the lower test RMSE in 55\% to 62\% of the pairs. In the remaining 17\% of the pairs ($\Delta\text{DL} > 2$ nats), MO-DL ends with a higher DL than MO-Len-DL and a higher NLL, and has the lower test RMSE in only 17\% to 33\% of the pairs. Over all pairs these effects cancel (49\%). The Spearman rank correlation of $\Delta$DL with the difference in log test RMSE is $0.25$. For the NLL part of $\Delta$DL it is $0.26$, and for the penalty part (function and parameter complexity) it is $-0.06$. A lower penalty at the same fit on the training set is therefore not related to a lower test error.

\begin{table}[h!]
  \centering
  
  \caption{Test error of the expressions selected by MO-DL and MO-Len-DL in the same run, grouped by $\Delta\text{DL} = \text{DL}(\text{MO-DL}) - \text{DL}(\text{MO-Len-DL})$ in nats. $\Delta$NLL and $\Delta$penalty are the median differences of the NLL and of the penalty (function and parameter complexity) in nats. The last two columns give the percentage of pairs in which MO-DL has the lower test RMSE and the median relative difference of the test RMSE of MO-DL to that of MO-Len-DL.}
  \label{tab:modl-dl-bins}
  \begin{tabular}{l|r|r|r|r|r}
    $\Delta$DL & Pairs & $\Delta$NLL & $\Delta$penalty & MO-DL lower RMSE$_\text{test}$ (\%) & $\Delta$RMSE$_\text{test}$ (\%) \\
    \hline
$(-\infty, -50]$         & 282 & $-125.2$ & $1.4$ & $62$ & $-3.5$ \\
$(-50, -20]$             & 222 & $-5.0$ & $-22.4$ & $57$ & $-1.2$ \\
$(-20, -10]$             & 309 & $-1.1$ & $-11.3$ & $60$ & $-0.7$ \\
$(-10, -5]$              & 552 & $-0.4$ & $-6.3$ & $55$ & $-0.1$ \\
$(-5, -2]$               & 723 & $0.0$ & $-2.7$ & $54$ & $-0.1$ \\
$(-2, -0.5]$             & 435 & $0.0$ & $-1.1$ & $49$ & $0.0$ \\
$(-0.5, 0.5]$            & 186 & $0.0$ & $0.0$ & $48$ & $0.0$ \\
$(0.5, 2]$               & 107 & $1.5$ & $-0.5$ & $52$ & $-0.2$ \\
$(2, 5]$                 & 127 & $2.6$ & $0.4$ & $28$ & $8.2$ \\
$(5, 10]$                & 101 & $4.5$ & $2.9$ & $33$ & $7.8$ \\
$(10, 20]$               & 102 & $5.9$ & $7.6$ & $24$ & $13.1$ \\
$(20, 50]$               & 127 & $18.7$ & $14.2$ & $20$ & $17.7$ \\
$(50, \infty)$           & 108 & $157.7$ & $-38.3$ & $17$ & $9.8$ \\
\hline
All                      & 3381 & $0.0$ & $-2.9$ & $49$ & $0.0$ \\
  \end{tabular}
  
\end{table}

\clearpage
\subsection{Single-objective (SO) Results} \label{sec:appendix-single-objective}

\begin{table}[h!]
  \centering
\caption{Median DL of solutions produced by SO-DL, MO-DL, and MO-Len (with DL selection). Smallest DL is marked bold. MO-DL achieved the smallest median DL values for 30 of the 34 datasets.}
\begin{tabular}{l|r|r|r}
Dataset & SO-DL & MO-DL & MO-Len \\
\hline
Kotanchek                        & $-191$ & $\mathbf{-199}$ & $-196$ \\
RatPol2d                         & $63$ & $\mathbf{28}$ & $32$ \\
RatPol3d                         & $-373$ & $\mathbf{-378}$ & $-371$ \\
Ripple                           & $537$ & $163$ & $\mathbf{162}$ \\
Salustowicz                      & $-138$ & $\mathbf{-155}$ & $-150$ \\
Salustowicz2d                    & $708$ & $\mathbf{-409}$ & $-392$ \\
Friedman-1 $m=50$                & $100$ & $\mathbf{97}$ & $99$ \\
Friedman-1 $m=100$               & $186$ & $\mathbf{177}$ & $180$ \\
Friedman-1 $m=200$               & $334$ & $\mathbf{325}$ & $328$ \\
Friedman-1 $m=500$               & $774$ & $\mathbf{764}$ & $767$ \\
Friedman-1 $m=1000$              & $1486$ & $\mathbf{1475}$ & $1480$ \\
Friedman-2 $\sigma=0.5$          & $242$ & $147$ & $\mathbf{138}$ \\
Friedman-2 $\sigma=1.0$          & $251$ & $207$ & $\mathbf{199}$ \\
Friedman-2 $\sigma=2.0$          & $270$ & $\mathbf{260}$ & $\mathbf{260}$ \\
Friedman-2 $\sigma=4.8$          & $323$ & $\mathbf{322}$ & $323$ \\
Friedman-2 $\sigma=13.6$         & $408$ & $\mathbf{407}$ & $408$ \\
Nikuradse (2d)                   & $-710$ & $\mathbf{-824}$ & $-813$ \\
Nikuradse (1d)                   & $-345$ & $\mathbf{-350}$ & $-344$ \\
Chem. 1 Tower                    & $14337$ & $14233$ & $\mathbf{14197}$ \\
Chem. 2 Comp.                    & $-282$ & $\mathbf{-330}$ & $-312$ \\
Friction dyn.                    & $-4207$ & $\mathbf{-4478}$ & $-4301$ \\
Friction stat.                   & $-3882$ & $\mathbf{-4142}$ & $-4036$ \\
192 vineyard                     & $77$ & $\mathbf{76}$ & $79$ \\
210 cloud                        & $26$ & $\mathbf{22}$ & $25$ \\
522 pm10                         & $371$ & $\mathbf{352}$ & $356$ \\
557 analcat apnea1               & $2667$ & $\mathbf{2301}$ & $2305$ \\
579 fri c0 250 5                 & $123$ & $\mathbf{39}$ & $47$ \\
606 fri c2 1000 10               & $-185$ & $\mathbf{-199}$ & $-191$ \\
650 fri c0 500 50                & $187$ & $\mathbf{6}$ & $14$ \\
678 vis. env.                    & $\mathbf{169}$ & $\mathbf{169}$ & $170$ \\
1028 SWD                         & $600$ & $\mathbf{597}$ & $600$ \\
1089 USCrime                     & $127$ & $\mathbf{124}$ & $127$ \\
1193 BNG lowbwt                  & $7606$ & $\mathbf{7593}$ & $7605$ \\
1199 BNG echoMon                 & $3907$ & $\mathbf{3905}$ & $3911$ \\
\end{tabular}
\end{table}

\subsubsection{Test $R^2$ and test RMSE}
\Cref{tab:r2-so} shows the median test $R^2$ of the solutions produced by SO in comparison to MO-Len-DL for all datasets, together with the BBT win probabilities of MO-Len-DL against each SO variant. \Cref{tab:rmse-so} shows the 10\%, 50\%, and 90\% quantiles of the test RMSE of the solutions produced by SO in comparison to MO-Len-DL, as a complement to \Cref{tab:r2-so}. MO-Len-DL has the lowest median test RMSE for 18 of the 34 datasets. The median test RMSE of SO-BIC, SO-BIC$_\text{SR}$, SO-FBF, and SO-DL is a median $4\%$, $15\%$, $15\%$, and $17\%$ higher than that of MO-Len-DL, and the median test RMSE of SO-AIC is a median $55\%$ higher. The 95\% bootstrap confidence intervals of these differences in \Cref{tab:rmse-rel} are all above zero.

\begin{table}[ht]
\centering
\small
\caption{Median $R^2$ on the test set of solutions produced by SO in comparison to MO-Len with DL model selection (best is marked bold). The last two rows give the median posterior win probabilities of MO-Len with DL selection vs. each criterion from the BBT model. Substantially higher win probabilities outside of the ROPE $45\% - 55\%$ are marked with $\dagger$ (more than 99\% of the probability mass outside the ROPE, one match per dataset, see \Cref{sec:statistical-comparisons}). The 10\%, 50\%, and 90\% quantiles of the test RMSE are given in \Cref{tab:rmse-so}, and the median differences to MO-Len-DL with bootstrap confidence intervals in \Cref{tab:rmse-rel}.}\label{tab:r2-so}
\begin{tabular}{l|r|r|r|r|r|r}
Dataset & SO-AIC & SO-BIC & SO-BIC$_\text{SR}$ & SO-FBF & SO-DL & MO-Len-DL \\
\hline
Kotanchek                        & $-\infty$ & $0.9448$ & $0.9601$ & $0.9597$ & $0.9461$ & $\mathbf{0.9938}$ \\
RatPol2d                         & $0.5689$ & $0.8155$ & $-0.0787$ & $-0.0887$ & $-0.4514$ & $\mathbf{0.9808}$ \\
RatPol3d                         & $0.9988$ & $0.9996$ & $0.9995$ & $0.9995$ & $0.9996$ & $\mathbf{0.9997}$ \\
Ripple                           & $0.9984$ & $0.9996$ & $0.8873$ & $0.8859$ & $0.8874$ & $\mathbf{0.9998}$ \\
Salustowicz                      & $0.9898$ & $0.9961$ & $0.9894$ & $0.9830$ & $0.9893$ & $\mathbf{0.9962}$ \\
Salustowicz2d                    & $0.9388$ & $0.6321$ & $0.2324$ & $0.2039$ & $0.2535$ & $\mathbf{0.9967}$ \\
Friedman-1 $m=50$                & $0.2053$ & $\mathbf{0.8196}$ & $0.7768$ & $0.7634$ & $0.7826$ & $0.7938$ \\
Friedman-1 $m=100$               & $0.7496$ & $\mathbf{0.9498}$ & $0.8714$ & $0.8535$ & $0.8630$ & $0.9382$ \\
Friedman-1 $m=200$               & $0.9339$ & $\mathbf{0.9803}$ & $0.9609$ & $0.9533$ & $0.9684$ & $0.9773$ \\
Friedman-1 $m=500$               & $0.9769$ & $0.9914$ & $0.9853$ & $0.9857$ & $0.9904$ & $\mathbf{0.9929}$ \\
Friedman-1 $m=1000$              & $0.9907$ & $0.9950$ & $0.9940$ & $0.9941$ & $0.9941$ & $\mathbf{0.9953}$ \\
Friedman-2 $\sigma=0.5$          & $0.9908$ & $0.9957$ & $0.7457$ & $0.7430$ & $0.7472$ & $\mathbf{0.9982}$ \\
Friedman-2 $\sigma=1.0$          & $0.9685$ & $0.9814$ & $0.7400$ & $0.7402$ & $0.7426$ & $\mathbf{0.9948}$ \\
Friedman-2 $\sigma=2.0$          & $0.7863$ & $0.8573$ & $0.7339$ & $0.7293$ & $0.7358$ & $\mathbf{0.9092}$ \\
Friedman-2 $\sigma=4.8$          & $-0.1833$ & $\mathbf{0.5860}$ & $0.4805$ & $0.4656$ & $0.5577$ & $0.5129$ \\
Friedman-2 $\sigma=13.6$         & $-6.9236$ & $-0.5008$ & $-0.0496$ & $-0.0548$ & $\mathbf{-0.0032}$ & $-0.1375$ \\
Nikuradse (2d)                   & $0.9606$ & $0.9702$ & $0.9809$ & $0.9697$ & $0.9860$ & $\mathbf{0.9948}$ \\
Nikuradse (1d)                   & $0.9792$ & $0.9819$ & $\mathbf{0.9820}$ & $\mathbf{0.9820}$ & $0.9816$ & $0.9819$ \\
Chem. 1 Tower                    & $0.9480$ & $\mathbf{0.9492}$ & $0.9397$ & $0.9395$ & $0.9401$ & $0.9468$ \\
Chem. 2 Comp.                    & $0.6909$ & $0.7536$ & $0.7632$ & $\mathbf{0.7823}$ & $0.7607$ & $0.7486$ \\
Friction dyn.                    & $0.9065$ & $0.8983$ & $0.8829$ & $0.8951$ & $0.9005$ & $\mathbf{0.9357}$ \\
Friction stat.                   & $0.9089$ & $0.9044$ & $0.8820$ & $0.8802$ & $0.8800$ & $\mathbf{0.9131}$ \\
192 vineyard                     & $0.4143$ & $0.4517$ & $0.5872$ & $0.5759$ & $\mathbf{0.6245}$ & $0.5604$ \\
210 cloud                        & $0.0885$ & $0.7139$ & $0.8007$ & $\mathbf{0.8057}$ & $0.8030$ & $0.7949$ \\
522 pm10                         & $0.1718$ & $0.2255$ & $0.1459$ & $0.1561$ & $0.1607$ & $\mathbf{0.2557}$ \\
557 analcat apnea1               & $0.7729$ & $0.8100$ & $0.8412$ & $\mathbf{0.8499}$ & $0.1834$ & $0.7803$ \\
579 fri c0 250 5                 & $0.9365$ & $\mathbf{0.9472}$ & $0.7402$ & $0.7367$ & $0.7562$ & $0.9467$ \\
606 fri c2 1000 10               & $0.9770$ & $0.9775$ & $0.9767$ & $0.9770$ & $\mathbf{0.9778}$ & $\mathbf{0.9778}$ \\
650 fri c0 500 50                & $0.9497$ & $0.8701$ & $0.7928$ & $0.7945$ & $0.8442$ & $\mathbf{0.9562}$ \\
678 vis. env.                    & $0.0145$ & $0.2318$ & $\mathbf{0.3183}$ & $0.3169$ & $0.3000$ & $0.2971$ \\
1028 SWD                         & $0.3733$ & $\mathbf{0.3740}$ & $0.3484$ & $0.3463$ & $0.3630$ & $0.3654$ \\
1089 USCrime                     & $-1.6910$ & $0.6373$ & $\mathbf{0.7371}$ & $\mathbf{0.7371}$ & $0.7253$ & $0.7212$ \\
1193 BNG lowbwt                  & $\mathbf{0.5773}$ & $0.5739$ & $0.5624$ & $0.5652$ & $0.5657$ & $0.5699$ \\
1199 BNG echoMon                 & $0.4199$ & $\mathbf{0.4332}$ & $0.4254$ & $0.4256$ & $0.4258$ & $0.4255$ \\
\hline
Win prob. & DL$>$AIC & DL$>$BIC & DL$>$BIC$_\text{SR}$ & DL$>$FBF & DL$>$SO-DL & \\
($34$ datasets) & $85.2\%^\dagger$ & $61.4\%$ & $81.3\%^\dagger$ & $83.6\%^\dagger$ & $78.3\%^\dagger$ & \\
Num. best median $R^2$ & 1 & 8 & 3 & 5 & 3 & 17 \\
\end{tabular}
\end{table}

\begin{sidewaystable}
  \centering
  \footnotesize
  \setlength{\tabcolsep}{2.5pt}
  
  \caption{Quantiles of the test RMSE of the solutions produced by SO in comparison to MO-Len with DL model selection. P10, P50, and P90 are the 10\%, 50\% (median), and 90\% quantiles over the 100 runs; the lowest median is marked bold. For the synthetic datasets the test set is without noise. A non-finite test RMSE (e.g., caused by a pole of the expression within the range of the test inputs) counts as $\infty$.}
  \label{tab:rmse-so}
  \begin{tabular}{l|rrr|rrr|rrr|rrr|rrr|rrr}
Dataset & \multicolumn{3}{c|}{SO-AIC} & \multicolumn{3}{c|}{SO-BIC} & \multicolumn{3}{c|}{SO-BIC$_\text{SR}$} & \multicolumn{3}{c|}{SO-FBF} & \multicolumn{3}{c|}{SO-DL} & \multicolumn{3}{c}{MO-Len-DL} \\
 & P10 & P50 & P90 & P10 & P50 & P90 & P10 & P50 & P90 & P10 & P50 & P90 & P10 & P50 & P90 & P10 & P50 & P90 \\
\hline
Kotanchek                        & $0.0206$ & $\infty$ & $\infty$ & $0.0120$ & $0.0478$ & $\infty$ & $0.0109$ & $0.0407$ & $\infty$ & $0.00897$ & $0.0409$ & $\infty$ & $0.0142$ & $0.0473$ & $\infty$ & $0.00758$ & $\mathbf{0.0160}$ & $0.0735$ \\
RatPol2d                         & $0.555$ & $1.67$ & $739$ & $0.400$ & $1.09$ & $32.9$ & $0.819$ & $2.64$ & $132$ & $2.47$ & $2.65$ & $130$ & $2.38$ & $3.06$ & $141$ & $0.218$ & $\mathbf{0.352}$ & $0.827$ \\
RatPol3d                         & $0.0145$ & $0.0286$ & $0.0683$ & $0.0103$ & $0.0173$ & $0.0367$ & $0.00849$ & $0.0179$ & $0.0488$ & $0.00825$ & $0.0186$ & $0.0470$ & $0.00945$ & $0.0167$ & $0.0293$ & $0.00620$ & $\mathbf{0.0135}$ & $0.0379$ \\
Ripple                           & $0.0695$ & $0.163$ & $2.44$ & $0.0490$ & $0.0804$ & $1.16$ & $0.0652$ & $1.35$ & $1.36$ & $0.0702$ & $1.36$ & $1.36$ & $0.0785$ & $1.35$ & $1.38$ & $0.0306$ & $\mathbf{0.0537}$ & $0.0862$ \\
Salustowicz                      & $0.0191$ & $0.0307$ & $1.61$ & $0.0127$ & $0.0189$ & $0.0370$ & $0.0170$ & $0.0312$ & $0.223$ & $0.0181$ & $0.0396$ & $0.223$ & $0.0155$ & $0.0314$ & $0.224$ & $0.0137$ & $\mathbf{0.0187}$ & $0.0308$ \\
Salustowicz2d                    & $0.0340$ & $0.248$ & $\infty$ & $0.0327$ & $0.608$ & $\infty$ & $0.0451$ & $0.879$ & $\infty$ & $0.0521$ & $0.895$ & $\infty$ & $0.0420$ & $0.867$ & $\infty$ & $0.0295$ & $\mathbf{0.0576}$ & $0.144$ \\
Friedman-1 $m=50$                & $1.49$ & $2.23$ & $47.4$ & $0.745$ & $\mathbf{1.06}$ & $1.60$ & $0.995$ & $1.18$ & $1.51$ & $1.03$ & $1.21$ & $1.57$ & $0.983$ & $1.16$ & $1.45$ & $0.879$ & $1.13$ & $1.52$ \\
Friedman-1 $m=100$               & $0.803$ & $1.25$ & $20.0$ & $0.400$ & $\mathbf{0.559}$ & $0.792$ & $0.731$ & $0.895$ & $1.05$ & $0.756$ & $0.955$ & $1.11$ & $0.730$ & $0.924$ & $1.06$ & $0.436$ & $0.620$ & $0.921$ \\
Friedman-1 $m=200$               & $0.459$ & $0.642$ & $1.01$ & $0.240$ & $\mathbf{0.350}$ & $0.459$ & $0.376$ & $0.493$ & $0.758$ & $0.383$ & $0.540$ & $0.760$ & $0.376$ & $0.444$ & $0.724$ & $0.265$ & $0.376$ & $0.481$ \\
Friedman-1 $m=500$               & $0.283$ & $0.380$ & $0.575$ & $0.180$ & $0.232$ & $0.295$ & $0.203$ & $0.303$ & $0.408$ & $0.202$ & $0.299$ & $0.397$ & $0.201$ & $0.244$ & $0.392$ & $0.170$ & $\mathbf{0.210}$ & $0.304$ \\
Friedman-1 $m=1000$              & $0.172$ & $0.241$ & $0.318$ & $0.140$ & $0.176$ & $0.215$ & $0.166$ & $0.193$ & $0.224$ & $0.171$ & $0.192$ & $0.233$ & $0.170$ & $0.192$ & $0.223$ & $0.142$ & $\mathbf{0.171}$ & $0.199$ \\
Friedman-2 $\sigma=0.5$          & $0.293$ & $0.489$ & $1.59$ & $0.231$ & $0.336$ & $1.68$ & $1.90$ & $2.57$ & $2.68$ & $2.26$ & $2.58$ & $2.68$ & $1.70$ & $2.56$ & $2.66$ & $0.110$ & $\mathbf{0.216}$ & $0.286$ \\
Friedman-2 $\sigma=1.0$          & $0.555$ & $0.903$ & $3.69$ & $0.363$ & $0.694$ & $1.90$ & $2.44$ & $2.60$ & $2.67$ & $2.34$ & $2.60$ & $2.71$ & $1.98$ & $2.58$ & $2.67$ & $0.232$ & $\mathbf{0.366}$ & $0.900$ \\
Friedman-2 $\sigma=2.0$          & $1.47$ & $2.35$ & $6.08$ & $1.02$ & $1.92$ & $2.88$ & $2.56$ & $2.63$ & $3.13$ & $2.57$ & $2.65$ & $3.21$ & $2.56$ & $2.62$ & $2.95$ & $0.532$ & $\mathbf{1.53}$ & $2.62$ \\
Friedman-2 $\sigma=4.8$          & $3.89$ & $5.54$ & $49.9$ & $2.78$ & $\mathbf{3.28}$ & $7.68$ & $3.19$ & $3.67$ & $4.64$ & $3.20$ & $3.72$ & $4.62$ & $3.07$ & $3.39$ & $4.42$ & $2.86$ & $3.55$ & $4.26$ \\
Friedman-2 $\sigma=13.6$         & $6.83$ & $14.3$ & $82.1$ & $4.47$ & $6.24$ & $20.3$ & $4.62$ & $5.22$ & $6.14$ & $4.62$ & $5.23$ & $6.14$ & $4.61$ & $\mathbf{5.10}$ & $5.96$ & $4.71$ & $5.43$ & $7.74$ \\
Nikuradse (2d)                   & $0.00662$ & $0.0198$ & $0.0959$ & $0.00680$ & $0.0172$ & $0.152$ & $0.00695$ & $0.0138$ & $0.101$ & $0.00690$ & $0.0173$ & $0.0605$ & $0.00692$ & $0.0118$ & $0.0684$ & $0.00640$ & $\mathbf{0.00715}$ & $0.00932$ \\
Nikuradse (1d)                   & $0.0370$ & $0.0394$ & $0.0441$ & $0.0364$ & $0.0368$ & $0.0386$ & $0.0365$ & $\mathbf{0.0367}$ & $0.0388$ & $0.0365$ & $\mathbf{0.0367}$ & $0.0388$ & $0.0366$ & $0.0371$ & $0.0375$ & $0.0366$ & $0.0368$ & $0.0438$ \\
Chem. 1 Tower                    & $18.1$ & $19.0$ & $20.0$ & $17.9$ & $\mathbf{18.7}$ & $19.5$ & $19.6$ & $20.4$ & $21.3$ & $19.3$ & $20.4$ & $21.1$ & $19.5$ & $20.3$ & $21.3$ & $18.4$ & $19.2$ & $20.4$ \\
Chem. 2 Comp.                    & $0.142$ & $0.208$ & $0.451$ & $0.140$ & $0.186$ & $0.289$ & $0.156$ & $0.182$ & $0.281$ & $0.154$ & $\mathbf{0.175}$ & $0.227$ & $0.157$ & $0.183$ & $0.246$ & $0.144$ & $0.188$ & $0.300$ \\
Friction dyn.                    & $0.00264$ & $0.00364$ & $0.00758$ & $0.00278$ & $0.00379$ & $0.00671$ & $0.00353$ & $0.00407$ & $0.00444$ & $0.00323$ & $0.00385$ & $0.00433$ & $0.00290$ & $0.00375$ & $0.00606$ & $0.00254$ & $\mathbf{0.00302}$ & $0.00424$ \\
Friction stat.                   & $0.00338$ & $0.00372$ & $0.00425$ & $0.00343$ & $0.00381$ & $0.00443$ & $0.00381$ & $0.00424$ & $0.00472$ & $0.00380$ & $0.00427$ & $0.00477$ & $0.00378$ & $0.00427$ & $0.00529$ & $0.00326$ & $\mathbf{0.00364}$ & $0.00407$ \\
192 vineyard                     & $2.50$ & $3.13$ & $4.08$ & $2.34$ & $2.86$ & $4.71$ & $2.12$ & $2.78$ & $3.29$ & $2.09$ & $2.81$ & $3.42$ & $2.11$ & $\mathbf{2.66}$ & $3.29$ & $2.19$ & $2.74$ & $3.68$ \\
210 cloud                        & $0.476$ & $0.999$ & $10.4$ & $0.334$ & $0.546$ & $2.03$ & $0.306$ & $0.507$ & $8.14$ & $0.306$ & $0.507$ & $4.89$ & $0.310$ & $\mathbf{0.496}$ & $0.897$ & $0.305$ & $0.507$ & $0.648$ \\
522 pm10                         & $0.748$ & $0.807$ & $0.857$ & $0.722$ & $0.776$ & $0.817$ & $0.773$ & $0.816$ & $0.865$ & $0.761$ & $0.815$ & $0.861$ & $0.764$ & $0.810$ & $0.855$ & $0.712$ & $\mathbf{0.766}$ & $0.821$ \\
557 analcat apnea1               & $1084$ & $1447$ & $2087$ & $941$ & $1368$ & $1977$ & $888$ & $1209$ & $3138$ & $844$ & $\mathbf{1179}$ & $3077$ & $911$ & $2530$ & $3267$ & $988$ & $1478$ & $2029$ \\
579 fri c0 250 5                 & $0.226$ & $0.252$ & $0.313$ & $0.209$ & $\mathbf{0.227}$ & $0.251$ & $0.451$ & $0.507$ & $0.558$ & $0.461$ & $0.509$ & $0.560$ & $0.428$ & $0.485$ & $0.543$ & $0.209$ & $\mathbf{0.227}$ & $0.256$ \\
606 fri c2 1000 10               & $0.146$ & $0.152$ & $0.161$ & $0.144$ & $0.149$ & $0.160$ & $0.145$ & $0.152$ & $0.162$ & $0.145$ & $0.152$ & $0.162$ & $0.144$ & $0.150$ & $0.160$ & $0.143$ & $\mathbf{0.148}$ & $0.160$ \\
650 fri c0 500 50                & $0.206$ & $0.226$ & $0.389$ & $0.203$ & $0.359$ & $0.391$ & $0.399$ & $0.456$ & $0.507$ & $0.388$ & $0.453$ & $0.525$ & $0.358$ & $0.390$ & $0.461$ & $0.197$ & $\mathbf{0.210}$ & $0.238$ \\
678 vis. env.                    & $2.77$ & $3.43$ & $4.19$ & $2.59$ & $2.98$ & $3.44$ & $2.51$ & $\mathbf{2.80}$ & $3.24$ & $2.51$ & $\mathbf{2.80}$ & $3.24$ & $2.52$ & $2.82$ & $3.25$ & $2.51$ & $2.86$ & $3.32$ \\
1028 SWD                         & $0.619$ & $0.644$ & $0.665$ & $0.619$ & $\mathbf{0.641}$ & $0.660$ & $0.634$ & $0.654$ & $0.680$ & $0.633$ & $0.656$ & $0.676$ & $0.622$ & $0.647$ & $0.667$ & $0.623$ & $0.645$ & $0.669$ \\
1089 USCrime                     & $24.4$ & $59.2$ & $1089$ & $15.7$ & $22.2$ & $72.2$ & $17.5$ & $\mathbf{20.0}$ & $23.7$ & $17.5$ & $20.1$ & $27.3$ & $17.5$ & $20.3$ & $27.7$ & $16.3$ & $20.4$ & $29.0$ \\
1193 BNG lowbwt                  & $467$ & $\mathbf{471}$ & $479$ & $469$ & $473$ & $476$ & $475$ & $479$ & $484$ & $475$ & $478$ & $484$ & $474$ & $477$ & $483$ & $471$ & $475$ & $480$ \\
1199 BNG echoMon                 & $11.9$ & $12.0$ & $14.4$ & $11.8$ & $\mathbf{11.9}$ & $12.0$ & $11.9$ & $12.0$ & $12.0$ & $11.9$ & $12.0$ & $12.0$ & $11.9$ & $12.0$ & $12.0$ & $11.9$ & $12.0$ & $12.1$ \\
  \end{tabular}
  
\end{sidewaystable}

\subsubsection{Line charts for all datasets}
\begin{center}
\includegraphics[page=1]{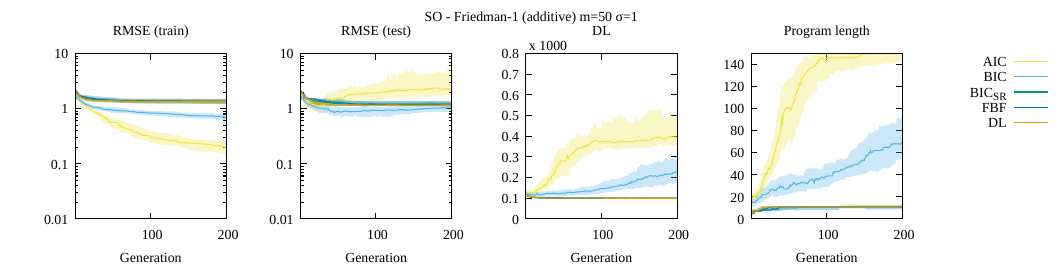}\\
\includegraphics[page=2]{neogp_linecharts_so.pdf}\\
\includegraphics[page=3]{neogp_linecharts_so.pdf}\\
\includegraphics[page=4]{neogp_linecharts_so.pdf}\\
\includegraphics[page=5]{neogp_linecharts_so.pdf}\\
\includegraphics[page=6]{neogp_linecharts_so.pdf}\\
\includegraphics[page=7]{neogp_linecharts_so.pdf}\\
\includegraphics[page=8]{neogp_linecharts_so.pdf}\\
\includegraphics[page=9]{neogp_linecharts_so.pdf}\\
\includegraphics[page=10]{neogp_linecharts_so.pdf}\\
\includegraphics[page=11]{neogp_linecharts_so.pdf}\\
\includegraphics[page=12]{neogp_linecharts_so.pdf}\\
\includegraphics[page=13]{neogp_linecharts_so.pdf}\\
\includegraphics[page=14]{neogp_linecharts_so.pdf}\\
\includegraphics[page=15]{neogp_linecharts_so.pdf}\\
\includegraphics[page=16]{neogp_linecharts_so.pdf}\\
\includegraphics[page=17]{neogp_linecharts_so.pdf}\\
\includegraphics[page=18]{neogp_linecharts_so.pdf}\\
\includegraphics[page=19]{neogp_linecharts_so.pdf}\\
\includegraphics[page=20]{neogp_linecharts_so.pdf}\\
\includegraphics[page=21]{neogp_linecharts_so.pdf}\\
\includegraphics[page=22]{neogp_linecharts_so.pdf}\\
\includegraphics[page=23]{neogp_linecharts_so.pdf}\\
\includegraphics[page=24]{neogp_linecharts_so.pdf}\\
\includegraphics[page=25]{neogp_linecharts_so.pdf}\\
\includegraphics[page=26]{neogp_linecharts_so.pdf}\\
\includegraphics[page=27]{neogp_linecharts_so.pdf}\\
\includegraphics[page=28]{neogp_linecharts_so.pdf}\\
\includegraphics[page=29]{neogp_linecharts_so.pdf}\\
\includegraphics[page=30]{neogp_linecharts_so.pdf}\\
\includegraphics[page=31]{neogp_linecharts_so.pdf}\\
\includegraphics[page=32]{neogp_linecharts_so.pdf}\\
\includegraphics[page=33]{neogp_linecharts_so.pdf}\\
\includegraphics[page=34]{neogp_linecharts_so.pdf}\\
\end{center}

\subsubsection{Effect of Hyperparameters on SO-DL}
The results of single-objective GP using DL for fitness calculation (SO-DL) with population size~1000, tournament group size~2, and mutation rate 5~\% showed early convergence with short expressions and high training RMSE (underfitting).
We analysed the effect of different hyperparameter settings, e.g., to reach different selection pressures, in a smaller experiment.
We executed ten runs for the grid of all combinations of population sizes $(1000, 10000, 20000)$, tournament group sizes $(2, 3, 5, 7)$, and mutation rates $(5 \%, 15 \%, 25 \%, 50\%)$ using the Friedman-2 dataset with $m=100$ and $\sigma_\text{err}=1.0$.

\begin{figure}
  \center
  \includegraphics[page=1]{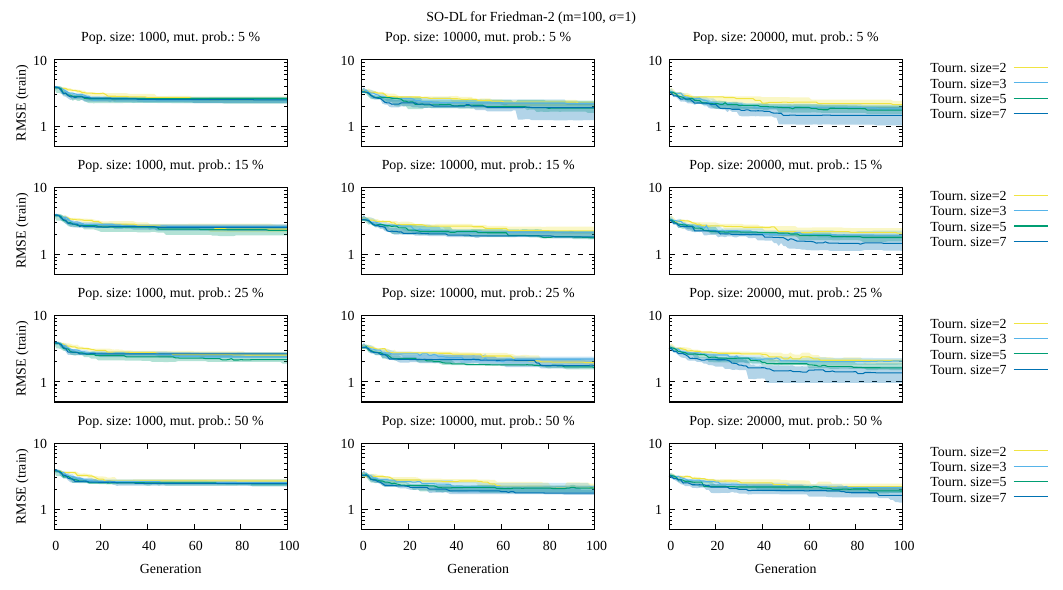}
  \caption{Training RMSE of SO-DL with different hyperparameter settings. The dashed line indicates the irreducible error threshold.}
  \label{fig:so-dl-hyperparameters-training-rmse}
\end{figure}

As shown in \Cref{fig:so-dl-hyperparameters-training-rmse} and \Cref{fig:so-dl-hyperparameters-size}, the results are similar to the setting used for all experiments. With larger population sizes and tournaments  the training error is noticeably lower, at 10–20× the evaluations, but the median stays clearly above $\sigma_\text{err}$. Higher mutation rates do not affect the results. We are confident in our claim that the complexity penalty imposed by the function complexity term is too strong for single-objective GP.

\begin{figure}
  \center
  \includegraphics[page=2]{neogp_linecharts_sodl_hyperparam.pdf}
  \caption{Size of the evolved expressions in SO-DL with different hyperparameter settings. The dashed line indicates the length of the generating function.}
  \label{fig:so-dl-hyperparameters-size}
\end{figure}

\subsection{Runtimes}\label{sec:appendix-runtime}
\subsubsection{Runtimes per dataset}
\Cref{tab:median-runtime-all-datasets-so} and \Cref{tab:median-runtime-all-datasets-mo} show the median runtimes for all datasets and all methods, and \Cref{tab:median-runtime-all-datasets-relative} shows the runtime differences relative to MO-Len-DL. The runtimes were measured in separate runs with the same code, settings, and seeds as the main experiments, but with only 10 runs per dataset (runs 1 to 10) and 100 generations (including the initial population). In these runs, only the selection criterion that is required by the method is calculated, so every MO-Len criterion has its own runs. All runs were executed single-threaded on a desktop PC: Apple M2 Pro, 32GB RAM, macOS Sequoia, Julia v1.13.1. Dataset 1089 USCrime is not included in the tables.

\begin{table}[h!]
  \centering
  
  \caption{Median runtimes (single-core) in seconds for SO and all datasets.}
  \label{tab:median-runtime-all-datasets-so}
  \begin{tabular}{l|rrrrr}
    Dataset  & SO-AIC & SO-BIC & SO-BIC$_\text{SR}$ & SO-DL & SO-FBF  \\
    \hline
    1028 SWD                     & $168.9$ & $51.3$ & $12.5$ & $28.3$ & $12.5$ \\
    1193 BNG lowbwt              & $109.4$ & $19.2$ & $7.5$ & $20.7$ & $7.9$ \\
    1199 BNG echoMonths          & $59.6$ & $20.3$ & $10.2$ & $21.0$ & $10.4$ \\
    192 vineyard                 & $4.2$ & $2.5$ & $0.6$ & $0.9$ & $0.5$ \\
    210 cloud                    & $20.9$ & $5.4$ & $0.8$ & $1.3$ & $0.8$ \\
    522 pm10                     & $49.7$ & $13.8$ & $1.9$ & $3.2$ & $2.0$ \\
    557 analcatdata apnea1       & $82.7$ & $44.4$ & $1.1$ & $1.5$ & $1.1$ \\
    579 fri c0 250 5             & $55.1$ & $34.1$ & $2.7$ & $6.3$ & $2.6$ \\
    606 fri c2 1000 10           & $308.9$ & $296.7$ & $334.7$ & $550.2$ & $327.7$ \\
    650 fri c0 500 50            & $71.4$ & $40.1$ & $3.1$ & $13.2$ & $3.6$ \\
    678 vis. env.                & $5.1$ & $1.8$ & $1.0$ & $1.6$ & $1.0$ \\
    Friedman-1 $m=100$           & $28.3$ & $12.5$ & $1.7$ & $3.3$ & $1.6$ \\
    Friedman-2 $\sigma=1.0$      & $26.8$ & $13.4$ & $1.8$ & $3.8$ & $1.6$ \\
    Kotanchek                    & $44.8$ & $41.7$ & $18.0$ & $30.9$ & $14.8$ \\
    Ratpol2d                     & $23.3$ & $15.7$ & $2.9$ & $4.1$ & $2.6$ \\
    Ratpol3d                     & $80.7$ & $78.6$ & $42.6$ & $114.7$ & $48.2$ \\
    Ripple                       & $105.0$ & $52.1$ & $16.4$ & $37.2$ & $18.3$ \\
    Salustowicz2d                & $191.3$ & $133.9$ & $40.2$ & $21.4$ & $60.9$ \\
    Salustowicz                  & $50.8$ & $39.0$ & $11.2$ & $6.6$ & $11.7$ \\
  \end{tabular}
  
\end{table}

\begin{table}[h!]
  \centering
  
  \caption{Median runtimes (single-core) in seconds for MO and all datasets.}
  \label{tab:median-runtime-all-datasets-mo}
  \begin{tabular}{l|rrrrrr}
    Dataset  & MO-Len-AIC & MO-Len-BIC & MO-Len-BIC$_\text{SR}$ & MO-Len-DL & MO-Len-FBF & MO-DL  \\
    \hline
    1028 SWD                     & $112.2$ & $112.9$ & $112.4$ & $116.1$ & $112.5$ & $128.7$ \\
    1193 BNG lowbwt              & $97.5$ & $97.7$ & $98.0$ & $100.8$ & $97.6$ & $129.1$ \\
    1199 BNG echoMonths          & $153.2$ & $153.2$ & $155.7$ & $161.0$ & $154.5$ & $121.3$ \\
    192 vineyard                 & $13.5$ & $13.5$ & $13.5$ & $13.7$ & $13.5$ & $17.0$ \\
    210 cloud                    & $19.1$ & $19.1$ & $19.2$ & $19.9$ & $19.1$ & $30.8$ \\
    522 pm10                     & $66.2$ & $66.2$ & $66.3$ & $68.2$ & $66.2$ & $81.5$ \\
    557 analcatdata apnea1       & $62.1$ & $62.2$ & $62.1$ & $63.0$ & $61.9$ & $49.2$ \\
    579 fri c0 250 5             & $36.7$ & $36.9$ & $36.6$ & $38.3$ & $36.6$ & $50.3$ \\
    606 fri c2 1000 10           & $148.5$ & $149.1$ & $149.8$ & $150.5$ & $147.1$ & $112.0$ \\
    650 fri c0 500 50            & $57.1$ & $57.0$ & $56.8$ & $59.1$ & $57.1$ & $79.9$ \\
    678 vis. env.                & $23.6$ & $23.7$ & $23.5$ & $24.2$ & $23.6$ & $24.2$ \\
    Friedman-1 $m=100$           & $23.5$ & $23.4$ & $23.5$ & $24.6$ & $23.5$ & $31.9$ \\
    Friedman-2 $\sigma=1.0$      & $21.7$ & $21.7$ & $21.7$ & $22.5$ & $21.7$ & $23.1$ \\
    Kotanchek                    & $32.7$ & $32.7$ & $32.7$ & $33.9$ & $32.6$ & $31.6$ \\
    Ratpol2d                     & $17.2$ & $17.2$ & $17.3$ & $18.0$ & $17.2$ & $13.5$ \\
    Ratpol3d                     & $53.5$ & $53.6$ & $53.7$ & $55.5$ & $53.6$ & $33.6$ \\
    Ripple                       & $59.3$ & $59.2$ & $59.2$ & $60.2$ & $59.1$ & $29.2$ \\
    Salustowicz2d                & $119.0$ & $119.1$ & $119.2$ & $123.2$ & $119.1$ & $67.7$ \\
    Salustowicz                  & $31.4$ & $31.5$ & $31.4$ & $32.4$ & $31.4$ & $18.7$ \\
  \end{tabular}
  
\end{table}

\begin{table}[h!]
  \centering
  
  \caption{Median runtime differences relative to MO-Len-DL runtime in percent.}
  \label{tab:median-runtime-all-datasets-relative}
  \begin{tabular}{l|rrrrr|rrrrr}
             & \multicolumn{5}{c|}{Single-objective} & \multicolumn{5}{c}{Multi-objective} \\
    Dataset  & AIC & BIC & BIC$_\text{SR}$ & DL & FBF & Len-AIC & Len-BIC & Len-BIC$_\text{SR}$ & Len-FBF & DL  \\
    \hline
    1028 SWD                     & $45.5$ & $-55.8$ & $-89.2$ & $-75.6$ & $-89.2$ & $-3.3$ & $-2.7$ & $-3.2$ & $-3.1$ & $10.8$ \\
    1193 BNG lowbwt              & $8.5$ & $-81.0$ & $-92.5$ & $-79.5$ & $-92.2$ & $-3.3$ & $-3.1$ & $-2.8$ & $-3.2$ & $28.0$ \\
    1199 BNG echoMonths          & $-63.0$ & $-87.4$ & $-93.7$ & $-87.0$ & $-93.5$ & $-4.8$ & $-4.9$ & $-3.3$ & $-4.0$ & $-24.7$ \\
    192 vineyard                 & $-69.2$ & $-81.7$ & $-95.8$ & $-93.6$ & $-96.3$ & $-1.7$ & $-1.8$ & $-1.4$ & $-1.6$ & $23.9$ \\
    210 cloud                    & $4.9$ & $-73.0$ & $-96.0$ & $-93.5$ & $-96.0$ & $-4.1$ & $-3.8$ & $-3.6$ & $-4.2$ & $54.7$ \\
    522 pm10                     & $-27.2$ & $-79.8$ & $-97.2$ & $-95.3$ & $-97.1$ & $-3.0$ & $-2.9$ & $-2.8$ & $-2.9$ & $19.5$ \\
    557 analcatdata apnea1       & $31.2$ & $-29.5$ & $-98.3$ & $-97.6$ & $-98.2$ & $-1.4$ & $-1.4$ & $-1.5$ & $-1.9$ & $-21.9$ \\
    579 fri c0 250 5             & $43.7$ & $-10.9$ & $-92.9$ & $-83.5$ & $-93.3$ & $-4.2$ & $-3.7$ & $-4.4$ & $-4.4$ & $31.3$ \\
    606 fri c2 1000 10           & $105.2$ & $97.1$ & $122.3$ & $265.5$ & $117.7$ & $-1.4$ & $-0.9$ & $-0.5$ & $-2.3$ & $-25.6$ \\
    650 fri c0 500 50            & $20.7$ & $-32.2$ & $-94.7$ & $-77.7$ & $-93.9$ & $-3.5$ & $-3.6$ & $-4.0$ & $-3.5$ & $35.1$ \\
    678 vis. env.                & $-79.1$ & $-92.6$ & $-96.0$ & $-93.4$ & $-95.9$ & $-2.2$ & $-1.8$ & $-2.6$ & $-2.5$ & $0.1$ \\
    Friedman-1 $m=100$           & $15.0$ & $-49.1$ & $-93.0$ & $-86.5$ & $-93.6$ & $-4.4$ & $-4.7$ & $-4.4$ & $-4.6$ & $29.9$ \\
    Friedman-2 $\sigma=1.0$      & $19.2$ & $-40.3$ & $-91.9$ & $-83.0$ & $-93.0$ & $-3.6$ & $-3.4$ & $-3.3$ & $-3.4$ & $2.8$ \\
    Kotanchek                    & $32.2$ & $23.1$ & $-47.0$ & $-8.6$ & $-56.2$ & $-3.5$ & $-3.4$ & $-3.5$ & $-3.7$ & $-6.8$ \\
    Ratpol2d                     & $29.2$ & $-13.1$ & $-83.7$ & $-77.0$ & $-85.6$ & $-4.5$ & $-4.4$ & $-4.2$ & $-4.5$ & $-25.2$ \\
    Ratpol3d                     & $45.4$ & $41.5$ & $-23.2$ & $106.6$ & $-13.2$ & $-3.5$ & $-3.5$ & $-3.3$ & $-3.4$ & $-39.4$ \\
    Ripple                       & $74.6$ & $-13.4$ & $-72.7$ & $-38.3$ & $-69.5$ & $-1.5$ & $-1.6$ & $-1.6$ & $-1.7$ & $-51.5$ \\
    Salustowicz2d                & $55.3$ & $8.7$ & $-67.4$ & $-82.6$ & $-50.5$ & $-3.5$ & $-3.3$ & $-3.2$ & $-3.4$ & $-45.0$ \\
    Salustowicz                  & $56.7$ & $20.2$ & $-65.4$ & $-79.6$ & $-64.1$ & $-3.1$ & $-3.1$ & $-3.1$ & $-3.3$ & $-42.4$ \\
    \hline
    Average                      & $18.4$ & $-28.9$ & $-72.0$ & $-50.5$ & $-71.2$ & $-3.2$ & $-3.1$ & $-3.0$ & $-3.2$ & $-2.4$ \\
  \end{tabular}
  
\end{table}

\subsubsection{Runtime profile of MO-DL and MO-Len-DL}\label{sec:appendix-runtime-profile}
To find out where the runtime goes, we repeated runs 1 to 3 of MO-DL and MO-Len-DL for all 20 datasets of the runtime experiment with an instrumented version of the run script.
We measure the time of each parameter fit (Levenberg-Marquardt), of the FIM and its SVD in the DL objective of MO-DL, of the selection of the model from the first front, and the remaining time (variation, non-dominated sorting, truncation). \Cref{tab:runtime-profile-shares} shows the shares of these parts in the runtime of the 100 generations after the initial population.

 Despite the cost of the DL objective, MO-DL is not slower than MO-Len-DL, because the expressions in the MO-DL population are shorter (median of the average length 31.5 vs.\ 44.3) and have fewer parameters (8.8 vs.\ 11.5). A fit in MO-Len-DL takes 1.96 times as long as a fit in MO-DL.
 Relative to the runtime of MO-Len-DL, MO-DL saves 46\% on parameter fitting and spends 39\% on the DL objective, a net difference of $-7\%$ (median over the datasets, $-48\%$ to $+86\%$) for these three runs.

\begin{table}[h!]
  \centering
  
  \caption{Shares of the runtime of the 100 generations after the initial population in percent (median [min, max] over the 20 datasets, three runs each).}
  \label{tab:runtime-profile-shares}
  \begin{tabular}{l|r|r}
    Part & MO-DL & MO-Len-DL \\
    \hline
    Parameter fitting (LM)           & $53.2$ [$46.6$, $62.5$] & $94.3$ [$91.6$, $96.3$] \\
    NLL objective                    & $0.6$ [$0.3$, $1.0$]    & $1.0$ [$0.7$, $1.5$] \\
    DL objective: FIM                & $34.1$ [$17.4$, $43.6$] & -- \\
    DL objective: SVD                & $3.4$ [$0.9$, $26.8$]   & -- \\
    DL objective: rest               & $0.3$ [$0.1$, $0.8$]    & -- \\
    DL selection: FIM + SVD          & $3.8$ [$2.5$, $4.9$]    & $3.1$ [$2.0$, $4.3$] \\
    DL selection: rest               & $0.1$ [$0.0$, $0.2$]    & $0.1$ [$0.1$, $0.1$] \\
    Rest                             & $1.4$ [$0.4$, $3.6$]    & $1.3$ [$0.4$, $4.9$] \\
  \end{tabular}
  
\end{table}

\clearpage
\subsection{Ablation of the parameter complexity}\label{sec:appendix-ablation}
We compare the parameter complexity of DL, which uses the full FIM and a rotated lattice (\cref{eqn:dl}), with two other variants and with DL without parameter complexity:
\begin{itemize}
  \item DL$_\text{diag}$: the axis-parallel lattice with the precisions from the diagonal of the FIM, as in \citet{Bartlett_ESR_2023}, i.e., $p \log 2 + \sum_{i=1}^{p} \max\left(0, \tfrac{1}{2}\log \mathrm{FIM}_{ii} + \log |\hat{\vartheta}_i| - \tfrac{1}{2}\log 12\right)$.
  \item DL$_\text{det}$: the rectangular lattice with free orientation described by \citet{Bartlett_Priors_2023}, which replaces the precisions of the individual parameters by the determinant of the FIM in relative coordinates, i.e., $p \log 2 + \sum_{i=1}^{p} \max\left(0, \tfrac{1}{2}\log \mu_i + \tfrac{1}{2}\log \nu_p - \log 2\right)$, where $\mu_i$ are the eigenvalues of $D\,\mathrm{FIM}\,D$ with $D = \mathrm{diag}(|\hat{\vartheta}|)$ and $\log \nu_p = 1 - \log 3$.
  \item NLL + func.compl.: DL without the parameter complexity term.
\end{itemize}
As for DL, the sign of each parameter always costs $\log 2$, negative eigenvalues of the FIM count as zero precision, and negative contributions are clamped to zero. All variants select the expressions from the same final MO-Len Pareto fronts as DL ($34 \times 100$ runs), so the differences between them are caused only by the selection. \Cref{tab:ablation-r2} shows the median test $R^2$ and the BBT win probabilities of DL against each variant, \Cref{tab:ablation-length} the median length, and \Cref{tab:ablation-complexity} the median function and parameter complexity of the selected expressions.

The test RMSE of the expressions selected with all three parameter complexities is similar  ($P(\text{DL}>\text{DL}_\text{diag}) = 55.5\%$, $P(\text{DL}>\text{DL}_\text{det}) = 48.5\%$). DL and DL$_\text{diag}$ select the same expression in 61\% of the runs, DL and DL$_\text{det}$ in 49\%. Without parameter complexity, the selected expressions are on average 25\% longer than with DL (without 1089 USCrime), and the median test $R^2$ is slightly higher for many datasets, but clearly lower for 1089 USCrime, Friedman-2 $\sigma=13.6$, and 678 vis. env., where NLL + func.compl. selects overfit expressions. Over all datasets $P(\text{DL}>\text{NLL + func.compl.}) = 37.8\%$, 10\% of the probability mass in the ROPE.

For the expressions selected by DL, the function complexity is larger than the parameter complexity for 24 of the 34 datasets, and the parameter complexity is a median 41\% of the penalty (function and parameter complexity).

\begin{table}[h!]
  \centering
  
  \caption{Median $R^2$ on the test set of the expressions selected from the MO-Len Pareto fronts by DL with the different parameter complexities and without parameter complexity (best is marked bold). The last two rows give the median posterior win probabilities of DL vs. each variant from the BBT model (one match per dataset). None of the win probabilities has more than 99\% of the probability mass outside the ROPE $45\% - 55\%$.}
  \label{tab:ablation-r2}
  \begin{tabular}{l|r|r|r|r}
Dataset & DL & DL$_\text{diag}$ & DL$_\text{det}$ & NLL + func.compl. \\
\hline
Kotanchek                        & $0.9938$ & $0.9920$ & $0.9942$ & $\mathbf{0.9960}$ \\
RatPol2d                         & $0.9808$ & $0.9796$ & $\mathbf{0.9812}$ & $0.9762$ \\
RatPol3d                         & $0.9997$ & $0.9997$ & $0.9997$ & $\mathbf{0.9998}$ \\
Ripple                           & $\mathbf{0.9998}$ & $\mathbf{0.9998}$ & $\mathbf{0.9998}$ & $\mathbf{0.9998}$ \\
Salustowicz                      & $0.9962$ & $0.9957$ & $\mathbf{0.9966}$ & $0.9965$ \\
Salustowicz2d                    & $0.9967$ & $0.9973$ & $0.9976$ & $\mathbf{0.9982}$ \\
Friedman-1 $m=50$                & $0.7938$ & $0.7853$ & $0.7912$ & $\mathbf{0.8259}$ \\
Friedman-1 $m=100$               & $0.9382$ & $0.9370$ & $0.9433$ & $\mathbf{0.9511}$ \\
Friedman-1 $m=200$               & $0.9773$ & $0.9764$ & $0.9772$ & $\mathbf{0.9790}$ \\
Friedman-1 $m=500$               & $\mathbf{0.9929}$ & $0.9925$ & $0.9924$ & $0.9925$ \\
Friedman-1 $m=1000$              & $0.9953$ & $0.9954$ & $0.9954$ & $\mathbf{0.9957}$ \\
Friedman-2 $\sigma=0.5$          & $0.9982$ & $0.9981$ & $0.9982$ & $\mathbf{0.9983}$ \\
Friedman-2 $\sigma=1.0$          & $0.9948$ & $0.9947$ & $\mathbf{0.9952}$ & $0.9949$ \\
Friedman-2 $\sigma=2.0$          & $0.9092$ & $0.8949$ & $0.9028$ & $\mathbf{0.9537}$ \\
Friedman-2 $\sigma=4.8$          & $0.5129$ & $\mathbf{0.5296}$ & $0.5273$ & $0.5238$ \\
Friedman-2 $\sigma=13.6$         & $-0.1375$ & $\mathbf{-0.1174}$ & $\mathbf{-0.1174}$ & $-1.2115$ \\
Nikuradse (2d)                   & $\mathbf{0.9948}$ & $0.9946$ & $\mathbf{0.9948}$ & $\mathbf{0.9948}$ \\
Nikuradse (1d)                   & $0.9819$ & $0.9818$ & $0.9819$ & $\mathbf{0.9820}$ \\
Chem. 1 Tower                    & $0.9468$ & $0.9414$ & $0.9492$ & $\mathbf{0.9508}$ \\
Chem. 2 Comp.                    & $0.7486$ & $0.7122$ & $0.7543$ & $\mathbf{0.7632}$ \\
Friction dyn.                    & $\mathbf{0.9357}$ & $0.9336$ & $0.9345$ & $0.9322$ \\
Friction stat.                   & $0.9131$ & $0.9117$ & $0.9132$ & $\mathbf{0.9158}$ \\
192 vineyard                     & $\mathbf{0.5604}$ & $0.5352$ & $0.5542$ & $0.5544$ \\
210 cloud                        & $\mathbf{0.7949}$ & $0.7938$ & $0.7932$ & $0.7644$ \\
522 pm10                         & $\mathbf{0.2557}$ & $0.2371$ & $0.2419$ & $0.2520$ \\
557 analcat apnea1               & $0.7803$ & $\mathbf{0.7940}$ & $0.7862$ & $0.7643$ \\
579 fri c0 250 5                 & $0.9467$ & $0.9475$ & $0.9495$ & $\mathbf{0.9498}$ \\
606 fri c2 1000 10               & $0.9778$ & $0.9781$ & $0.9781$ & $\mathbf{0.9782}$ \\
650 fri c0 500 50                & $0.9562$ & $0.9563$ & $0.9570$ & $\mathbf{0.9578}$ \\
678 vis. env.                    & $0.2971$ & $\mathbf{0.3000}$ & $0.2971$ & $0.2313$ \\
1028 SWD                         & $0.3654$ & $0.3644$ & $0.3663$ & $\mathbf{0.3706}$ \\
1089 USCrime                     & $\mathbf{0.7212}$ & $0.6817$ & $0.6623$ & $-0.2064$ \\
1193 BNG lowbwt                  & $0.5699$ & $0.5693$ & $0.5705$ & $\mathbf{0.5743}$ \\
1199 BNG echoMon                 & $0.4255$ & $0.4250$ & $0.4255$ & $\mathbf{0.4265}$ \\
\hline
Win prob. & & DL$>$DL$_\text{diag}$ & DL$>$DL$_\text{det}$ & DL$>$NLL + func.compl. \\
($34$ datasets) & & $55.5$ & $48.5$ & $37.8$ \\
  \end{tabular}
  
\end{table}

\begin{table}[h!]
  \centering
  
  \caption{Median length of the expressions selected from the MO-Len Pareto fronts by DL with the different parameter complexities and without parameter complexity (shortest is marked bold).}
  \label{tab:ablation-length}
  \begin{tabular}{l|r|r|r|r}
Dataset & DL & DL$_\text{diag}$ & DL$_\text{det}$ & NLL + func.compl. \\
\hline
Kotanchek                        & $17.0$ & $\mathbf{16.0}$ & $17.0$ & $19.0$ \\
RatPol2d                         & $22.0$ & $\mathbf{21.0}$ & $22.0$ & $24.0$ \\
RatPol3d                         & $\mathbf{15.0}$ & $16.0$ & $16.0$ & $\mathbf{15.0}$ \\
Ripple                           & $\mathbf{20.5}$ & $\mathbf{20.5}$ & $22.0$ & $22.0$ \\
Salustowicz                      & $24.0$ & $\mathbf{23.0}$ & $26.0$ & $26.0$ \\
Salustowicz2d                    & $\mathbf{35.0}$ & $\mathbf{35.0}$ & $38.0$ & $37.0$ \\
Friedman-1 $m=50$                & $13.0$ & $\mathbf{12.0}$ & $\mathbf{12.0}$ & $18.0$ \\
Friedman-1 $m=100$               & $20.0$ & $\mathbf{19.0}$ & $20.0$ & $22.0$ \\
Friedman-1 $m=200$               & $\mathbf{24.0}$ & $\mathbf{24.0}$ & $\mathbf{24.0}$ & $26.0$ \\
Friedman-1 $m=500$               & $27.0$ & $\mathbf{26.0}$ & $27.0$ & $28.0$ \\
Friedman-1 $m=1000$              & $\mathbf{27.0}$ & $\mathbf{27.0}$ & $28.0$ & $30.0$ \\
Friedman-2 $\sigma=0.5$          & $28.0$ & $\mathbf{27.0}$ & $29.0$ & $28.0$ \\
Friedman-2 $\sigma=1.0$          & $\mathbf{26.0}$ & $\mathbf{26.0}$ & $27.0$ & $27.0$ \\
Friedman-2 $\sigma=2.0$          & $21.0$ & $\mathbf{20.0}$ & $\mathbf{20.0}$ & $26.0$ \\
Friedman-2 $\sigma=4.8$          & $\mathbf{10.0}$ & $\mathbf{10.0}$ & $\mathbf{10.0}$ & $14.0$ \\
Friedman-2 $\sigma=13.6$         & $\mathbf{4.0}$ & $\mathbf{4.0}$ & $\mathbf{4.0}$ & $9.0$ \\
Nikuradse (2d)                   & $39.0$ & $\mathbf{38.0}$ & $43.0$ & $43.0$ \\
Nikuradse (1d)                   & $\mathbf{13.0}$ & $\mathbf{13.0}$ & $14.0$ & $\mathbf{13.0}$ \\
Chem. 1 Tower                    & $208.0$ & $\mathbf{149.5}$ & $256.5$ & $299.5$ \\
Chem. 2 Comp.                    & $62.5$ & $\mathbf{49.0}$ & $77.0$ & $83.0$ \\
Friction dyn.                    & $\mathbf{145.5}$ & $146.5$ & $201.5$ & $214.0$ \\
Friction stat.                   & $97.5$ & $\mathbf{96.5}$ & $136.0$ & $158.0$ \\
192 vineyard                     & $\mathbf{7.0}$ & $\mathbf{7.0}$ & $\mathbf{7.0}$ & $\mathbf{7.0}$ \\
210 cloud                        & $11.0$ & $\mathbf{10.0}$ & $11.0$ & $17.0$ \\
522 pm10                         & $15.0$ & $\mathbf{14.5}$ & $15.0$ & $18.5$ \\
557 analcat apnea1               & $39.5$ & $\mathbf{39.0}$ & $44.0$ & $68.0$ \\
579 fri c0 250 5                 & $\mathbf{32.0}$ & $32.5$ & $34.0$ & $34.5$ \\
606 fri c2 1000 10               & $38.0$ & $\mathbf{36.0}$ & $38.0$ & $37.0$ \\
650 fri c0 500 50                & $33.0$ & $\mathbf{32.0}$ & $35.0$ & $34.0$ \\
678 vis. env.                    & $\mathbf{5.0}$ & $\mathbf{5.0}$ & $\mathbf{5.0}$ & $10.0$ \\
1028 SWD                         & $17.0$ & $\mathbf{15.5}$ & $16.0$ & $19.5$ \\
1089 USCrime                     & $7.0$ & $\mathbf{5.0}$ & $8.0$ & $120.0$ \\
1193 BNG lowbwt                  & $18.0$ & $\mathbf{17.0}$ & $18.0$ & $23.0$ \\
1199 BNG echoMon                 & $13.0$ & $\mathbf{12.5}$ & $13.0$ & $16.0$ \\
  \end{tabular}
  
\end{table}

\begin{table}[h!]
  \centering
  
  \caption{Median function complexity (FC) and parameter complexity (PC) in nats of the expressions selected from the MO-Len Pareto fronts by DL with the different parameter complexities. PC is the parameter complexity of the variant used for the selection. For NLL + func.compl. only FC is part of the criterion.}
  \label{tab:ablation-complexity}
  \begin{tabular}{l|rr|rr|rr|r}
    & \multicolumn{2}{c|}{DL} & \multicolumn{2}{c|}{DL$_\text{diag}$} & \multicolumn{2}{c|}{DL$_\text{det}$} & NLL + func.compl. \\
    Dataset & FC & PC & FC & PC & FC & PC & FC \\
    \hline
Kotanchek                        & $20.3$ & $19.4$ & $19.7$ & $22.1$ & $20.8$ & $21.1$ & $20.8$ \\
RatPol2d                         & $25.3$ & $23.7$ & $25.0$ & $26.8$ & $25.3$ & $25.5$ & $27.1$ \\
RatPol3d                         & $19.5$ & $20.9$ & $21.1$ & $25.9$ & $21.1$ & $24.2$ & $19.5$ \\
Ripple                           & $25.3$ & $29.1$ & $25.3$ & $33.6$ & $26.9$ & $30.2$ & $25.3$ \\
Salustowicz                      & $29.1$ & $25.1$ & $27.9$ & $28.7$ & $31.2$ & $22.5$ & $30.8$ \\
Salustowicz2d                    & $43.7$ & $43.9$ & $42.8$ & $54.3$ & $47.3$ & $27.4$ & $45.7$ \\
Friedman-1 $m=50$                & $14.3$ & $6.3$ & $13.6$ & $6.8$ & $13.6$ & $8.1$ & $20.3$ \\
Friedman-1 $m=100$               & $22.9$ & $12.1$ & $22.9$ & $13.4$ & $22.9$ & $15.2$ & $27.0$ \\
Friedman-1 $m=200$               & $29.1$ & $16.6$ & $28.6$ & $19.4$ & $28.8$ & $20.5$ & $32.2$ \\
Friedman-1 $m=500$               & $34.8$ & $19.8$ & $34.5$ & $23.9$ & $34.8$ & $23.6$ & $35.2$ \\
Friedman-1 $m=1000$              & $36.8$ & $23.0$ & $35.2$ & $28.1$ & $38.4$ & $26.0$ & $39.1$ \\
Friedman-2 $\sigma=0.5$          & $38.4$ & $26.3$ & $36.8$ & $32.8$ & $39.1$ & $28.0$ & $36.8$ \\
Friedman-2 $\sigma=1.0$          & $36.0$ & $21.0$ & $36.4$ & $27.3$ & $37.1$ & $25.3$ & $36.8$ \\
Friedman-2 $\sigma=2.0$          & $29.9$ & $13.6$ & $25.1$ & $14.1$ & $26.4$ & $16.7$ & $34.5$ \\
Friedman-2 $\sigma=4.8$          & $10.8$ & $5.0$ & $10.8$ & $5.0$ & $10.8$ & $6.1$ & $15.6$ \\
Friedman-2 $\sigma=13.6$         & $1.4$ & $3.2$ & $1.4$ & $3.2$ & $1.4$ & $4.2$ & $8.0$ \\
Nikuradse (2d)                   & $52.7$ & $47.3$ & $51.3$ & $65.8$ & $58.6$ & $44.1$ & $57.4$ \\
Nikuradse (1d)                   & $12.5$ & $20.2$ & $12.5$ & $24.5$ & $14.5$ & $22.8$ & $11.3$ \\
Chem. 1 Tower                    & $383.2$ & $153.8$ & $263.5$ & $294.1$ & $465.3$ & $88.9$ & $536.6$ \\
Chem. 2 Comp.                    & $94.8$ & $48.2$ & $73.1$ & $83.3$ & $117.3$ & $34.9$ & $130.0$ \\
Friction dyn.                    & $288.2$ & $130.6$ & $282.8$ & $165.8$ & $382.2$ & $65.8$ & $418.0$ \\
Friction stat.                   & $182.6$ & $108.2$ & $176.1$ & $122.8$ & $255.4$ & $59.7$ & $305.6$ \\
192 vineyard                     & $4.4$ & $6.1$ & $3.3$ & $6.7$ & $3.3$ & $7.0$ & $3.3$ \\
210 cloud                        & $12.5$ & $7.2$ & $11.3$ & $8.0$ & $11.9$ & $9.0$ & $17.9$ \\
522 pm10                         & $18.7$ & $8.8$ & $17.5$ & $9.2$ & $18.7$ & $10.5$ & $22.9$ \\
557 analcat apnea1               & $62.1$ & $62.2$ & $63.1$ & $64.8$ & $72.9$ & $37.5$ & $116.1$ \\
579 fri c0 250 5                 & $44.3$ & $23.5$ & $43.7$ & $29.2$ & $45.1$ & $24.7$ & $45.6$ \\
606 fri c2 1000 10               & $48.4$ & $52.4$ & $46.1$ & $55.1$ & $50.0$ & $25.6$ & $48.0$ \\
650 fri c0 500 50                & $43.2$ & $28.3$ & $41.4$ & $34.7$ & $45.6$ & $25.0$ & $43.5$ \\
678 vis. env.                    & $3.3$ & $4.6$ & $2.3$ & $5.4$ & $3.3$ & $6.5$ & $7.6$ \\
1028 SWD                         & $22.9$ & $9.8$ & $22.0$ & $10.1$ & $22.0$ & $11.0$ & $26.4$ \\
1089 USCrime                     & $4.4$ & $7.0$ & $1.4$ & $8.4$ & $6.2$ & $8.8$ & $187.9$ \\
1193 BNG lowbwt                  & $22.9$ & $14.9$ & $20.3$ & $15.7$ & $22.0$ & $17.4$ & $28.8$ \\
1199 BNG echoMon                 & $14.0$ & $10.1$ & $13.1$ & $10.9$ & $14.3$ & $12.3$ & $17.5$ \\
  \end{tabular}
  
\end{table}

\subsection{Recovery of the generating functions}\label{sec:appendix-recovery}
Checking symbolic equivalence of the selected expressions with the generating functions is difficult. We therefore check functional equivalence numerically. For each expression selected for the synthetic datasets, all numerical constants are treated as parameters and refitted on the noise-free test set, starting from the values estimated on the training set. If the expression is a form of the generating function, the test RMSE after refitting is zero up to the numerical precision. We count the generating function as recovered if the test RMSE after refitting is below $10^{-6}\,\sigma_\text{err}$.
For all most of the synthetic datasets, the test sets cover a larger domain than the training sets (\Cref{tab:synthetic-datasets}), so approximations that are accurate only in the training domain are not counted. However, expressions that contain the generating function as a special case, e.g., with additional terms whose parameters can be refitted to zero, also count as recovered.

\Cref{tab:recovery} shows the number of runs in which the generating function was recovered. The generating functions are recovered only for the datasets with low noise and short generating functions, mainly for Ripple and for Friedman-2 with $\sigma_\text{err} \le 2$.
With MO-Len, BIC$_\text{SR}$, FBF, and DL recover the generating function in practically the same number of runs (140 to 141 of 1600), BIC in 90, and AIC in only 39. The expressions recovered by AIC are much longer (median length 108 instead of 22), i.e., they contain the generating function together with additional terms. DL$_\text{diag}$, DL$_\text{det}$, and NLL + func.compl.\ (\Cref{sec:appendix-ablation}) recover the generating function in 144, 128, and 140 runs, respectively. MO-DL recovers the generating function in fewer runs than MO-Len-DL (116), mainly because of Friedman-2 (13 instead of 41 runs), but more often for RatPol3d (13 instead of 1 run). All SO variants recover the generating function in fewer runs (21 to 55) than MO-Len-DL, which is consistent with the premature convergence of SO (\Cref{sec:appendix-single-objective}).

\begin{table}[h!]
  \centering
  \footnotesize
  \setlength{\tabcolsep}{3pt}
  
  \caption{Number of runs (of 100) in which the selected expression is a form of the generating function, i.e., in which its test RMSE after refitting the parameters on the noise-free test set is below $10^{-6}\,\sigma_\text{err}$ (the largest number is marked bold). The last row gives the median length of the recovered expressions.}
  \label{tab:recovery}
  \begin{tabular}{l|rrrrr|r|rrrrr}
 & \multicolumn{5}{c|}{MO-Len} & & \multicolumn{5}{c}{SO} \\
Dataset & AIC & BIC & BIC$_\text{SR}$ & FBF & DL & MO-DL & AIC & BIC & BIC$_\text{SR}$ & FBF & DL \\
\hline
Kotanchek                        & $0$ & $0$ & $3$ & $3$ & $\mathbf{4}$ & $2$ & $0$ & $0$ & $0$ & $0$ & $0$ \\
RatPol2d                         & $0$ & $0$ & $0$ & $0$ & $0$ & $0$ & $0$ & $0$ & $0$ & $0$ & $0$ \\
RatPol3d                         & $0$ & $0$ & $2$ & $2$ & $1$ & $\mathbf{13}$ & $1$ & $1$ & $6$ & $5$ & $1$ \\
Ripple                           & $37$ & $77$ & $90$ & $90$ & $\mathbf{91}$ & $83$ & $29$ & $51$ & $15$ & $14$ & $16$ \\
Salustowicz                      & $0$ & $0$ & $0$ & $0$ & $0$ & $0$ & $0$ & $\mathbf{2}$ & $0$ & $1$ & $1$ \\
Salustowicz2d                    & $1$ & $4$ & $4$ & $4$ & $3$ & $4$ & $1$ & $0$ & $0$ & $2$ & $\mathbf{5}$ \\
Friedman-1 $m=50$                & $0$ & $0$ & $0$ & $0$ & $0$ & $0$ & $0$ & $0$ & $0$ & $0$ & $0$ \\
Friedman-1 $m=100$               & $0$ & $0$ & $0$ & $0$ & $0$ & $0$ & $0$ & $0$ & $0$ & $0$ & $0$ \\
Friedman-1 $m=200$               & $0$ & $0$ & $0$ & $0$ & $0$ & $0$ & $0$ & $0$ & $0$ & $0$ & $0$ \\
Friedman-1 $m=500$               & $0$ & $0$ & $0$ & $0$ & $0$ & $0$ & $0$ & $0$ & $0$ & $0$ & $0$ \\
Friedman-1 $m=1000$              & $0$ & $0$ & $0$ & $0$ & $0$ & $\mathbf{1}$ & $0$ & $0$ & $0$ & $0$ & $0$ \\
Friedman-2 $\sigma=0.5$          & $1$ & $7$ & $\mathbf{18}$ & $\mathbf{18}$ & $\mathbf{18}$ & $6$ & $0$ & $1$ & $0$ & $0$ & $0$ \\
Friedman-2 $\sigma=1.0$          & $0$ & $1$ & $\mathbf{17}$ & $\mathbf{17}$ & $15$ & $5$ & $0$ & $0$ & $0$ & $0$ & $0$ \\
Friedman-2 $\sigma=2.0$          & $0$ & $1$ & $7$ & $7$ & $\mathbf{8}$ & $2$ & $0$ & $0$ & $0$ & $0$ & $0$ \\
Friedman-2 $\sigma=4.8$          & $0$ & $0$ & $0$ & $0$ & $0$ & $0$ & $0$ & $0$ & $0$ & $0$ & $0$ \\
Friedman-2 $\sigma=13.6$         & $0$ & $0$ & $0$ & $0$ & $0$ & $0$ & $0$ & $0$ & $0$ & $0$ & $0$ \\
\hline
Total (of 1600)                  & $39$ & $90$ & $\mathbf{141}$ & $\mathbf{141}$ & $140$ & $116$ & $31$ & $55$ & $21$ & $22$ & $23$ \\
Median length                    & $108$ & $25$ & $22$ & $22$ & $22$ & $28$ & $60$ & $38$ & $36$ & $42$ & $48$ \\
  \end{tabular}
  
\end{table}

\end{document}